\documentclass[letterpaper,journal]{IEEEtran}
\usepackage[caption=false,font=normalsize,labelfont=sf,textfont=sf]{subfig}
\usepackage{marvosym}
\usepackage{stfloats}
\usepackage{verbatim}
\usepackage{cite}
\usepackage{graphicx} 
\usepackage{multirow} 
\usepackage{amsmath, amssymb, amsfonts} 
\usepackage{amsthm} 
\usepackage{mathrsfs} 
\usepackage{balance}
\usepackage{xcolor} 
\usepackage{textcomp} 
\usepackage{booktabs} 
\usepackage{algorithm} 
\usepackage{algpseudocode} %
\usepackage{listings} 
\usepackage{array} 
\usepackage{tabularx} 
\usepackage{url} 
\usepackage{makecell} 
\usepackage[table]{xcolor} 
\definecolor{deperr}{RGB}{214,228,247}
\definecolor{depacc}{RGB}{218,232,213}
\colorlet{metricblue}{deperr}
\colorlet{metricred}{depacc}

\usepackage{placeins}
\usepackage[hidelinks,hypertexnames=false,colorlinks=true,citecolor=red!70!black,urlcolor=black]{hyperref}
\usepackage{orcidlink}
\usepackage{etoolbox}

\newcommand{\settablefont}{\footnotesize}

\AtBeginEnvironment{table}{%
    \settablefont
    \setlength{\tabcolsep}{1.8pt}%
}

\AtBeginEnvironment{table*}{%
    \settablefont
    \setlength{\tabcolsep}{1.8pt}%
}
\usepackage{tikz}
\newcolumntype{L}[1]{>{\raggedright\arraybackslash}p{#1}}
\newcolumntype{C}[1]{>{\centering\arraybackslash}p{#1}}
\newcolumntype{R}[1]{>{\raggedleft\arraybackslash}p{#1}}

\newtheorem{proposition}{Proposition}

\begin{document}
\bstctlcite{BSTcontrol}

\title{

Scale-Consistent Monocular Fisheye Visual Odometry via Spherical Bundle Adjustment-Guided Unsupervised Depth Learning
}

\author{    
Wanhui~Luo\,\orcidlink{0009-0008-8591-232X},    
Hongbo~Zhao\,\orcidlink{0009-0008-2198-2484},    
Yi~Zhou\,\orcidlink{0000-0003-3201-8873},    
Qijun~Chen\,\orcidlink{0000-0001-5644-1188},    
~\IEEEmembership{Senior Member,~IEEE},\\    
Yiming~Nie\,\orcidlink{0000-0003-0421-595X},    
and Rui~Fan\,\orcidlink{0000-0003-2593-6596}%    
\textsuperscript{\Letter},    
~\IEEEmembership{Senior Member,~IEEE}%    
\thanks{Wanhui Luo and Qijun Chen are with the Department of Control    
Science \& Engineering, the College of Electronic    
\& Information Engineering, Tongji University,    
Shanghai 201804, China    
(e-mail: luowanhui@tongji.edu.cn, qjchen@tongji.edu.cn).}%    
\thanks{Hongbo Zhao is with the Shanghai Research    
Institute for Intelligent Autonomous Systems,    
Tongji University, Shanghai 201210, China    
(e-mail: hongbozhao@tongji.edu.cn).}%    
\thanks{Yi Zhou is with the School of Artificial Intelligence and Robotics,    
Hunan University, 410082, Changsha, China    
(e-mail: eeyzhou@hnu.edu.cn).}%    
\thanks{Yiming Nie is with the Unmanned Systems    
Technology Research Center, Defense Innovation    
Institute, Academy of Military Science,    
Beijing 100071, China    
(e-mail: nieym@alumni.nudt.edu.cn).}%    
\thanks{Rui Fan is with the Department of Control    
Science \& Engineering, the College of Electronic    
\& Information Engineering, Shanghai Research    
Institute for Intelligent Autonomous Systems,    
the State Key Laboratory of Intelligent Autonomous    
Systems, and Frontiers Science Center for    
Intelligent Autonomous Systems, Tongji University,    
Shanghai 201804, China    
(e-mail: rui.fan@ieee.org).}%    
\thanks{\Letter~Corresponding author: Rui Fan.}%
}

\maketitle
\begin{abstract}
The wide field of view of fisheye cameras and the simplicity of a monocular setup make monocular fisheye visual odometry (VO) attractive for robotic localization and mapping. Nevertheless, scale drift remains a major challenge over long trajectories. Unsupervised depth learning can help to mitigate this issue, but the pronounced distortion of fisheye images poses challenges for accurate depth estimation. This article presents S$^2$D-FVO, a monocular fisheye VO framework integrating spherical bundle adjustment (SBA)-guided unsupervised depth learning to improve scale consistency and depth estimation accuracy. Given that pixel reprojection residuals do not uniformly reflect angular deviations between viewing rays and that the first-order linearization of a scalar angular reprojection residual is insensitive to perturbations along one tangent direction, SBA employs tangent-space angular reprojection residuals derived from the spherical logarithmic map. These residuals retain sensitivity along both tangent directions, with their norms equal to the corresponding scalar angular residuals. A median depth-based scale regularization term is further integrated into SBA to constrain the reconstruction scale. Building on these formulations, SBA replaces the pose estimation network during unsupervised depth learning, providing more reliable relative poses for view synthesis. In turn, depth estimates from the well-trained depth estimation network provide scale references for SBA, improving scale consistency while allowing individual 3-D landmark positions to be refined. Extensive experiments conducted on multiple public datasets, as well as two newly collected large-scale datasets, demonstrate that the proposed framework outperforms representative existing approaches in both depth and trajectory estimation accuracy and is applicable to different camera models. 
\end{abstract}

\begin{figure*}[!tp]
\centering
\includegraphics[width=\textwidth]{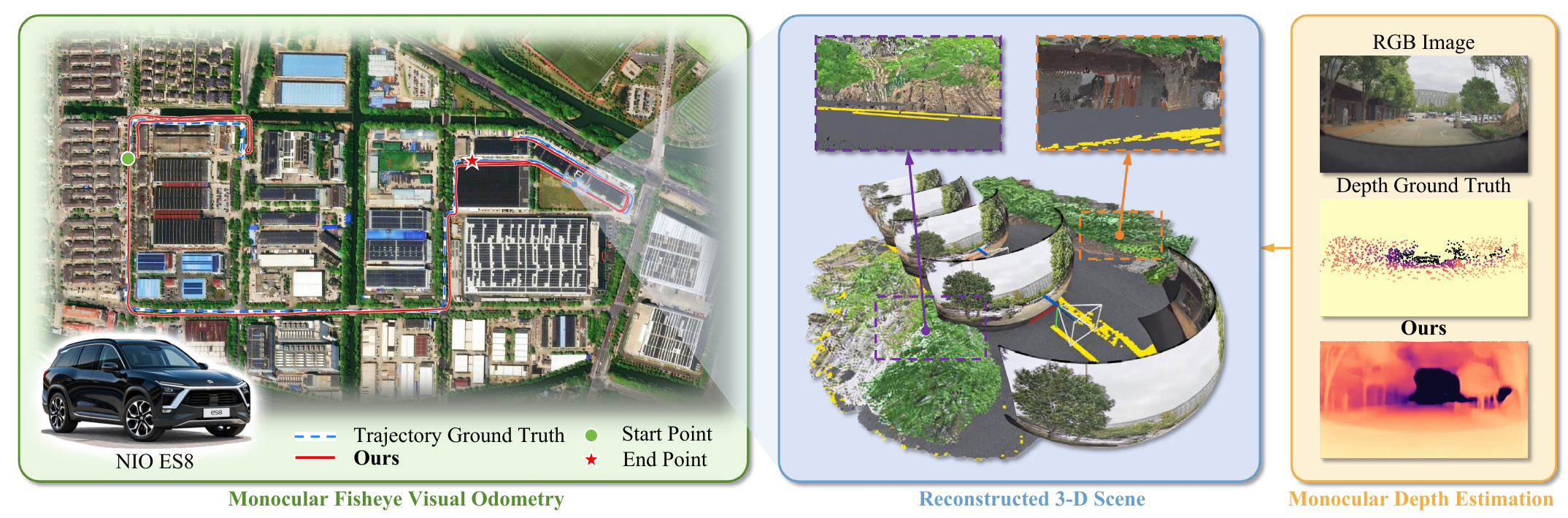}
\caption{Illustration of the proposed S$^2$D-FVO framework for monocular fisheye visual odometry and depth estimation. An NIO ES8 vehicle is used for data collection. The estimated trajectory and depth map are compared with the ground truth, and a reconstructed local 3-D scene is visualized.}
\label{fig:teaser}
\end{figure*}

\section{Introduction}
\label{sec.intro}

\IEEEPARstart{M}{ONOCULAR} fisheye visual odometry (VO) serves as a core component of visual navigation systems for mobile robots~\cite{matsuki2018omnidirectional,boche2025okvis2x}, unmanned aerial vehicles~\cite{forster2017svo,xu2025airslam}, and autonomous vehicles~\cite{wang2018cubemapslam,yin2025iloc}. Benefiting from the wide field of view (FoV), a single fisheye camera preserves greater visual overlap between consecutive frames, enabling more stable camera tracking under substantial inter-frame viewpoint changes and thereby improving the accuracy and robustness of visual odometry~\cite{zhang2016benefit,zhang2023bamf,wang2025lfpgvio}. 
Moreover, compared with using multiple perspective cameras to achieve comparable visual coverage, a single fisheye camera significantly simplifies the VO system by eliminating the need for hardware synchronization and extrinsic calibration~\cite{eckenhoff2019multi,guan2026affine}. Consequently, monocular fisheye VO provides an attractive solution that combines wide visual coverage with a simple sensing configuration.

Similar to conventional VO frameworks based on a monocular perspective camera, monocular fisheye VO frameworks suffer from inherent scale ambiguity, which leaves the magnitude of camera translation and 3-D landmark
coordinates undetermined, posing a fundamental challenge to trajectory estimation, especially over long trajectories~\cite{strasdat2010scale,ju2026scalemaster}. This ambiguity cannot be resolved by bundle adjustment (BA), since jointly scaling the camera translations and 3-D landmark coordinates leaves their image projections unchanged~\cite{hartley2003multiple,strasdat2010scale}. Consequently, as new keyframes are localized and new landmarks are triangulated, the resulting scale deviations can propagate through successive estimates and accumulate into pronounced scale drift, which cannot be fully eliminated through reprojection error minimization alone~\cite{wu2026sceslam}. An additional scale reference is therefore required to achieve scale consistency in visual odometry~\cite{zhong2026dynamic}.

Monocular depth estimation networks (hereafter referred to as \textit{DepthNets}) have been extensively incorporated into monocular VO systems to mitigate scale drift~\cite{lai2025zerovo,wu2026foundationslam,zhong2026dynamic}.
Although recent monocular depth estimation foundation models can produce reliable depth for fisheye images~\cite{guo2025depthanycamera,piccinelli2025unik3d,ganesan2026unidac}, their computational demands can limit their suitability for VO frameworks that typically require real-time performance. Furthermore, training a fisheye DepthNet in a fully supervised manner requires accurate dense ground-truth depth annotations, which are expensive and labor-intensive to acquire~\cite{kumar2020fisheyedistancenet,zhao2024fisheyedepth}.
In contrast, unsupervised depth learning from unlabeled image sequences provides a more practical alternative by minimizing a photometric loss computed through view synthesis using estimated depth maps and relative camera poses~\cite{zhou2017unsupervised,godard2019digging,wu2025geodepth,li2026seeingdepth}. Nevertheless, the photometric loss depends on both estimated depth and relative pose, making DepthNet training susceptible to errors in PoseNet outputs. Moreover, minimizing photometric loss alone does not guarantee accurate relative pose estimates. Therefore, a reliable pose estimation algorithm is required to enable effective unsupervised depth learning on fisheye images.

Monocular depth estimates can be further incorporated into the BA objective to provide additional constraints, enabling the joint optimization of relative camera poses and 3-D landmarks by minimizing both feature reprojection and depth consistency errors~\cite{zhong2026dynamic}. 
Fisheye cameras, however, exhibit severe radial distortion, resulting in the nonuniform projection of viewing rays onto the image plane~\cite{kannala2006generic}. 
Consequently, as the camera moves, the same scene region may be observed from different viewpoints in consecutive frames, causing the corresponding 3-D landmarks to be projected with different degrees of image distortion. This weakens feature descriptor consistency and degrades the quality of visual correspondences~\cite{huang2024_360loc,jung2025edm,liu2025priorflow}. 
Moreover, under the nonlinear fisheye projection model, an identical angular deviation between observed and predicted viewing rays may produce different pixel reprojection residuals across image regions, resulting in nonuniform reprojection error measurements across the FoV and additional nonlinearity in the BA objective~\cite{lee2019angular,tiradogarin2025anycalib,guo2026panovggt}.
To mitigate severe distortion in wide-FoV images, several existing methods remap each image into multiple cubemap faces or local tangent-plane patches~\cite{shen2026panogabor}. Nevertheless, such piecewise image remapping strategies may introduce discontinuities in depth estimates across the boundaries between adjacent faces or patches, potentially reducing their reliability as additional constraints in BA~\cite{shen2026panogabor}.
Therefore, developing a continuity-preserving image remapping strategy together with an appropriate reprojection residual formulation warrants further investigation to enable robust explicit pose estimation for scale-consistent monocular fisheye visual odometry.

To address the aforementioned limitations, this article proposes S$^2$D-FVO, a scale-consistent monocular fisheye visual odometry framework that employs spherical bundle adjustment to guide unsupervised depth learning and leverages the DepthNet estimates as a scale reference for visual odometry.
First, rather than remapping raw fisheye images into multiple cubemap faces or local tangent-plane patches, the proposed S$^2$D-FVO processes them in the equirectangular projection (ERP) domain~\cite{guo2025depthanycamera}, allowing images captured by fisheye cameras calibrated based on different projection models or having different FoVs to be represented in a common domain. Such an image remapping strategy better preserves spatial continuity across the FoV, and the resulting images are used as inputs for both camera tracking and monocular depth estimation. 
Additionally, bundle adjustment is formulated directly on the unit sphere using the spherical logarithmic map to construct 2-D tangent-space angular reprojection residuals. The discrepancy between predicted and observed unit viewing rays is quantified by their angular deviation rather than the pixel displacement, ensuring that identical angular deviations produce exactly the same residual magnitude irrespective of the feature location in the image. 
When incorporating monocular depth estimates into bundle adjustment, the proposed framework does not enforce pointwise consistency between individual landmark depth values and their corresponding DepthNet outputs. Instead, only their median values are enforced to be consistent, thereby anchoring the overall reconstruction scale while allowing individual landmark depth to be further refined during optimization. 
The spherical BA formulation is then applied to both unsupervised depth learning and monocular fisheye visual odometry. During unsupervised depth learning, it replaces the PoseNet and jointly refines relative poses and corresponding 3-D landmarks. The refined poses and current depth estimates are subsequently used for view synthesis, from which a photometric loss is computed and minimized to train the DepthNet. 
This process alternates between pose refinement and DepthNet optimization throughout training. During monocular fisheye visual odometry, the learned depth estimates serve as a scale reference within spherical bundle adjustment, where the angular reprojection and depth consistency errors jointly constrain the estimated poses and 3-D landmarks, thereby improving scale consistency over long trajectories. 
Extensive experiments on multiple public fisheye datasets, as well as the newly collected real-world and synthetic datasets, demonstrate the effectiveness of using relative poses obtained through spherical bundle adjustment to improve unsupervised fisheye depth learning and incorporating depth estimates as scale references for visual odometry, with representative results illustrated in Fig.~\ref{fig:teaser}. The proposed framework outperforms state-of-the-art (SoTA) approaches across the evaluated datasets in terms of both depth estimation and visual odometry, while maintaining continuous camera tracking after pose initialization.  

The main contributions of this study are summarized as follows:
\begin{itemize}
\item A scale-consistent monocular fisheye visual odometry framework that leverages learned monocular depth estimates as a scale reference.
\item A spherical BA formulation that combines tangent-space angular reprojection residuals with a median depth-based scale regularization term, thereby enabling robust pose estimation for scale-consistent monocular fisheye VO.
\item A spherical BA-guided unsupervised depth learning strategy that replaces conventional PoseNets with a robust explicit pose estimation algorithm, providing more reliable photometric supervision and enabling
iterative refinement of depth and pose throughout training.
\item Extensive evaluations on multiple public fisheye datasets and newly collected real-world and synthetically rendered fisheye datasets, demonstrating consistent improvements in both depth estimation and trajectory accuracy over SoTA approaches. 
\end{itemize}

The rest of this article is organized as follows:
Sect.~\ref{sec.Relatedwork} reviews SoTA monocular depth estimation and VO approaches.
Sect.~\ref{sec.Method} presents the proposed VO framework.
Sect.~\ref{sec.Experiments} reports extensive experimental results and ablation studies.
Sect.~\ref{sec.Discussion} discusses the applicability of the proposed framework to different camera models and its limitations.
Finally, Sect.~\ref{sec.Conclusion} concludes this article and discusses
possible future work.

\section{Related Work}
\label{sec.Relatedwork}

\subsection{Fisheye Image Processing}
Fisheye cameras provide a wide FoV through a nonlinear mapping of viewing rays onto the image plane, resulting in spatially varying distortion across the FoV~\cite{athwale2025darswinunet,gangopadhyay2025extending}. For downstream vision tasks that use fisheye images as inputs, such distortion must be properly handled, as it can degrade the performance of downstream vision algorithms. Common remapping approaches transform fisheye images using alternative projection models. As a representative remapping strategy, fisheye images can be rectified using perspective projection, allowing methods developed for perspective cameras to be applied directly~\cite{schneider2016accuracy}. Since a single perspective image typically covers only part of the original FoV, piecewise remapping can retain broader visual coverage by projecting viewing rays onto the faces of a cube~\cite{wang2018cubemapslam,wang2020bifuse} or local tangent-plane patches~\cite{eder2020tangent,li2022omnifusion,cao2025panda}, thereby reducing distortion within each projected region. Nevertheless, such piecewise remapping introduces discontinuities between adjacent views, which may disrupt spatial consistency, particularly in dense prediction tasks such as depth estimation, while also requiring additional cross-view processing~\cite{wang2018cubemapslam,wang2020bifuse,li2022omnifusion}. In contrast, equirectangular, Cassini, and other cylindrical projections enable fisheye images to be remapped into a single image, making them suitable for tasks that benefit from processing the full FoV within a unified representation. These projections differ primarily in the manner in which the angular coordinates of viewing rays are mapped to image coordinates~\cite{meuleman2021real,li2022mode,yun2023egformer,jung2025edm}. 
For learning-based vision tasks, a variety of methods directly process fisheye images and adapt feature extraction to fisheye distortion by adjusting convolutional sampling locations~\cite{ahmad2022fisheyehdk,tateno2018distortion} or introducing calibration tokens that modulate the latent embeddings of fisheye images~\cite{gangopadhyay2025extending}. However, fisheye cameras can differ substantially in their FoVs and distortion characteristics, making it challenging for these methods to generalize consistently across different fisheye camera models~\cite{ahmad2022fisheyehdk,gangopadhyay2025extending}. Therefore, no single fisheye image processing strategy is universally suitable for all vision tasks. In this study, preserving a wide FoV within a single image is important for both dense depth estimation and sparse feature tracking. Accordingly, equirectangular projection is adopted to provide a unified spherical representation with a direct mapping between image coordinates and viewing rays, thereby preserving the wide FoV and facilitating the application of the proposed framework to different fisheye cameras.

\subsection{Monocular Fisheye Depth Estimation}
Existing monocular fisheye depth estimation approaches can be categorized into fully supervised and unsupervised methods~\cite{lee2023slabins,kumar2020fisheyedistancenet,kumar2020unrectdepthnet,kumar2021syndistnet}.
The former require large-scale dense depth annotations, which remain scarce, particularly for real fisheye images~\cite{zhao2024fisheyedepth,indyk2026widedepth}, making it difficult to train depth estimation models with extensive supervised data. This limitation motivates the reuse of monocular depth estimation foundation models pretrained on perspective images. However, their direct application to fisheye images remains challenging due to the substantial differences in projection geometry and distortion characteristics between perspective and fisheye images~\cite{gangopadhyay2025extending}. To address this issue, existing studies~\cite{gangopadhyay2025extending,guo2025depthanycamera,ganesan2026unidac} mainly attempt to adapt pretrained monocular depth estimation models to fisheye images through different strategies. 
For instance, DAC~\cite{guo2025depthanycamera} and UniDAC~\cite{ganesan2026unidac} remap input images according to a common projection model during both training and inference, enabling a single network to process images acquired by different cameras. Other approaches instead retain fisheye images as inputs and incorporate camera projection information into the DepthNet. For example, UniK3D~\cite{piccinelli2025unik3d} jointly predicts the viewing ray and radial distance corresponding to each image pixel. The study~\cite{gangopadhyay2025extending} adopts a different strategy by introducing calibration tokens that adapt monocular depth estimation foundation models pretrained on perspective images to fisheye inputs by aligning fisheye features with the feature space learned from perspective images.

On the other hand, unsupervised monocular fisheye depth estimation approaches avoid the need for dense depth annotations by enforcing photometric consistency between a target image and its synthesized counterpart generated from an adjacent source image~\cite{kumar2020fisheyedistancenet,kumar2020unrectdepthnet}. Subsequent studies~\cite{kumar2020fisheyedistancenet,kumar2021syndistnet} have explored different strategies to mitigate violations of the photometric consistency assumption in unsupervised fisheye depth learning. For example, FisheyeDistanceNet~\cite{kumar2020fisheyedistancenet} introduces more robust photometric losses, while SynDistNet~\cite{kumar2021syndistnet} leverages semantic information to mask dynamic objects. Although these methods improve the reliability of photometric consistency enforcement, the quality of view synthesis still depends heavily on the accuracy of the relative pose between the target and source images. To address this issue, subsequent studies~\cite{xu2026distortion,zhao2024fisheyedepth} have explored alternative strategies for exploiting temporal information. For instance, FisheyeDepth~\cite{zhao2024fisheyedepth} utilizes ground-truth relative poses for view synthesis, eliminating the need for a PoseNet. Nonetheless, this strategy is infeasible when ground-truth relative poses are unavailable. More recently, DDS-Net~\cite{xu2026distortion} constructs a cost volume by matching features extracted from adjacent fisheye frames and iteratively refines the estimated depth. Although these studies improve the exploitation of temporal information for unsupervised fisheye depth learning, the multi-view constraints coupling camera motion and scene structure remain underexplored. To address this limitation, bundle adjustment is employed in this study to jointly refine the camera poses and 3-D landmarks across multiple frames, thereby more fully exploiting multi-view geometry while providing refined relative poses for view synthesis.

\begin{figure*}[!tp]
    \centering
    \includegraphics[width=\textwidth]{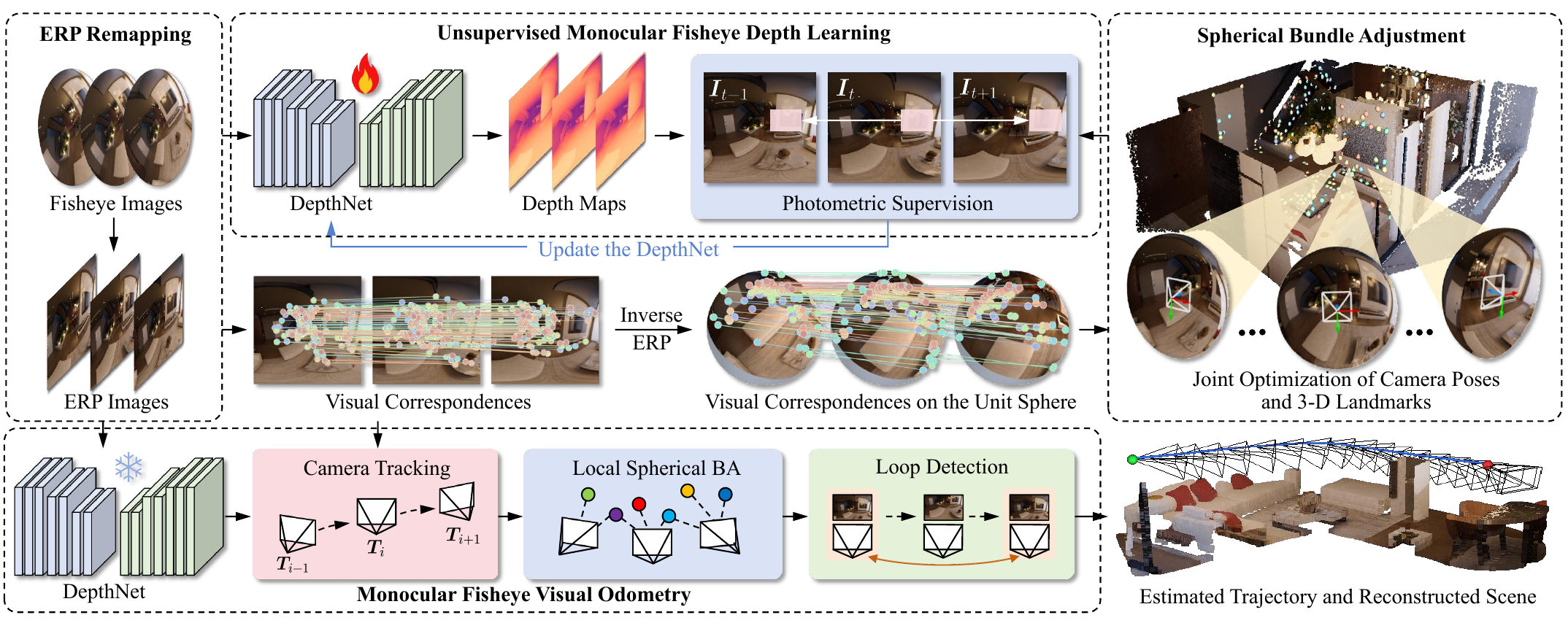}
    \caption{An overview of the proposed S$^2$D-FVO framework. Fisheye images are first remapped to ERP images for both camera tracking and unsupervised depth learning. Meanwhile, visual correspondences are unprojected onto the unit sphere for spherical bundle adjustment. Subsequently, the refined camera poses are incorporated into the unsupervised depth learning process, and the well-trained DepthNet is used to provide a scale reference for monocular fisheye visual odometry.}
    \label{fig:overview}
\end{figure*}

\subsection{Monocular Visual Odometry}
Classical monocular visual odometry approaches estimate camera poses and 3-D landmarks primarily by minimizing either feature reprojection errors~\cite{mur2015orbslam,wang2018cubemapslam,sumikura2019openvslam} or photometric errors~\cite{engel2018dso,matsuki2018omnidirectional}. With the growing use of wide-FoV cameras, these methods have been extended to accommodate projection models beyond the perspective camera model. For example, OpenVSLAM~\cite{sumikura2019openvslam}, ORB-SLAM3~\cite{campos2021orbslam3}, and OV$^{2}$SLAM~\cite{fer2021ov2slam} extend their pipelines to fisheye cameras by formulating projection and reprojection functions according to specific fisheye camera models, while Omnidirectional DSO~\cite{matsuki2018omnidirectional} incorporates the unified omnidirectional camera model into DSO~\cite{engel2018dso}. SVO~\cite{forster2017svo} and HSO~\cite{luo2022hso} further extend hybrid visual odometry to wide-FoV cameras. Although these methods enable direct processing of fisheye images, their performance may degrade under severe radial distortion. Alternatively, CubemapSLAM~\cite{wang2018cubemapslam} reprojects each fisheye image onto the faces of a cube to generate multiple perspective images, alleviating distortion effects and enabling more reliable feature extraction and matching. More recently, the study~\cite{zhang2026p2uslam} models uncertainty in camera poses and 3-D landmarks during bundle adjustment to improve the accuracy and robustness of pose estimation with wide-FoV cameras. Despite these advances, existing monocular fisheye VO methods still suffer from scale drift due to the inherent scale ambiguity.

To mitigate the scale drift problem, monocular depth estimates have been introduced into monocular VO as an additional scale reference~\cite{zhang2022towards,wang2025kpdepthvo,wu2026sceslam}. One common strategy is to use monocular depth estimates for scale recovery. For example, KPDepthVO~\cite{wang2025kpdepthvo} introduces an uncertainty-aware scale recovery strategy that incorporates uncertainty in monocular depth estimates during scale alignment. Nevertheless, monocular depth estimates may exhibit spatially varying scale inconsistencies that cannot be fully compensated by a single scale factor. To address this issue, HI-SLAM2~\cite{zhang2025hislam2} introduces a grid-based scale alignment strategy rather than assuming a uniform scale factor across the entire image. While such scale alignment strategies establish the relationship between the depth estimates predicted by the network and those of reconstructed 3-D landmarks, SCE-SLAM~\cite{wu2026sceslam} further introduces scene coordinate bundle adjustment that anchors the reconstruction to a reference scale through explicit 3-D coordinate constraints. Nevertheless, these methods are primarily developed for perspective cameras. For fisheye cameras, monocular depth estimates have mainly been used for scale recovery and dense mapping~\cite{gallagher2023densefisheye}, without directly constraining the reconstruction scale during joint optimization of camera poses and 3-D landmarks. To address this limitation, this study incorporates monocular fisheye depth estimates into bundle adjustment as a scale reference, directly constraining the reconstruction scale during joint optimization rather than performing separate scale recovery, thereby improving the scale consistency of monocular fisheye visual odometry.

\section{Methodology}
\label{sec.Method}

An overview of the proposed S$^2$D-FVO framework is illustrated in Fig.~\ref{fig:overview}. Sect.~\ref{subsec.preliminaries} first introduces the relevant preliminaries, based on which Sect.~\ref{subsec.Spherical_BA} formulates spherical bundle adjustment as the common optimization basis for both monocular fisheye depth learning and visual odometry. Building upon this formulation, Sect.~\ref{subsec.depth_learning} presents the unsupervised monocular fisheye depth learning strategy, where spherical bundle adjustment replaces the PoseNet to refine the relative poses used for view synthesis during DepthNet training. Sect.~\ref{subsec.visual_odometry} subsequently presents the monocular fisheye visual odometry system, in which the well-trained DepthNet provides a scale reference for spherical bundle adjustment, thereby improving the long-term scale consistency of the estimated trajectory.

\subsection{Preliminaries}
\label{subsec.preliminaries}

Let $\boldsymbol{p}_j^W\in\mathbb{R}^3$ denote the coordinates of the $j$-th 3-D landmark in the world coordinate system, and let $\boldsymbol{p}_{ij}^C\in\mathbb R^3$ denote its coordinates in the camera coordinate system associated with the $i$-th video frame $\boldsymbol I_i\in\mathbb R^{H\times W\times 3}$, where $H$ and $W$ denote the image height and width, respectively. The two coordinates can be related using the following expression:
\begin{equation}
\begin{pmatrix}
\boldsymbol{p}_{ij}^C\\1
\end{pmatrix}=
\underbrace{
\begin{pmatrix}
\boldsymbol R_i & \boldsymbol t_i\\
\boldsymbol 0^\top & 1
\end{pmatrix}
}_{\boldsymbol{T}_i\in\mathrm{SE}(3)}
\begin{pmatrix}
\boldsymbol{p}_j^W\\1
\end{pmatrix},
\end{equation}
where $\boldsymbol T_{i}$ denotes the rigid transformation matrix from the world coordinate system to the camera coordinate system, $\boldsymbol R_{i}\in\mathrm{SO}(3)$ and $\boldsymbol t_{i}\in\mathbb R^3$ represent the rotation matrix and translation vector, respectively, and $\boldsymbol{0}$ is a column vector of zeros. $\boldsymbol{p}_{ij}^C$ can be projected onto a pixel $\boldsymbol{p}_{ij}\in\mathbb{R}^2$ in the $i$-th video frame through $\boldsymbol{p}_{ij}=\pi_f(\boldsymbol{p}_{ij}^C;\boldsymbol\kappa)$, where $\pi_f(\cdot;\cdot)$ represents the fisheye projection function and $\boldsymbol{\kappa}$ is a vector containing the camera calibration parameters. The corresponding bearing vector (also known as a unit viewing ray) $\boldsymbol b_{ij}\in\mathbb{S}^2$ can be obtained through inverse projection. Let $d_{ij}>0$ denote the radial depth of the 3-D landmark with respect to the camera center. Its coordinates in the camera coordinate system can then be expressed as
$\boldsymbol{p}_{ij}^C=d_{ij}\boldsymbol{b}_{ij}$.

To enable unified processing across fisheye cameras with different projection models or FoVs, the original fisheye images are remapped using
equirectangular projection~\cite{guo2025depthanycamera} in this study. A unit viewing ray $\boldsymbol{b}\in\mathbb{S}^2$ can be parameterized by spherical longitude
$\lambda\in[-\pi,\pi)$ and latitude $\varphi\in[-\pi/2,\pi/2]$ as
$\boldsymbol b=
(\cos\varphi\sin\lambda,\,
\sin\varphi,\,
\cos\varphi\cos\lambda)^\top$, 
which is related to a pixel in the ERP image using the following projection function:
\begin{equation}
\pi_{ERP}(\boldsymbol b)
=
\begin{pmatrix}
\dfrac{W}{2\pi}(\lambda+\pi)\\[2mm]
\dfrac{H}{\pi}
\left(
\dfrac{\pi}{2}-\varphi
\right)
\end{pmatrix}.
\label{eq:erp_projection}
\end{equation}
Under this parameterization, each fisheye image is remapped to the region of the full ERP domain corresponding to its FoV.

Since $\boldsymbol{b}$ lies on the unit sphere, the tangent space at $\boldsymbol b\in\mathbb S^2$ is defined as
$T_{\boldsymbol{b}}\mathbb{S}^2
=
\{\boldsymbol{v}\in\mathbb{R}^3
\mid
\boldsymbol{b}^\top\boldsymbol{v}=0\}$, where $\boldsymbol{v}$ denotes an arbitrary tangent vector. The exponential map
$\operatorname{Exp}_{\boldsymbol b}:T_{\boldsymbol b}\mathbb S^2
\rightarrow\mathbb S^2$
is defined as follows:
\begin{equation}
\operatorname{Exp}_{\boldsymbol b}(\boldsymbol v)
=
\begin{cases}
\cos\!\left(\|\boldsymbol v\|_2\right)\boldsymbol b
+
\dfrac{\sin\!\left(\|\boldsymbol v\|_2\right)}
{\|\boldsymbol v\|_2}\boldsymbol v,
& \boldsymbol v\neq\boldsymbol 0,
\\[0.6em]
\boldsymbol b,
& \boldsymbol v=\boldsymbol 0.
\end{cases}
\label{eq:exp_sphere}
\end{equation}
Given another unit viewing ray $\boldsymbol{a}\in\mathbb S^2\setminus\{-\boldsymbol{b}\}$, the angle between $\boldsymbol{a}$ and $\boldsymbol{b}$ is given by
$\phi=\arccos(\boldsymbol{a}^\top\boldsymbol{b})$, and the logarithmic map
$\operatorname{Log}_{\boldsymbol{b}}:\mathbb S^2\setminus\{-\boldsymbol{b}\}
\rightarrow T_{\boldsymbol{b}}\mathbb S^2$
is defined as follows:
\begin{equation}
\operatorname{Log}_{\boldsymbol{b}}(\boldsymbol{a})
=
\begin{cases}
\dfrac{\phi}{\sin\phi}
\left(
\boldsymbol{a}
-
(\boldsymbol{b}^\top\boldsymbol{a})\boldsymbol{b}
\right),
& 0<\phi<\pi,
\\[0.6em]
\boldsymbol 0,
& \phi=0.
\end{cases}
\label{eq:log_sphere}
\end{equation}
\eqref{eq:exp_sphere} and \eqref{eq:log_sphere} establish the relationship between points on the unit sphere and their corresponding local tangent spaces, which will be used in the following subsection.

\subsection{Spherical Bundle Adjustment}
\label{subsec.Spherical_BA}

Bundle adjustment jointly refines camera poses and 3-D landmarks by minimizing reprojection errors of all observations. For an observed pixel $\boldsymbol{p}_{ij}$, the pixel reprojection residual is typically defined as follows:
\begin{equation}
\boldsymbol{e}_{ij}
=
\boldsymbol{p}_{ij}
-
\pi_f
\left(
\hat{\boldsymbol R}_i\hat{\boldsymbol p}_j^W
+
\hat{\boldsymbol t}_i;
\boldsymbol\kappa
\right),
\end{equation}
where $\hat{\boldsymbol R}_i$, $\hat{\boldsymbol t}_i$, and $\hat{\boldsymbol p}_j^W$ denote the current estimates of the camera rotation, camera translation, and world coordinates of the 3-D landmark, respectively. $\|\boldsymbol{e}_{ij}\|_2$, the magnitude of $\boldsymbol{e}_{ij}$, is defined as the 
pixel reprojection error~\cite{triggs2000bundle}.
However, pairs of viewing rays with the same angular deviation may correspond to different pixel distances on the image plane, which is an inherent limitation of bundle adjustment based on pixel reprojection errors, particularly for fisheye cameras. Therefore, it is natural to introduce a new reprojection error formulated directly based on angular deviations.

Since all 3-D landmarks on the same viewing ray are projected onto the same pixel in the image regardless of depth~\cite{ramalingam2010generic}, the projection function can be formulated directly in terms of unit viewing rays. Let $\mathcal{U}\subset\mathbb{S}^{2}$ be a nonempty open set of valid unit viewing rays, and let 
$\Pi:\mathcal{U}\rightarrow\Theta\subset\mathbb{R}^{2}$
denote a smooth projection function for a given camera model, satisfying $\operatorname{rank}D\Pi(\boldsymbol{b})=2$ for all $\boldsymbol{b}\in\mathcal{U}$. Let
$d_{\mathbb{S}^{2}}:\mathbb{S}^2\times\mathbb{S}^2\rightarrow[0,\pi]$
denote the geodesic distance on $\mathbb{S}^2$, \textit{i.e.}, the angular deviation between two points on the unit sphere.

\begin{proposition}
\label{prop:nonuniform_pixel_reprojection}
There exist 
$\boldsymbol b_1,\boldsymbol b_1',
\boldsymbol b_2,\boldsymbol b_2'\in\mathcal U$
such that
$d_{\mathbb S^{2}}(\boldsymbol b_1,\boldsymbol b_1')
=
d_{\mathbb S^{2}}(\boldsymbol b_2,\boldsymbol b_2')$, 
whereas
$\|\Pi(\boldsymbol b_1)-\Pi(\boldsymbol b_1')\|_2
\neq
\|\Pi(\boldsymbol b_2)-\Pi(\boldsymbol b_2')\|_2$.
Therefore, equal angular deviations between viewing rays can correspond to different pixel displacements under the projection $\Pi$.
\end{proposition}

\begin{IEEEproof}
Suppose, for contradiction, that
Proposition~\ref{prop:nonuniform_pixel_reprojection} does not hold. For any $\boldsymbol{b}_0\in\mathcal{U}$, since $\mathcal{U}$ is open,
there exists an open geodesic ball $\mathcal{B}$ centered at
$\boldsymbol{b}_0$ with radius $0<\delta<\frac{\pi}{2}$  such that
$
\mathcal{B}
=
\left\{
\boldsymbol{b}\in\mathbb S^2
\mid
d_{\mathbb{S}^2}(\boldsymbol{b},\boldsymbol{b}_0)<\delta
\right\}
\subset\mathcal{U}.
$ Then, for all
$\boldsymbol{b}_1,\boldsymbol{b}_2\in\mathcal{B}$, the Euclidean distance between their projections depends only on $d_{\mathbb S^2}(\boldsymbol{b}_1,\boldsymbol{b}_2)$. Therefore, there exists a function
$h:[0, 2\delta)\rightarrow\mathbb R_{\geq0}$ such that
\begin{equation}
\left\|
\Pi(\boldsymbol b_1)-\Pi(\boldsymbol b_2)
\right\|_2
=
h\!\left(
d_{\mathbb S^{2}}
(\boldsymbol b_1,\boldsymbol b_2)
\right).
\label{eq:uniform_pixel_distance}
\end{equation}

For any $\boldsymbol{b}\in\mathcal{B}$ and any unit tangent vector
$\boldsymbol{u}\in T_{\boldsymbol{b}}\mathbb{S}^{2}$, define
$\boldsymbol{b}_{\epsilon}
=
\operatorname{Exp}_{\boldsymbol b}(\epsilon\boldsymbol u)$,
where $0<\epsilon<\delta-d_{\mathbb{S}^2}(\boldsymbol b,\boldsymbol b_0)$.
Since
$d_{\mathbb S^{2}}(\boldsymbol b,\boldsymbol b_{\epsilon})=\epsilon$,
~\eqref{eq:uniform_pixel_distance} implies that
$\|\Pi(\boldsymbol b_{\epsilon})-\Pi(\boldsymbol b)\|_{2}=h(\epsilon)$.
Dividing both sides by $\epsilon$ and taking the limit as
$\epsilon\rightarrow0^+$ yields the following expression:

\begin{equation}
\left\|
D\Pi(\boldsymbol b)[\boldsymbol u]
\right\|_{2}
=
\lim_{\epsilon\rightarrow0^{+}}
\frac{h(\epsilon)}{\epsilon}
=:
s,
\label{eq:uniform_local_scale}
\end{equation}
where $D\Pi(\boldsymbol{b})[\boldsymbol{u}]$ denotes the differential of $\Pi$ at $\boldsymbol{b}$ applied to $\boldsymbol{u}$. The scalar $s$ is independent of both $\boldsymbol b$ and $\boldsymbol u$, and $s>0$ follows from the full-rank assumption on $D\Pi(\cdot)$.

By the linearity of the differential and the absolute homogeneity of
the Euclidean norm~\cite{axler2015linear}, \eqref{eq:uniform_local_scale} extends to every tangent vector
$\boldsymbol{v}\in T_{\boldsymbol{b}}\mathbb S^2$ as
$\|D\Pi({\boldsymbol b})[\boldsymbol v]\|_2
=
s\|\boldsymbol v\|_2$.
Define the rescaled projection function
$\widetilde{\Pi}:=\frac{1}{s}\Pi$.
It follows that
$\|D\widetilde{\Pi}({\boldsymbol b})[\boldsymbol v]\|_2
=
\|\boldsymbol v  \|_2$.
Hence, $\widetilde{\Pi}|_{\mathcal{B}}$ preserves the Riemannian metric and is a local isometry between
$\mathcal{B}$ and
$\widetilde{\Pi}(\mathcal{B})\subset\mathbb{R}^{2}$.
By Gauss's Theorema Egregium, Gaussian curvature is invariant under local isometries~\cite{docarmo2016differential}.
However, $\mathcal{B}$ inherits the nonzero Gaussian curvature of $\mathbb{S}^2$, whereas $\widetilde{\Pi}(\mathcal{B})$, being an open subset of the Euclidean plane, has zero Gaussian curvature. Therefore, $\widetilde{\Pi}|_{\mathcal{B}}$ cannot be a local isometry, yielding
a contradiction to~\eqref{eq:uniform_pixel_distance}.
\end{IEEEproof}

To avoid the spatial nonuniformity of pixel reprojection errors, this study directly measures the angular deviation between the observed and predicted viewing rays on the unit sphere. 
Given the current estimates of the camera pose and 3-D landmark, the predicted unit viewing ray $\hat{\boldsymbol{b}}_{ij}$ is computed using the following expression:
\begin{equation}
\hat{\boldsymbol b}_{ij}
=
\frac{\hat{\boldsymbol R}_{i}\hat{\boldsymbol p}_{j}^{W}+\hat{\boldsymbol t}_{i}}
{\left\|\hat{\boldsymbol R}_{i}\hat{\boldsymbol p}_{j}^{W}+\hat{\boldsymbol t}_{i}\right\|_{2}},
\label{eq:predicted_bearing}
\end{equation}
For a fixed observed unit viewing ray $\boldsymbol{b}_{ij}$, the scalar angular reprojection residual function is defined as follows:
\begin{equation}
\mathcal {A}_{ij}(\cdot)
=
d_{\mathbb S^2}(\boldsymbol {b}_{ij} , \cdot)
\label{eq:angular_residual_function}
\end{equation}
The corresponding scalar angular reprojection residual $\alpha_{ij}$ is then defined as follows:
\begin{equation}
\alpha_{ij}
=
\mathcal A_{ij}(\hat{\boldsymbol b}_{ij})
=
\arccos
(
\boldsymbol {b}_{ij}^{\top}
\hat{\boldsymbol b}_{ij}
).
\label{eq:angular_reprojection_error}
\end{equation}
Although this residual directly quantifies the angular deviation between the observed and predicted viewing rays, incorporating it into nonlinear least-squares optimization typically involves a local first-order approximation obtained through linearization~\cite{madsen2004lm,triggs2000bundle}. The linearization of the scalar angular reprojection residual in~\eqref{eq:angular_reprojection_error} is necessarily insensitive to perturbations along one direction in the tangent space
$T_{\hat{\boldsymbol{b}}_{ij}}\mathbb{S}^2$, as formalized in the following proposition.

\begin{proposition}
\label{prop:scalar_angle_linearization}
For
$\boldsymbol{b}_{ij},\hat{\boldsymbol{b}}_{ij}\in\mathbb{S}^2$ satisfying
$\hat{\boldsymbol{b}}_{ij}\neq\pm\boldsymbol{b}_{ij}$,
the differential 
$
D\mathcal{A}_{ij}(\hat{\boldsymbol{b}}_{ij})
:
T_{\hat{\boldsymbol{b}}_{ij}}\mathbb{S}^2
\rightarrow
\mathbb R
$, \textit{i.e.}, the linear map giving the first-order variation of $\mathcal{A}_{ij}(\cdot)$ at $\hat{\boldsymbol{b}}_{ij}$, has a 1-D kernel, namely
\[
\dim
\ker
D\mathcal{A}_{ij}(\hat{\boldsymbol b}_{ij})
=
1.
\]
Therefore, there exists a nonzero tangent vector along which perturbing the predicted viewing ray produces no first-order change in the scalar angular reprojection residual, making its linearization insensitive to a 1-D subspace of the local tangent space at the predicted viewing ray.
\end{proposition}

\begin{IEEEproof}
Let
$\boldsymbol{\tau}_{ij}^{\parallel}
\in T_{\hat{\boldsymbol {b}}_{ij}}\mathbb S^2$ denote the unit tangent vector along the great-circle geodesic from
$\hat{\boldsymbol {b}}_{ij}$ toward $\boldsymbol {b}_{ij}$.
$\boldsymbol{\tau}_{ij}^{\parallel}$ can be expressed as follows:
\begin{equation}
\begin{aligned}
\boldsymbol\tau_{ij}^{\parallel}
&=
\frac{
(\hat{\boldsymbol{b}}_{ij}\times\boldsymbol{b}_{ij})
\times\hat{\boldsymbol{b}}_{ij}
}{
\|(\hat{\boldsymbol{b}}_{ij}\times\boldsymbol{b}_{ij})
\times\hat{\boldsymbol{b}}_{ij}\|_2
}
=
\frac{
\boldsymbol{b}_{ij}
-
\cos\alpha_{ij}\hat{\boldsymbol{b}}_{ij}
}{
\sin\alpha_{ij}
}.
\end{aligned}
\label{eq:tangent_direction_q}
\end{equation}
For any
$\boldsymbol{v}
\in T_{\hat{\boldsymbol{b}}_{ij}}\mathbb S^2$,
the differential of the angular reprojection residual function at
$\hat{\boldsymbol{b}}_{ij}$ applied to $\boldsymbol{v}$ satisfies
$
D\mathcal A_{ij}(\hat{\boldsymbol b}_{ij})
[\boldsymbol v]
=
-\frac{\boldsymbol b_{ij}^{\top}\boldsymbol v}
{\sin\alpha_{ij}}
=
-(\boldsymbol\tau_{ij}^{\parallel})^\top
\boldsymbol v.
$
In particular,
$
D\mathcal{A}_{ij}(\hat{\boldsymbol{b}}_{ij})
[\boldsymbol{\tau}_{ij}^{\parallel}]
=
-1.
$
Therefore,
$
D\mathcal{A}_{ij}(\hat{\boldsymbol {b}}_{ij})
:
T_{\hat{\boldsymbol{b}}_{ij}}\mathbb S^2
\rightarrow
\mathbb{R}
$
is a nonzero linear map, implying that
$
\dim \operatorname{Im}
D\mathcal{A}_{ij}(\hat{\boldsymbol{b}}_{ij})
=
1.
$
By the rank--nullity theorem~\cite{axler2015linear}, the dimension of the kernel of $D\mathcal{A}_{ij}(\hat{\boldsymbol b}_{ij})$ can be calculated as follows:
\begin{equation}
\begin{aligned}
\dim
\ker D\mathcal{A}_{ij}(\hat{\boldsymbol{b}}_{ij})
=
\underbrace{
\vphantom{
\dim \operatorname{Im}
D\mathcal{A}_{ij}(\hat{\boldsymbol{b}}_{ij})
}
\dim {T}_{\hat{\boldsymbol{b}}_{ij}}\mathbb S^2
}_{2}
-
\underbrace{
\vphantom{
\dim{T}_{\hat{\boldsymbol{b}}_{ij}}\mathbb S^2
}
\dim \operatorname{Im}
D\mathcal{A}_{ij}(\hat{\boldsymbol{b}}_{ij})
}_{1}
=
1.
\end{aligned}
\label{eq:angular_kernel_dimension}
\end{equation}

Let
$\boldsymbol{\tau}_{ij}^{\perp}
\in{T}_{\hat{\boldsymbol{b}}_{ij}}\mathbb S^2$
denote the unit tangent vector orthogonal to
$\boldsymbol{\tau}_{ij}^{\parallel}$. $\boldsymbol{\tau}_{ij}^{\perp}$ can be expressed as follows:
\begin{equation}
\boldsymbol{\tau}_{ij}^{\perp}
=
\frac{
\hat{\boldsymbol{b}}_{ij}\times\boldsymbol {b}_{ij}
}{
\left\|
\hat{\boldsymbol{b}}_{ij}\times\boldsymbol {b}_{ij}
\right\|_2
}.
\end{equation}
For any 
$\boldsymbol{v}'
=
\beta \boldsymbol{\tau}_{ij}^{\perp}$,
where $\beta \in \mathbb {R}$ is an arbitrary scalar,
the differential of the angular reprojection residual function at $\hat{\boldsymbol{b}}_{ij}$ applied to $\boldsymbol{v}'$ satisfies
$
D\mathcal{A}_{ij}(\hat{\boldsymbol {b}}_{ij})
[\boldsymbol{v}']
=
-\beta
(\boldsymbol{\tau}_{ij}^{\parallel})^\top
\boldsymbol{\tau}_{ij}^{\perp}
=
0.
$
Therefore,
$
\operatorname{span}
\left\{
\boldsymbol{\tau}_{ij}^{\perp}
\right\}
\subseteq
\ker
D\mathcal{A}_{ij}(\hat{\boldsymbol{b}}_{ij}).
$
Since both subspaces are 1-D, the following expression can be obtained:
\begin{equation}
\ker D\mathcal{A}_{ij}(\hat{\boldsymbol{b}}_{ij})
=
\operatorname{span}
\left\{
\boldsymbol{\tau}_{ij}^{\perp}
\right\}.
\label{eq:angular_kernel}
\end{equation}

Therefore, perturbations of the predicted viewing ray along
$\boldsymbol{\tau}_{ij}^{\perp}$
produce no first-order variation in the scalar angular reprojection residual, thereby proving Proposition~\ref{prop:scalar_angle_linearization}.

\end{IEEEproof}

\begin{figure}[!tp]
\centering
\includegraphics[width=0.65\columnwidth]{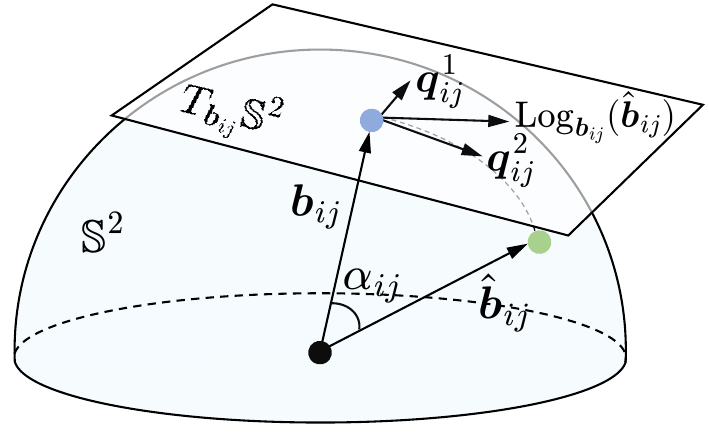}
\caption{An illustration of the tangent-space angular reprojection residual. The angle between the observed viewing ray $\boldsymbol{b}_{ij}$ and the predicted viewing ray $ \hat{\boldsymbol{b}}_{ij}$ is denoted by $\alpha_{ij}$. The predicted viewing ray is then mapped to the tangent space
$T_{\boldsymbol{b}_{ij}}\mathbb{S}^2$ via the logarithmic map $\operatorname{Log}_{\boldsymbol{b}_{ij}}(\cdot)$, and the coordinates of the resulting tangent vector with respect to the orthonormal basis
$\{\boldsymbol{q}_{ij}^{1},\boldsymbol{q}_{ij}^{2}\}$
form the residual used in the spherical bundle adjustment.}
\label{fig:spherical_ba}
\end{figure}

To address the above limitation while preserving the original angular reprojection optimization objective, this study defines a tangent-space angular reprojection residual function $\mathcal{R}_{ij}:
\mathbb S^2\setminus\{-\boldsymbol b_{ij}\}
\rightarrow\mathbb R^2$ as follows:
\begin{equation}
\mathcal{R}_{ij}(\cdot)
=
\left(
\boldsymbol{q}_{ij}^1,\boldsymbol{q}_{ij}^2\right)^{\top}
\operatorname{Log}_{\boldsymbol {b}_{ij}}(\cdot),
\label{eq:tangent_residual_function}
\end{equation}
where
$\left\{\boldsymbol{q}_{ij}^{1},\boldsymbol{q}_{ij}^{2}\right\}$
is an orthonormal basis for
$T_{\boldsymbol{b}_{ij}}\mathbb S^2$. Since the logarithmic map is locally nonsingular and the basis provides
an isometric coordinate representation of $T_{\boldsymbol b_{ij}}\mathbb S^2$,
the differential
$
D\mathcal{R}_{ij}(\hat{\boldsymbol{b}}_{ij})
:
T_{\hat{\boldsymbol{b}}_{ij}}\mathbb S^2
\rightarrow
\mathbb R^2
$
has full rank.
Therefore, the linearization of \eqref{eq:tangent_residual_function} remains sensitive to perturbations along both
$\boldsymbol{\tau}_{ij}^{\parallel}$
and
$\boldsymbol{\tau}_{ij}^{\perp}$. As illustrated in Fig.~\ref{fig:spherical_ba}, for the predicted viewing ray
$\hat{\boldsymbol{b}}_{ij}$, the corresponding tangent-space angular reprojection residual is expressed as follows:
\begin{equation}
\boldsymbol{r}_{ij}
=\mathcal{R}_{ij}(\hat{\boldsymbol{b}}_{ij})
=\frac{
\mathcal{A}_{ij}(\hat{\boldsymbol{b}}_{ij})
}{
\sin\mathcal{A}_{ij}(\hat{\boldsymbol{b}}_{ij})
}
\left(\boldsymbol{q}_{ij}^1,\boldsymbol{q}_{ij}^2\right)^{\top}
\hat{\boldsymbol{b}}_{ij}
\in\mathbb{R}^2.
\label{eq:tangent_angular_residual}
\end{equation}
The apparent singularity at $\alpha_{ij}=0$ is removable, since
$\lim\limits_{\alpha_{ij}\to 0}
\frac{\alpha_{ij}}{\sin\alpha_{ij}}=1.$ Furthermore, unlike the scalar angular reprojection residual, the tangent-space angular reprojection residual remains differentiable at zero angular deviation.

While the angular reprojection residual constrains the viewing directions on $\mathbb{S}^2$, it provides no constraint on the global reconstruction scale. To introduce complementary scale information along the viewing rays, this study defines a scale regularization term $r^S\in\mathbb{R}$ based on the median depth. Let $\mathcal{O}$ denote the set of observations used in the bundle adjustment, with $i$ and $j$ indexing the corresponding frame and 3-D landmark, respectively. Given an externally provided reference radial depth $d_{ij}>0$ for each observation $(i,j)\in\mathcal{O}$, the median depth-based scale regularization term is defined as follows:
\begin{equation}
r^{S}
=
\log
\frac{
\displaystyle
\underset{(i,j)\in\mathcal{O}}{\operatorname{med}}
\left\|
\hat{\boldsymbol{R}}_i\hat{\boldsymbol{p}}_j^W
+
\hat{\boldsymbol{t}}_i
\right\|_2
}{
\displaystyle
\underset{(i,j)\in\mathcal{O}}{\operatorname{med}}
d_{ij}
}.
\label{eq:median_depth_scale_residual}
\end{equation}

Minimizing the squared residual in \eqref{eq:median_depth_scale_residual} enables the alignment of the median radial depth of the reconstructed 3-D landmarks with the median reference radial depth over $\mathcal{O}$, thereby constraining the global reconstruction scale without enforcing consistency between the radial depth of each 3-D landmark and its corresponding reference radial depth. By combining the angular reprojection residual with the median depth-based scale regularization term, the spherical bundle adjustment jointly optimizes the camera poses and 3-D landmarks. 
Let
$\mathcal{T}=\{\boldsymbol{T}_{i}\}_{i\in\mathcal{K}_{o}}$
and
$\mathcal{P}=\{\boldsymbol{p}_{j}^{W}\}_{j\in\mathcal{Q}_{o}}$
denote the sets of camera poses and 3-D landmarks to be optimized, respectively, where
$\mathcal{K}_o$ and $\mathcal{Q}_o$ are the corresponding index sets for the video frames and 3-D landmarks.
Let $\mathcal{S}=\{{d}_{ij}\}_{(i,j)\in\mathcal{O}}$ denote the set of reference radial depth values used for scale regularization.
The complete optimization objective is therefore formulated as follows:
\begin{equation}
E_{{SBA}}
\left(
\mathcal{T},\mathcal {P};\mathcal{S}
\right)
=
\sum_{(i,j)\in\mathcal O}
\rho\!\left(
\left\|
\boldsymbol{r}_{ij}
\right\|_2^2
\right)
+
\lambda
\left(
r^{S}
\right)^2,
\label{eq:spherical_ba_objective}
\end{equation}
where $\rho(\cdot)$ denotes the Huber robust cost function~\cite{huber1964robust}
and $\lambda>0$ controls the contribution of the median depth-based scale regularization term. For optimization, the median operators in
\eqref{eq:median_depth_scale_residual} are approximated
using a differentiable smooth median
operator~\cite{brown2001smoothed}
to obtain well-defined derivatives for optimization. \eqref{eq:spherical_ba_objective} combines complementary constraints on the viewing directions and reconstruction scale. The tangent-space angular reprojection residual avoids projection-dependent weighting while preserving both local components of angular variation for each observation. Meanwhile, the median depth-based scale regularization term constrains the global scale, which is inherently unobservable from monocular geometry, without directly constraining the depth of each individual 3-D landmark. These two constraints, when used together, enable the joint refinement of the camera poses and 3-D landmarks while anchoring the global reconstruction scale and preserving the flexibility of bundle adjustment. The pseudocode for the proposed spherical bundle adjustment approach is provided in Algorithm~\ref{alg:spherical_ba}.

\algrenewcommand\algorithmicrequire{\textbf{Input:}}
\algrenewcommand\algorithmicensure{\textbf{Output:}}

\begin{algorithm}[H]
\caption{Spherical Bundle Adjustment}
\label{alg:spherical_ba}
\begin{algorithmic}[1]

\Require
Observations $\mathcal{O}$,
reference radial depths $\mathcal{S}$,
initial camera poses $\mathcal{T}$,
initial 3-D landmarks $\mathcal{P}$,
and scale regularization weight $\lambda$

\Ensure
Optimized camera poses $\hat{\mathcal{T}}$
and 3-D landmarks $\hat{\mathcal{P}}$

\State Fix the reference camera pose to eliminate
the pose gauge freedom;

\While{Stopping criterion not satisfied}
    \State Compute $\boldsymbol{r}_{ij}$ for each
    $(i,j)\in\mathcal{O}$ using
    \eqref{eq:predicted_bearing},
    \eqref{eq:angular_reprojection_error},
    and \eqref{eq:tangent_angular_residual};

    \State Compute $r^{S}$ using
    \eqref{eq:median_depth_scale_residual};

    \State Form $E_{\mathrm{SBA}}$ using
    \eqref{eq:spherical_ba_objective};

    \State Linearize the residuals;

    \State Solve for the Levenberg--Marquardt increment;

    \State Update $\mathcal{T}$ and $\mathcal{P}$;
\EndWhile

\State
$\hat{\mathcal{T}}\leftarrow\mathcal{T}$ and 
$\hat{\mathcal{P}}\leftarrow\mathcal{P}$.

\end{algorithmic}
\end{algorithm}
\FloatBarrier

\subsection{Spherical Bundle Adjustment-Guided Unsupervised Fisheye Depth Learning}
~\label{subsec.depth_learning}

In conventional unsupervised monocular depth learning frameworks developed specifically for perspective images~\cite{zhou2017unsupervised,godard2019digging}, a DepthNet estimates a depth map $\boldsymbol{D}_t$ for a target video frame $\boldsymbol{I}_t$, while a PoseNet estimates the relative pose $\boldsymbol{T}_{t\rightarrow t'}$ between $\boldsymbol{I}_t$ and an adjacent source video frame $\boldsymbol{I}_{t'}$, where $t'\in \{t-1,t+1\}$. $\boldsymbol{D}_t$ is first used to unproject the pixels of $\boldsymbol{I}_t$ into the target camera coordinate system. The resulting 3-D points are then transformed into the source camera coordinate system using $\boldsymbol{T}_{t\rightarrow t'}$ and subsequently projected onto $\boldsymbol{I}_{t'}$ to determine the corresponding sampling locations. The target view is then synthesized through differentiable warping based on these sampling locations. 

Unlike the depth estimated independently for each pixel in the target video frame, a single relative camera pose is estimated for each pair of target and
source video frames~\cite{nguyen2024mining}. During view synthesis, the estimated relative pose is applied to all 3-D
points back-projected from the target video frame. Consequently, pose estimation errors induce spatially coherent misalignments throughout the synthesized view, reducing the reliability of the photometric supervision used to train the DepthNet~\cite{ji2021monoindoor}. This issue is further exacerbated for fisheye images, in which severe distortion, particularly toward the image periphery, substantially deforms local structures and makes reliable relative pose estimation considerably more challenging. Accurate relative pose estimation is
thus critical to effective unsupervised fisheye depth learning.

To address this limitation, the proposed framework explicitly refines the relative poses used for view synthesis by minimizing the spherical bundle adjustment objective in \eqref{eq:spherical_ba_objective}, rather than estimating them directly from image pairs using a simple PoseNet. The refined relative poses are subsequently incorporated into the view synthesis process, and the DepthNet is optimized using the resulting photometric supervision, following the training strategy
proposed in the study~\cite{godard2019digging}. Given a target video frame $\boldsymbol{I}_t$ and its adjacent source video frames
$\boldsymbol{I}_{t'}$, the photometric reprojection loss $\mathcal{L}_{P}^{t}$ is defined as follows:
\begin{equation}
\begin{aligned}
\mathcal{L}_P^t
&=
\frac{1}{|\Omega|}
\sum_{\boldsymbol{p}\in\Omega}
\mu_t(\boldsymbol{p})
\min_{t'\in\{t-1,t+1\}}
\Bigg[
\alpha
\frac{
1-\operatorname{SSIM}
(\boldsymbol{I}_t,\hat{\boldsymbol{I}}_{t'\to t})
(\boldsymbol{p})
}{2}
\\
&\qquad\qquad+
(1-\alpha)
\left\|
\boldsymbol{I}_t(\boldsymbol{p})-
\hat{\boldsymbol{I}}_{t'\to t}(\boldsymbol{p})
\right\|_1
\Bigg],
\end{aligned}
\label{eq:photometric_error}
\end{equation}
where $\Omega$ denotes the image domain of
$\boldsymbol{I}_t$, $\boldsymbol{p}=(u,v)^\top$ denotes a pixel in $\boldsymbol{I}_t$, $\operatorname{SSIM}$ denotes the structural similarity index, and $\hat{\boldsymbol{I}}_{t'\to t}$ denotes the target view synthesized from
$\boldsymbol{I}_{t'}$ using the estimated depth map $\boldsymbol{D}_t$ and the refined relative pose $\boldsymbol{T}_{t\rightarrow t'}$. 
Following Monodepth2~\cite{godard2019digging},
$\alpha$ is set to $0.85$, and
$\mu_t(\boldsymbol{p})\in\{0,1\}$ denotes the
auto-masking indicator, which is set to one
when the minimum reprojection error of the
warped images is lower than that of the
original, unwarped source images at pixel
$\boldsymbol{p}$, and zero otherwise.

The following edge-aware smoothness loss is also utilized:
\begin{equation}
\begin{aligned}
\mathcal{L}_S^t
&=
\sum_{q\in\{u,v\}}
\frac{1}{|\Omega_q|}
\sum_{\boldsymbol{p}\in\Omega_q}
\frac{
\left|
\frac{1}{\boldsymbol{D}_t(\boldsymbol{p}+\boldsymbol{e}_q)}
-
\frac{1}{\boldsymbol{D}_t(\boldsymbol{p})}
\right|
}{
|\Omega|^{-1}\sum_{\boldsymbol{q}\in\Omega}\frac{1}{\boldsymbol{D}_t(\boldsymbol{q})}
}
\\
&\qquad\times
\exp{\left(
-\frac{1}{3}
\left\|
\boldsymbol{I}_t(\boldsymbol{p}+\boldsymbol{e}_q)
-
\boldsymbol{I}_t(\boldsymbol{p})
\right\|_1
\right)
}.
\end{aligned}
\label{eq:monodepth2_smoothness_loss}
\end{equation}
where $\boldsymbol{e}_u=(1,0)^\top$ and
$\boldsymbol{e}_v=(0,1)^\top$ represent the horizontal and vertical unit
pixel offsets, respectively, and $\Omega_q=\{\boldsymbol{p}\in\Omega\mid
\boldsymbol{p}+\boldsymbol{e}_q\in\Omega\}$
denotes the valid pixel domain along direction $q\in\{u,v\}$. 

\begin{figure}[!tp]
\centering
\includegraphics[width=\columnwidth]{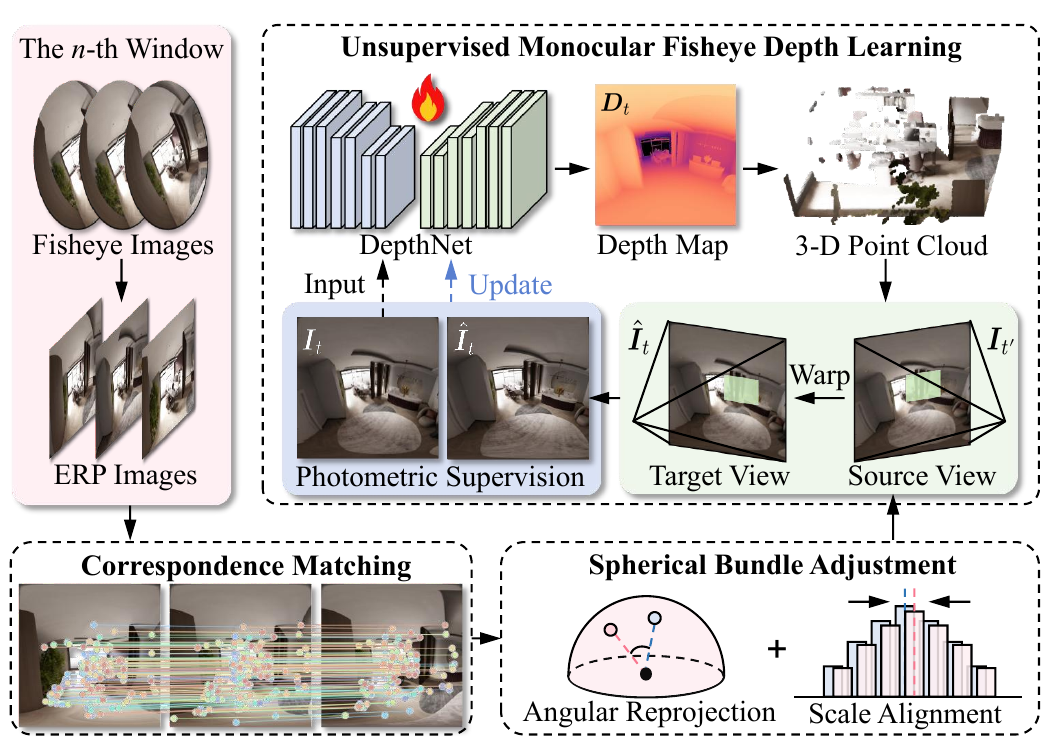}
    \caption{Training process within the $n$-th window of the proposed unsupervised fisheye depth learning framework. Spherical bundle adjustment refines the camera poses and 3-D landmarks, and the refined camera poses are subsequently used for view synthesis to train the DepthNet.}
    \label{fig:depth_learning}
\end{figure}

The photometric loss provides the primary supervision signal for depth
learning, whereas the smoothness loss serves as an edge-aware regularization
term. As illustrated in Fig.~\ref{fig:depth_learning}, each video sequence used for model training is organized into a set of non-overlapping local windows, each consisting of $K$ consecutive video frames. Specifically, the frame set of the $n$-th window is defined as
$\mathcal{W}_n=
\{\boldsymbol{I}_k\}_{k=(n-1)K+1}^{nK}$. The DepthNet estimates a depth map
$\boldsymbol{D}_k$ for each video frame
$\boldsymbol{I}_k\in\mathcal{W}_n$, and the resulting depth maps are stored in the set
$\mathcal{D}_n=
\{\boldsymbol{D}_k\}_{\boldsymbol{I}_k\in\mathcal{W}_n}$.
The camera poses estimated for the video
frames in $\mathcal{W}_n$ are stored in
$\mathcal{T}_n=
\{\boldsymbol{T}_k\}_{\boldsymbol{I}_k\in\mathcal{W}_n}$. Visual correspondences are then established among the video frames within
the window and used together with the estimated camera poses to triangulate a set of 3-D landmarks, denoted by $\mathcal{P}_n$. The resulting depth maps, camera poses, and 3-D landmarks are then
used as inputs to spherical bundle adjustment.

The training procedure alternates between spherical bundle adjustment
refinement and DepthNet optimization within each epoch.
During the $m$-th epoch, the training windows are processed sequentially.
Let $\boldsymbol{\Theta}_{n-1}^{(m)}$ denote the network parameters obtained after processing the preceding $n-1$ windows, with
$\boldsymbol{\Theta}_{0}^{(m)}=\boldsymbol{\Theta}^{(m)}$ denoting the
parameters at the beginning of the epoch. When processing the $n$-th window $\mathcal{W}_{n}$, the DepthNet parameterized
by $\boldsymbol{\Theta}_{n-1}^{(m)}$ first estimates a set of reference depth maps
$\mathcal{D}_{n}^{(m)}$ for the frames in $\mathcal{W}_{n}$. 
Let $\mathcal{O}_n$ denote the set of 3-D landmark observation indices within the $n$-th window. 
The depth estimates sampled at the corresponding
3-D landmark observations form the reference
radial depth set
$\mathcal{S}_n^{(m)}
=\{d_{ij}^{(m)}\}_{(i,j)\in\mathcal{O}_n}$,
which is used for scale regularization
in spherical bundle adjustment. Based on these scale references, the camera poses and 3-D landmarks are jointly refined by minimizing the objective in~\eqref{eq:spherical_ba_objective}, as follows:
\begin{equation}
\left(
\hat{\mathcal{T}}_n^{(m)},
\hat{\mathcal{P}}_n^{(m)}
\right)
=
\underset{
\mathcal{T}_n^{(m)},
\mathcal{P}_n^{(m)}
}{\arg\min}
E_{SBA}
\left(
\mathcal{T}_n^{(m)},
\mathcal{P}_n^{(m)};
\mathcal{S}_n^{(m)}
\right),
\label{eq:window_ba_update}
\end{equation}
where $\mathcal{T}_n^{(m)}$ and $\mathcal{P}_n^{(m)}$ denote the camera poses and 3-D landmarks associated with the $n$-th window during the
$m$-th epoch, respectively, and $\hat{\mathcal{T}}_n^{(m)}$ and $\hat{\mathcal{P}}_n^{(m)}$ denote their corresponding optimized estimates obtained through spherical bundle adjustment. 

Given a target video frame $\boldsymbol{I}_t$ and an adjacent source video frame $\boldsymbol{I}_{t'}$ in $\mathcal{W}_n$,
the refined relative pose during the $m$-th epoch is computed as
$\hat{\boldsymbol{T}}_{t\rightarrow t'}^{(m)}
=
\hat{\boldsymbol{T}}_{t'}^{(m)}
(\hat{\boldsymbol{T}}_{t}^{(m)})^{-1}$.
The refined relative poses are then incorporated into the view synthesis process to guide the optimization of the DepthNet. The training loss $\mathcal{L}_N$ for the $n$-th window $\mathcal{W}_n$ is defined as follows:
\begin{equation}
\begin{aligned}
\mathcal{L}_N
\left(
\boldsymbol{\Theta};
\hat{\mathcal{T}}_n^{(m)}
\right)
&=
\frac{1}{K}
\sum_{t=(n-1)K+1}^{nK}
\mathcal{L}_P^t
\left(
\boldsymbol{\Theta};
\hat{\mathcal{T}}_n^{(m)}
\right)
\\
&\quad+
\frac{\lambda_S}{K}
\sum_{t=(n-1)K+1}^{nK}
\mathcal{L}_S^t
\left(
\boldsymbol{\Theta}
\right),
\end{aligned}
\label{eq:window_depth_loss}
\end{equation}
where $\lambda_S$ denotes the weighting coefficient for the edge-aware
smoothness regularization term. The same loss formulation is applied to all windows.

Starting from $\boldsymbol{\Theta}_{n-1}^{(m)}$, the network parameters for
the $n$-th window are updated by minimizing $\mathcal{L}_N$, while
$\hat{\mathcal{T}}_n^{(m)}$ is held fixed. The
resulting parameters $\boldsymbol{\Theta}_n^{(m)}$ are then carried forward to
the next window. After all $N_W$ windows have been
processed, the network parameters for the next epoch are initialized as follows:
\begin{equation}
\boldsymbol{\Theta}^{(m+1)}
=
\boldsymbol{\Theta}_{N_W}^{(m)},
\label{eq:epoch_parameter_update}
\end{equation}
where $N_W$ denotes the total number of windows. This procedure is
repeated in each subsequent epoch.

Through the above unsupervised depth learning procedure, DepthNet optimization and spherical bundle adjustment are coupled in an alternating optimization process. The predicted depth maps provide scale references for spherical bundle adjustment, while the refined camera poses are subsequently incorporated into the view synthesis process to guide DepthNet optimization. The well-trained DepthNet therefore provides depth estimates that serve as scale references for the monocular fisheye visual odometry framework described in the next section.

\subsection{Scale-Consistent Monocular Fisheye Visual Odometry}
\label{subsec.visual_odometry}

The proposed S$^2$D-FVO framework builds upon the spherical bundle adjustment formulation introduced in Sect.~\ref{subsec.Spherical_BA} and the monocular fisheye DepthNet presented in Sect.~\ref{subsec.depth_learning}. During inference, the well-trained DepthNet provides depth estimates for the input video frames. 

These estimates provide scale references for spherical bundle adjustment, while visual correspondences are used for camera tracking and triangulation. In conventional feature-based SLAM systems such as ORB-SLAM3~\cite{campos2021orbslam3}, 2-D-to-3-D associations during camera tracking are commonly established through local searches around projected 3-D landmarks using camera poses estimated by a constant-velocity motion model. Nonetheless, inaccurately estimated relative poses may cause the corresponding 2-D visual features to fall outside the search regions, resulting in missed 2-D-to-3-D associations~\cite{mur2015orbslam} even when the same features remain visible across consecutive frames. To address this limitation, the proposed framework separates 2-D feature track management from 3-D landmark association.

For each incoming video frame, visual features are independently detected and matched to those retained from the preceding frame. Consecutive frame-to-frame correspondences are linked through persistent feature identities to construct multi-frame 2-D feature tracks, which are maintained independently of their current associations with 3-D landmarks. Consequently, rejecting a geometrically inconsistent association between a 2-D observation and a 3-D landmark does not terminate the underlying feature track. The retained track remains active and can later be reassociated with an existing landmark or used for triangulation or retriangulation once sufficient geometric information becomes available. This design preserves long-term visual correspondences even when individual 2-D-to-3-D associations are temporarily unreliable.

For two consecutive video frames $\boldsymbol{I}_{i}$ and
$\boldsymbol{I}_{i'}$, a set of
visual correspondences
$\mathcal{C}_{ii'}
=
\{(\boldsymbol{p}_{ij},\boldsymbol{p}_{i'j})\}_{j=1}^{N_{ii'}}$ is obtained,
where $\boldsymbol{p}_{ij}=(u_{ij},v_{ij})^\top$ and
$\boldsymbol{p}_{i'j}=(u_{i'j},v_{i'j})^\top$ denote the image
coordinates of the $j$-th pair of matched correspondences, and $N_{ii'}$ denotes
the total number of correspondences between
$\boldsymbol{I}_{i}$ and $\boldsymbol{I}_{i'}$. 
The spherical longitude $\lambda_{ij}$
and latitude $\varphi_{ij}$ corresponding to $\boldsymbol{p}_{ij}$ are computed as follows:
\begin{equation}
\left\{
\begin{aligned}
\lambda_{ij}
&=
\frac{2\pi u_{ij}}{W}-\pi,\\
\varphi_{ij}
&=
\frac{\pi}{2}-\frac{\pi v_{ij}}{H}.
\end{aligned}
\right.
\label{eq:erp_inverse_angles}
\end{equation}
The corresponding unit viewing ray $\boldsymbol{b}_{ij}$ can be obtained using the following expression:
\begin{equation}
\boldsymbol{b}_{ij}
=
\begin{pmatrix}
\cos\varphi_{ij}\sin\lambda_{ij}\\
\sin\varphi_{ij}\\
\cos\varphi_{ij}\cos\lambda_{ij}
\end{pmatrix}.
\label{eq:erp_inverse_projection}
\end{equation}
Similarly, the viewing ray $\boldsymbol{b}_{i'j}$ is obtained from
$\boldsymbol{p}_{i'j}$. The matched viewing rays are then used to estimate the relative rotation matrix
$\hat{\boldsymbol{R}}_{i\to i'}$ and the unit relative translation vector
$\hat{\boldsymbol{t}}_{i\to i'}$. Given the estimated radial depth $\hat{d}_{ij}$, the corresponding 3-D point in the camera coordinate system of the video frame $\boldsymbol{I}_{i}$ is
represented by $\hat{d}_{ij}\boldsymbol{b}_{ij}$. When transformed into the
camera coordinate system of the video frame $\boldsymbol{I}_{i'}$, this 3-D point must be collinear with the corresponding viewing ray
$\boldsymbol{b}_{i'j}$, yielding the following expression:
\begin{equation}
\boldsymbol{b}_{i'j}
\times
\left(
\hat{\boldsymbol{R}}_{i\to i'}
\hat{d}_{ij}\boldsymbol{b}_{ij}
+
s_{i\to i'}
\hat{\boldsymbol{t}}_{i\to i'}
\right)
=
\boldsymbol{0},
\label{eq:translation_scale_constraint}
\end{equation}
where $s_{i\to i'}$ denotes the unknown translation scale. For notational convenience, the following auxiliary vectors are defined:
\begin{equation}
\left\{
\begin{aligned}
\boldsymbol{a}_{j}
&=
[\boldsymbol{b}_{i'j}]_{\times}
\hat{\boldsymbol{t}}_{i\to i'},\\
\boldsymbol{c}_{j}
&=
[\boldsymbol{b}_{i'j}]_{\times}
\hat{\boldsymbol{R}}_{i\to i'}
\left(
\hat{d}_{ij}\boldsymbol{b}_{ij}
\right).
\end{aligned}
\right.
\label{eq:scale_auxiliary_vectors}
\end{equation}
The optimal translation scale is then obtained as the closed-form solution to the following least-squares problem:
\begin{equation}
\begin{aligned}
\hat{s}_{i\to i'}
&=
\underset{s\in\mathbb{R}}{\arg\min}
\sum_j
\left\|
\boldsymbol{c}_j+s\boldsymbol{a}_j
\right\|_2^2\\
&=
-\frac{
\sum_j\boldsymbol{a}_j^\top\boldsymbol{c}_j
}{
\sum_j\|\boldsymbol{a}_j\|_2^2
},
\quad
\sum_j\|\boldsymbol{a}_j\|_2^2>0.
\end{aligned}
\label{eq:depth_scale_init}
\end{equation}
Accordingly, the estimated relative pose is obtained as follows:
\begin{equation}
\hat{\boldsymbol{T}}_{i\to i'}
=
\begin{pmatrix}
\hat{\boldsymbol{R}}_{i\to i'}
&
\hat{s}_{i\to i'}
\hat{\boldsymbol{t}}_{i\to i'}
\\
\boldsymbol{0}^{\top}
&
1
\end{pmatrix}
\in \mathrm{SE}(3),
\label{eq:relative_pose_initialization}
\end{equation}
which provides the relative pose used to initialize the absolute pose during camera tracking.

Camera tracking begins with pose initialization, in which the absolute pose of each incoming video frame with respect to the world coordinate system is estimated through sequential pose propagation and refinement.
For the first video frame $\boldsymbol{I}_{0}$, its camera coordinate system is adopted as the world coordinate system, and the tracked absolute pose $\boldsymbol{T}_{0}^{\mathrm{trk}}$ is set to the identity transformation. 
Subsequently, for each video frame $\boldsymbol{I}_{i-1}$ with $i>1$, its tracked absolute pose $\boldsymbol{T}_{i-1}^{\mathrm{trk}}$ is obtained through the following robust pose-only optimization process:
\begin{equation}
\boldsymbol{T}_{i-1}^{\mathrm{trk}}
=
\underset{\boldsymbol{T}_{i-1}}{\arg\min}
\sum_{j\in\mathcal{O}_{i-1}}
\rho
\left(
\left\|
\boldsymbol{r}_{i-1,j}
\right\|_{2}^{2}
\right),
\label{eq:preceding_pose_tracking}
\end{equation}
where $\mathcal{O}_{i-1}$ denotes the index set of existing 3-D landmarks observed in the $(i-1)$-th video frame.
The resulting tracked absolute pose $\boldsymbol{T}_{i-1}^{\mathrm{trk}}$ is then composed with the relative pose $\hat{\boldsymbol{T}}_{i-1\to i}$, estimated using~\eqref{eq:erp_inverse_angles}--\eqref{eq:relative_pose_initialization}, to obtain the initial absolute pose of the $i$-th video frame as follows:
\begin{equation}
\boldsymbol{T}_{i}^{\mathrm{init}}
=
\hat{\boldsymbol{T}}_{i-1\to i}
\boldsymbol{T}_{i-1}^{\mathrm{trk}} 
\qquad (i\geq 1),
\label{eq:camera_pose_propagation}
\end{equation}
which is then utilized to initialize the robust pose-only optimization using~\eqref{eq:preceding_pose_tracking} over $\mathcal{O}_i$, yielding the tracked absolute pose $\boldsymbol{T}_i^{\mathrm{trk}}$.

\begin{figure}[!tp]
    \centering
    \includegraphics[width=\columnwidth]{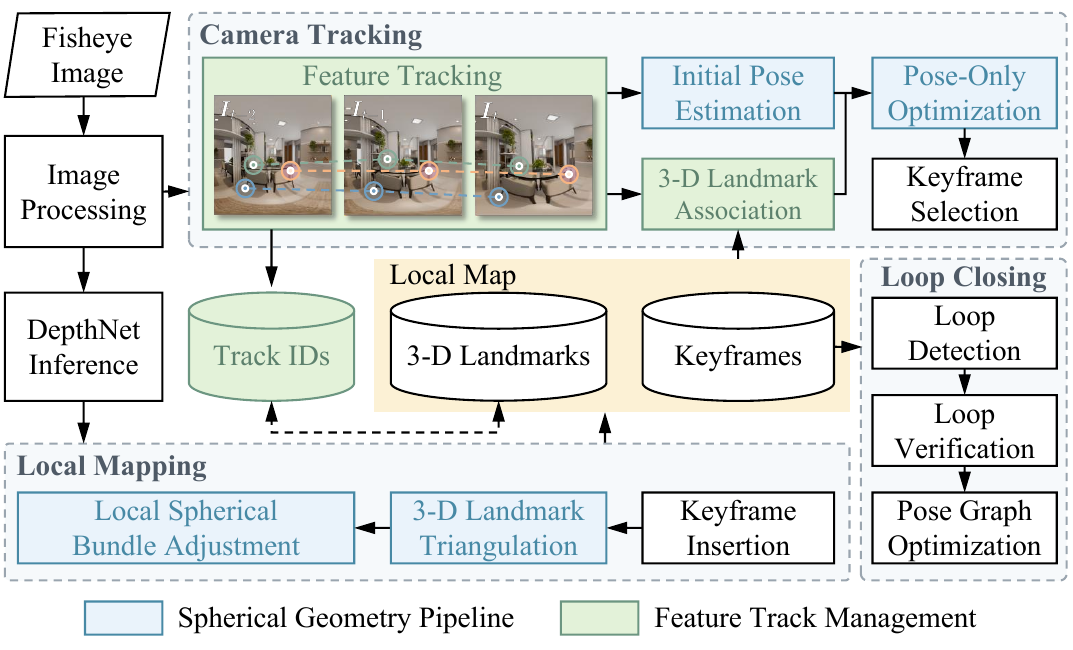}
    \caption{Architecture of the proposed monocular fisheye visual odometry framework, which consists of camera tracking, local mapping, and loop closing. Feature track management is shown in green, while the spherical geometry pipeline is shown in blue.}
    \label{fig:vo_pipeline}
\end{figure}

Each successfully tracked video frame is then analyzed for keyframe selection based on the number of tracked 3-D landmarks, the temporal interval since the preceding keyframe, as well as the available baseline and parallax for reliable triangulation. If the current frame is selected as a new keyframe, it is incorporated into the local map, and its tracked absolute pose serves as the initial absolute pose of the corresponding keyframe in the subsequent local spherical bundle adjustment. The newly selected keyframe and its covisible keyframes constitute the active keyframe set. Other keyframes that observe the same 3-D landmarks provide additional observation constraints, while their absolute poses are held fixed during optimization. Local spherical bundle adjustment jointly refines the absolute poses of the active keyframes and the associated 3-D landmarks by minimizing~\eqref{eq:spherical_ba_objective}. The radial depth estimates provided by the DepthNet at the corresponding 3-D landmarks serve as scale references for the median depth-based scale regularization term used in~\eqref{eq:spherical_ba_objective}. The local map is updated with the optimized absolute poses of the keyframes and the coordinates of 3-D landmarks.

Meanwhile, optimizing the absolute poses of the selected keyframes also refines the corresponding segments of the estimated camera trajectory.
Once a loop closure candidate is detected and validated using a geometrically consistent $\mathrm{Sim}(3)$ transformation, the corresponding loop constraint is incorporated into a global $\mathrm{Sim}(3)$ pose-graph optimization process~\cite{campos2021orbslam3}, thereby reducing accumulated drift and improving global consistency of the estimated trajectory.

Throughout the proposed system, 2-D visual features are represented as unit-sphere bearing vectors. This common spherical representation is consistently used for relative pose estimation, triangulation, pose optimization, and local spherical bundle adjustment, allowing these operations to be performed directly on the unit sphere rather than in the original image plane. The resulting spherical geometric pipeline also provides the basis for the camera model generalization discussed in Sect.~\ref{sec.Discussion}. The architecture of the proposed visual odometry system is illustrated in Fig.~\ref{fig:vo_pipeline}, which shows its main components and their interactions.

Overall, the proposed framework uses monocular fisheye depth estimates as scale references for camera tracking and spherical bundle adjustment, while maintaining global
trajectory consistency through loop closure-based $\mathrm{Sim}(3)$ pose-graph optimization. Together, these components form a unified framework for scale-consistent monocular fisheye visual odometry. 
                                                                        
\section{Experiments}
\label{sec.Experiments}

\subsection{Implementation Details}
\label{sec.exp_implementation}
The proposed monocular fisheye visual odometry system consists of a tracking front end, an optimization back end, and a well-trained DepthNet that provides scale references for camera tracking and spherical bundle adjustment.

DepthNet uses an encoder-decoder architecture with ResNet-18~\cite{he2016deep} as the encoder. Prior to spherical bundle adjustment-guided unsupervised training, DepthNet is first pretrained using depth maps generated by UniDAC. In the unsupervised training process, each video sequence is divided into non-overlapping temporal windows, each containing ten consecutive frames. The resolution of the input ERP image is $512\times512$ pixels. The DepthNet is trained for 20 epochs on an NVIDIA GeForce RTX~5090 GPU using the Adam optimizer. The learning rate is set to $1\times10^{-4}$ for the first 15 epochs and reduced to $1\times10^{-5}$ for the remaining 5 epochs. 

The well-trained DepthNet is then integrated into the monocular fisheye visual odometry system. The front end employs SuperPoint~\cite{detone2018superpoint} for feature extraction and SphereGlue~\cite{gava2023sphereglue} for correspondence matching. The resulting visual correspondences are then used to estimate relative camera poses with a recently proposed birotation solution~\cite{zhao2025birotation}, while robust pose-only optimization and spherical bundle adjustment are carried out in the back end using the Levenberg--Marquardt algorithm implemented in g2o~\cite{kuemmerle2011g2o}, a general graph optimization framework. The overall system is implemented in C++. For practical deployment, DepthNet, SuperPoint, and SphereGlue are exported from PyTorch to ONNX and executed using ONNX Runtime.

\subsection{Datasets}

Extensive experiments are conducted on four public datasets and two newly created datasets to evaluate the performance of the proposed framework. These data are collected from diverse real-world and synthetic environments. Their characteristics and experimental settings are detailed as follows:

\begin{itemize}
    \item 
    KITTI-360~\cite{liao2022kitti360} is a real-world driving dataset collected in suburban environments. Two fisheye cameras, each with a $180^{\circ}$ FoV, are mounted on opposite sides of the vehicle and are modeled using the unified omnidirectional camera model~\cite{mei2007single}, while a Velodyne HDL-64E LiDAR sensor is used to acquire 3-D point clouds. The resolution of the fisheye images is $1,400\times1,400$ pixels. In this study, sequences 00, 02, 03, 04, 05, 07, 09, and 10 captured by the left fisheye camera are used for DepthNet training, while sequence 06 captured by the same camera is used for testing. The training and test sets contain 57,635 and 9,186 fisheye images, respectively. Sparse depth maps are generated by projecting the LiDAR point clouds onto the corresponding fisheye images using the provided camera intrinsics and camera--LiDAR extrinsics. For the evaluation of visual odometry performance, sequences 00, 02, 03, 04, 05, 06, 07, 09, and 10 captured by the left fisheye camera are used.

     \item NIO-Fisheye is a real-world driving dataset collected in diverse urban environments. An NIO ES8 is equipped with a forward-facing wide-angle fisheye camera (Front120) with a FoV of $\sim120^{\circ}$, four surround-view fisheye cameras mounted at the front, left, rear, and right of the vehicle, a front-mounted Innovusion Falcon LiDAR sensor, and IMU/GNSS units. The Front120 camera is modeled using the Kannala--Brandt camera model~\cite{kannala2006generic}, whereas the surround-view cameras are modeled using the Scaramuzza polynomial camera model~\cite{scaramuzza2006toolbox}. The resolutions of the Front120 and surround-view fisheye images are $3,848\times2,168$ and $1,920\times1,536$ pixels, respectively. In this study, sequences 00, 01, 03, and 04 captured by the Front120 camera are used for DepthNet training, while sequence 02 captured by the Front120 camera is used for testing. The training and test sets contain 39,317 and 5,491 fisheye images, respectively. Sparse depth maps are generated by projecting the LiDAR point clouds onto the corresponding fisheye images using the provided camera intrinsics and camera--LiDAR extrinsics. For the evaluation of visual odometry performance, all five sequences captured by the Front120 camera are used.

    \item TUM-VI~\cite{schubert2018tum} is a real-world visual--inertial dataset comprising 28 sequences captured using a handheld stereo fisheye camera rig. The two fisheye cameras have a diagonal FoV of $\sim195^{\circ}$ and are modeled using the Kannala--Brandt camera model~\cite{kannala2006generic}. The resolution of the monochrome fisheye images is $1,024\times1,024$ pixels. Since the TUM-VI dataset does not provide ground-truth depth maps, it is used exclusively for the evaluation of visual odometry performance. Six sequences captured in indoor rooms are selected because they provide motion-capture ground-truth poses over their entire trajectories. Only images captured by the left fisheye camera are used as input to the monocular fisheye visual odometry system.
    
    \item ScanNet++~\cite{yeshwanth2023scannetpp} is a real-world indoor dataset comprising 1,006 scenes captured by a high-resolution laser scanner and a fisheye digital single-lens reflex camera. The fisheye camera is modeled using the Kannala--Brandt camera model~\cite{kannala2006generic}. The resolution of the resized fisheye images is $1,752\times1,168$ pixels. In this study, 35 scenes are selected for DepthNet training, whereas five scenes are selected for testing. The training and test sets contain 16,831 and 1,376 fisheye images, respectively. Radial depth maps for the selected images are rendered from the corresponding 3-D meshes using the provided calibration parameters and camera poses. The data split will be made publicly available upon publication to enable reproducible comparisons.
        
    \item TartanAir V2~\cite{wang2020tartanair} is a large-scale synthetic dataset encompassing a wide range of indoor and outdoor scenarios. The raw RGB images and depth maps are rendered using the Microsoft AirSim simulator with a virtual camera rig consisting of 12 perspective cameras oriented in different directions to provide $360^{\circ}$ coverage. Within each scenario, the virtual camera rig follows multiple distinct trajectories. The fisheye RGB images and corresponding radial depth maps are then generated by resampling the raw perspective RGB images and depth maps according to the linear spherical camera model~\cite{wang2020tartanair}. The resolution of the fisheye images is $1,000\times1,000$ pixels. For the indoor experiments, three scenarios are selected for DepthNet training and two additional scenarios for testing. All fisheye video sequences generated through resampling are included, yielding 35,689 training images and 4,879 test images. For the outdoor experiments, two scenarios are selected for DepthNet training and one additional scenario for testing, yielding 40,384 training images and 5,822 test images. The data split will be made publicly available upon publication to enable reproducible comparisons. For the evaluation of visual odometry performance, all fisheye video sequences in the test sets are used, including 14 indoor and eight outdoor sequences.

    \item MIAS-FisheyeVO is a synthetic dataset generated using Blender by rendering fisheye video sequences from high-quality indoor and outdoor 3-D scenes along controlled camera trajectories. The virtual fisheye cameras have a $180^{\circ}$ FoV and are modeled using the equidistant projection model. The resolution of the rendered fisheye images is $800\times800$ pixels. The radial depth maps are obtained directly during rendering. For the indoor experiments, 71 scenes are used for DepthNet training and six scenes are used for testing. The corresponding training and test sets contain 15,550 and 1,307 fisheye images, respectively. For the outdoor experiments, 41 scenes are used for DepthNet training, and five scenes are used for testing. The corresponding training and test sets contain 13,836 and 2,918 fisheye images, respectively. All six indoor scenes and five outdoor scenes from the test sets are used for the evaluation of visual odometry performance.
\end{itemize}

\subsection{Evaluation Metrics}

In this study, seven evaluation metrics are employed to quantify the performance of DepthNet, including mean absolute relative error (Abs Rel), mean squared relative error (Sq Rel), root mean squared error (RMSE), root mean squared logarithmic error (RMSE log), and the accuracy under three thresholds ($\delta<1.25^i$, $i=1,2,3$). Detailed expressions of these metrics are provided in the study~\cite{eigen2014depth}.

In addition, to ensure fair comparisons in the monocular fisheye visual odometry experiments, each estimated trajectory is first aligned with the corresponding reference trajectory using a single global $\mathrm{Sim}(3)$ transformation~\cite{zhang2018tutorial}. Visual odometry performance is quantified using absolute trajectory error (ATE) and relative pose error (RPE)~\cite{zhang2018tutorial}. The translational and rotational components of RPE, denoted by RPE Trans and RPE Rot, respectively, characterize local translational and rotational motion errors. The percentage of successfully tracked frames (Trck), for which a valid camera pose is estimated, is reported to quantify tracking completeness. For the experiments conducted on the KITTI-360 and NIO-Fisheye datasets, the translational error $e_{t}$ (\%) and rotational error $e_{r}$ ($^\circ$/100\,m) are additionally reported over trajectory segments of different lengths to evaluate translational and rotational drift with respect to the distance traveled. Detailed definitions of these evaluation metrics are provided in studies~\cite{zhang2018tutorial,geiger2012we,munoz2020ucoslam}. Each method is evaluated over five runs, and the median result is reported, consistent with common practice for nondeterministic visual SLAM systems~\cite{campos2021orbslam3}. In the presence of timestamp mismatches, trajectory errors are computed only over temporally overlapping trajectory segments, following the study~\cite{guo2026360dvo}. A run is considered unsuccessful if tracking is lost for more than half of the frames in a sequence. If at least three of the five runs are unsuccessful, the corresponding method on that sequence is marked with the symbol $\times$.

\subsection{Monocular Fisheye Depth Estimation Results}
\label{sec.exp_depth}
\begin{table*}[!tp] 
\centering 
\caption{  
Quantitative comparison with SoTA networks on the KITTI-360, NIO-Fisheye, and ScanNet++ datasets. The best results are shown in bold font. The symbols $\uparrow$ and $\downarrow$ indicate that higher and lower values correspond to better performance, respectively. 
} 
\label{tab:depth_results_aligned} 
 
% \settablefont 
% \setlength{\tabcolsep}{3.8pt} 
% \renewcommand{\arraystretch}{1.25} 
 
\begin{tabular*}{\linewidth}{
@{\extracolsep{\fill}}
cclccccccc
@{}
} 
\toprule 
\textbf{Dataset} 
& \textbf{Camera Model} 
& \textbf{Method} 
& \textbf{Abs Rel} $\downarrow$ 
& \textbf{Sq Rel} $\downarrow$ 
& \textbf{RMSE} $\downarrow$ 
& \textbf{RMSE$_{\log}$} $\downarrow$ 
& $\boldsymbol{\delta < 1.25} \uparrow$ 
& $\boldsymbol{\delta < 1.25^2} \uparrow$ 
& $\boldsymbol{\delta < 1.25^3} \uparrow$ \\ 
\midrule 
 
\multirow{8}{*}{KITTI-360} 
& \multirow{5}{*}{Perspective} 
& Lite-Mono~\cite{zhang2023litemono} 
& 0.123 & 0.586 & \textbf{3.526} & 0.194 
& \textbf{0.860} & 0.958 & 0.983 \\ 
 
& & 
Monodepth2~\cite{godard2019digging} 
& 0.212 & 1.186 & 4.616 & 0.307 
& 0.656 & 0.896 & 0.957 \\ 
 
& & 
SC-DepthV3~\cite{sun2023scdepthv3} 
& 0.147 & 0.849 & 5.217 & 0.240 
& 0.790 & 0.929 & 0.970 \\ 
 
& & 
RA-Depth~\cite{he2022radepth} 
& 0.132 & 0.791 & 4.159 & 0.201 
& 0.847 & 0.955 & 0.982 \\ 
 
& & 
Manydepth2~\cite{zhou2025manydepth2} 
& 0.145 & 0.656 & 3.821 & 0.216 
& 0.804 & 0.946 & 0.980 \\ 
 
\cmidrule(lr){2-10} 
 
& \multirow{3}{*}{Fisheye} 
& FisheyeDepth~\cite{zhao2024fisheyedepth} 
& 0.128 & 0.647 & 3.587 & 0.202 
& 0.850 & 0.954 & 0.981 \\ 
 
& & 
DDS-Net~\cite{xu2026distortion} 
& 0.129 & 0.715 & 4.045 & 0.207 
& 0.844 & 0.950 & 0.979 \\ 
 
& & 
\textbf{Ours} 
& \textbf{0.122} 
& \textbf{0.582} 
& 3.613 
& \textbf{0.183} 
& 0.858 
& \textbf{0.959} 
& \textbf{0.984} \\ 
 
\midrule 
 
\multirow{8}{*}{NIO-Fisheye} 
& \multirow{5}{*}{Perspective} 
& Lite-Mono~\cite{zhang2023litemono} 
& 0.222 & 2.660 & 9.490 & 0.308 
& 0.670 & 0.882 & 0.951 \\ 
 
& & 
Monodepth2~\cite{godard2019digging} 
& 0.214 & 2.790 & 9.323 & 0.314 
& 0.697 & 0.888 & 0.948 \\ 
 
& & 
SC-DepthV3~\cite{sun2023scdepthv3} 
& 0.394 & 4.600 & 11.466 & 0.441 
& 0.442 & 0.719 & 0.873 \\ 
 
& & 
RA-Depth~\cite{he2022radepth} 
& 0.207 & 2.281 & \textbf{9.102} & 0.299 
& 0.685 & 0.893 & 0.955 \\ 
 
& & 
Manydepth2~\cite{zhou2025manydepth2} 
& 0.204 & 2.491 & 9.160 & 0.306 
& 0.703 & 0.890 & 0.951 \\ 
 
\cmidrule(lr){2-10} 
 
& \multirow{3}{*}{Fisheye} 
& FisheyeDepth~\cite{zhao2024fisheyedepth} 
& 0.246 & 3.522 & 10.530 & 0.337 
& 0.630 & 0.852 & 0.934 \\ 
 
& & 
DDS-Net~\cite{xu2026distortion} 
& 0.259 & 3.725 & 11.103 & 0.362 
& 0.594 & 0.837 & 0.929 \\ 
 
& & 
\textbf{Ours} 
& \textbf{0.187} 
& \textbf{2.068} 
& 9.170
& \textbf{0.285} 
& \textbf{0.706} 
& \textbf{0.901} 
& \textbf{0.960} \\ 
 
\midrule 
 
\multirow{8}{*}{ScanNet++} 
& \multirow{5}{*}{Perspective} 
& Lite-Mono~\cite{zhang2023litemono} 
& 0.258 & 0.236 & 0.744 & 0.309 
& 0.589 & 0.851 & 0.949 \\ 
 
& & 
Monodepth2~\cite{godard2019digging} 
& 0.278 & 0.334 & 0.878 & 0.331 
& 0.567 & 0.837 & 0.937 \\ 
 
& & 
SC-DepthV3~\cite{sun2023scdepthv3} 
& 0.328 & 0.380 & 1.010 & 0.456 
& 0.465 & 0.721 & 0.851 \\ 
 
& & 
RA-Depth~\cite{he2022radepth} 
& 0.300 & 0.304 & 0.848 & 0.357 
& 0.491 & 0.792 & 0.925 \\ 
 
& & 
Manydepth2~\cite{zhou2025manydepth2} 
& 0.296 & 0.298 & 0.827 & 0.347 
& 0.537 & 0.810 & 0.924 \\ 
 
\cmidrule(lr){2-10} 
 
& \multirow{3}{*}{Fisheye} 
& FisheyeDepth~\cite{zhao2024fisheyedepth} 
& 0.208 & 0.375 & 0.833 & 0.283 
& 0.760 & 0.906 & 0.952 \\ 
 
& & 
DDS-Net~\cite{xu2026distortion} 
& 0.206 & 0.240 & 0.663 & 0.256 
& 0.726 & 0.908 & 0.966 \\ 
 
& & 
\textbf{Ours} 
& \textbf{0.110} 
& \textbf{0.056} 
& \textbf{0.361} 
& \textbf{0.152} 
& \textbf{0.894} 
& \textbf{0.975} 
& \textbf{0.993} \\ 
 
\bottomrule 
\end{tabular*}
\end{table*}

\begin{figure*}[!tp]
\centering
\includegraphics[width=\textwidth]{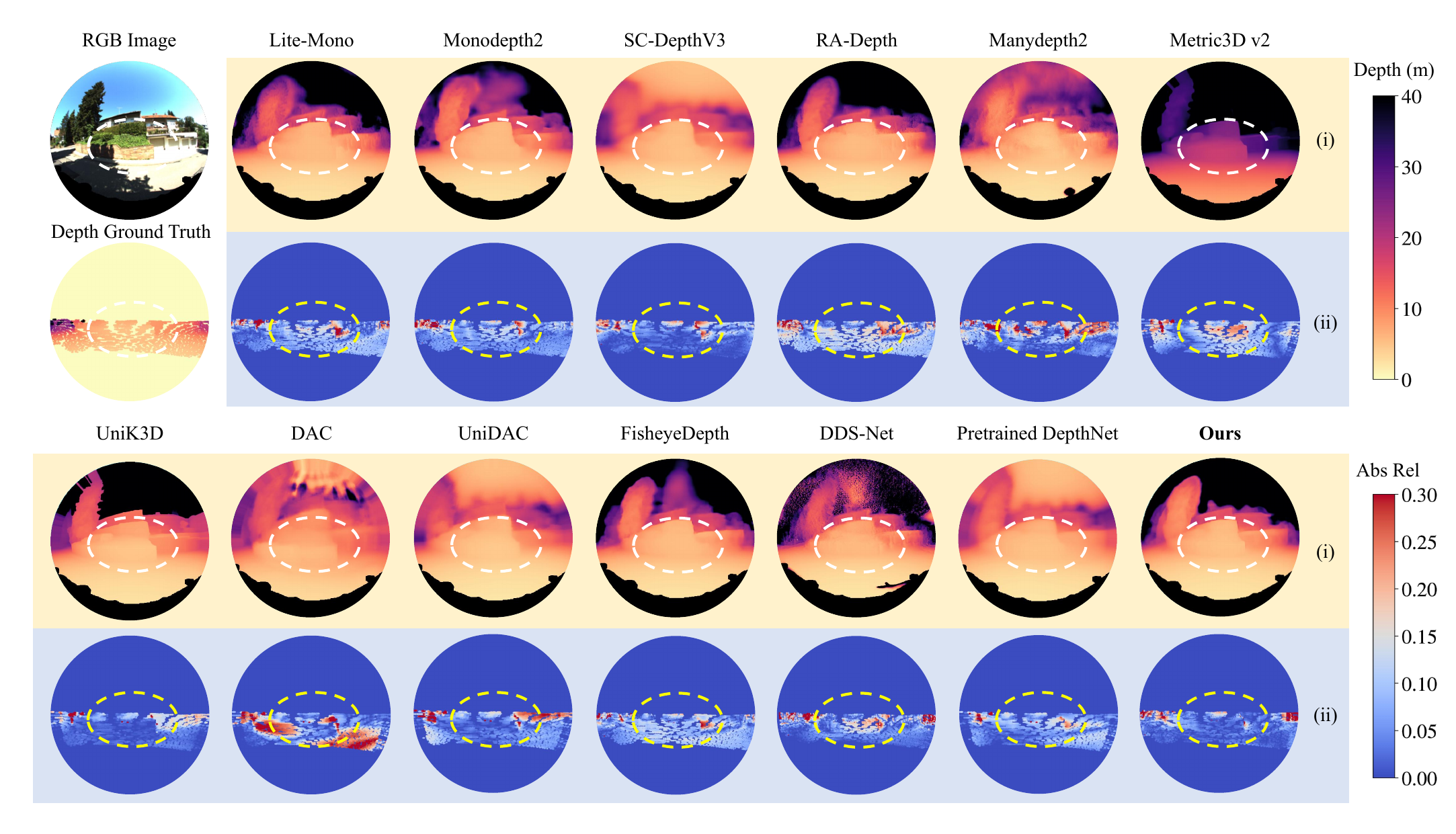}
\caption{
Qualitative comparison of monocular fisheye depth estimation on the KITTI-360 dataset: (i) estimated depth maps; (ii) corresponding error maps. White and yellow dashed contours highlight regions with substantial improvements.
}
\label{fig:qual_depth_kitti360}
\end{figure*}

\begin{figure*}[!tp]
\centering
\includegraphics[width=\textwidth]{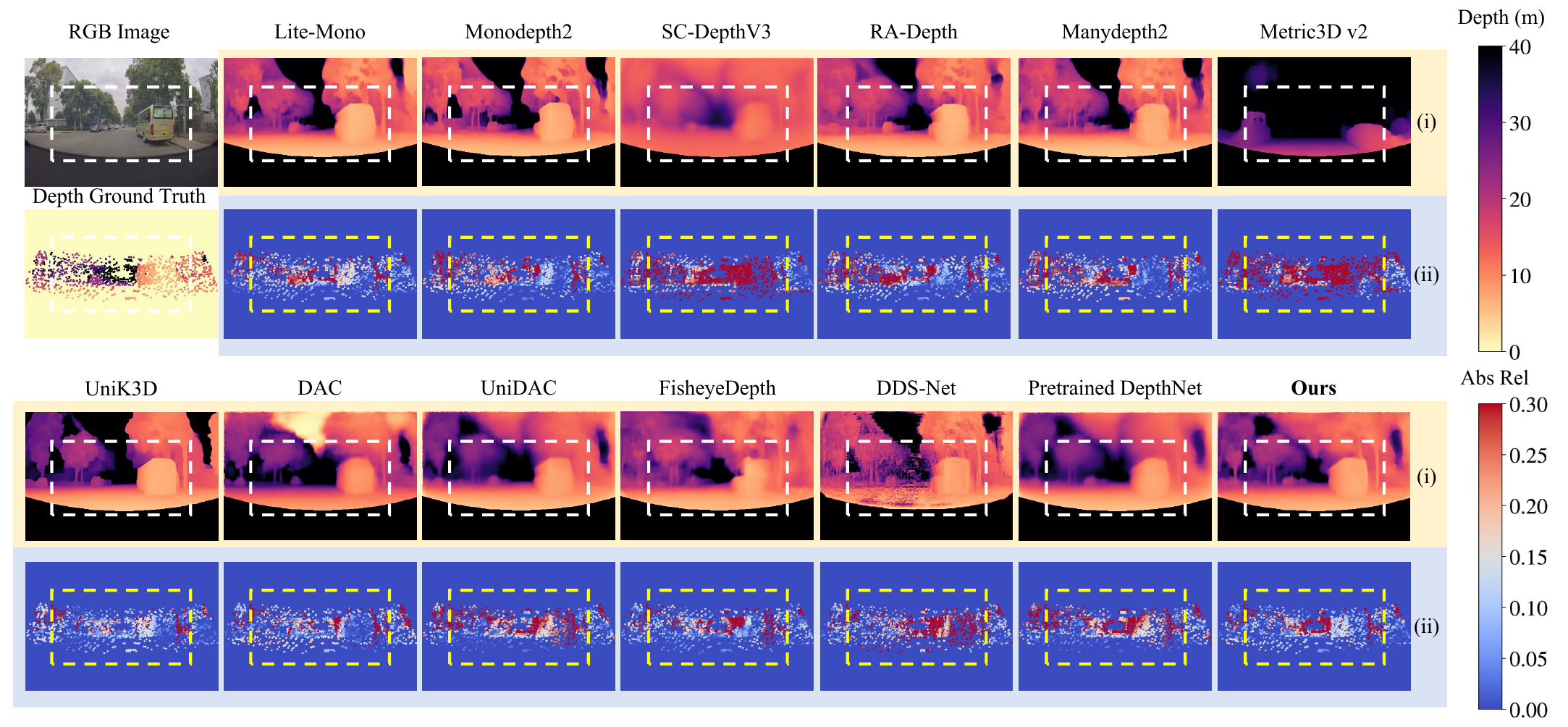}
\caption{
Qualitative comparison of monocular fisheye depth estimation on the NIO-Fisheye dataset: (i) estimated depth maps; (ii) corresponding error maps. White and yellow dashed contours highlight regions with substantial improvements.
}
\label{fig:qual_depth_nio}
\end{figure*}

\begin{figure*}[!tp]
\centering
\includegraphics[width=\textwidth]{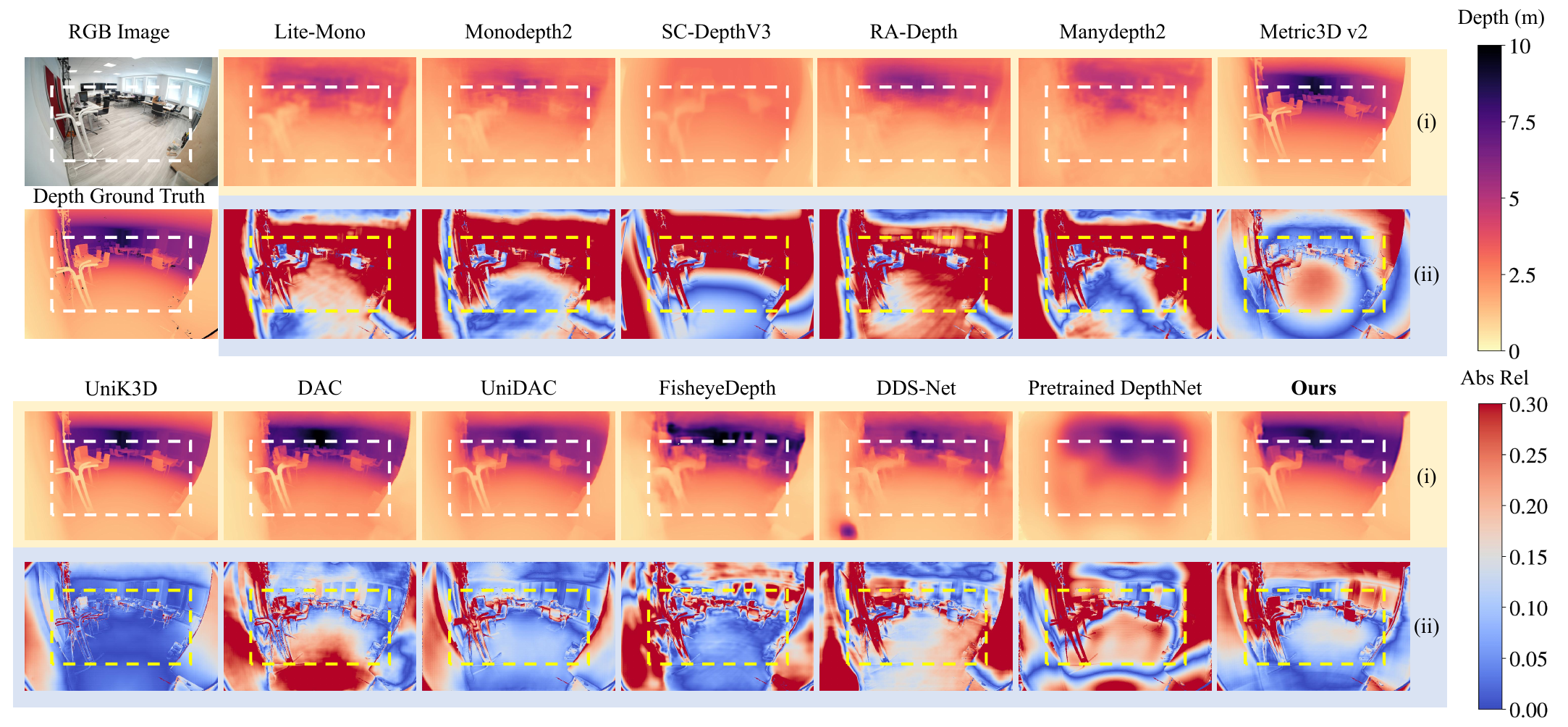}
\caption{
Qualitative comparison of monocular fisheye depth estimation on the ScanNet++ dataset: (i) estimated depth maps; (ii) corresponding error maps. White and yellow dashed contours highlight regions with substantial improvements.
}
\label{fig:qual_depth_scannetpp}
\end{figure*}

\begin{figure*}[!tp]
\centering
\includegraphics[width=\textwidth]{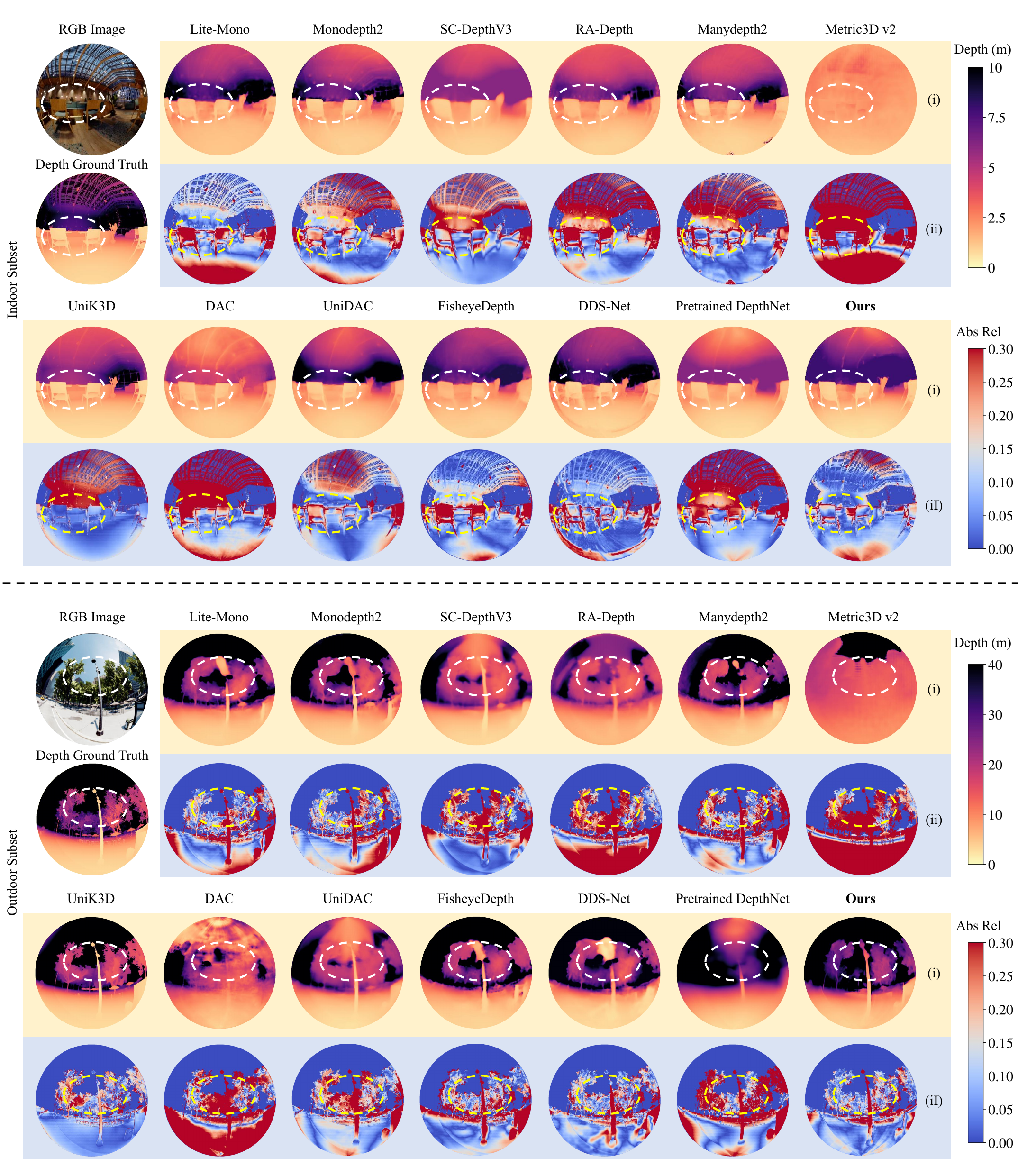}
\caption{
Qualitative comparison of monocular fisheye depth estimation on the TartanAir V2 dataset: (i) estimated depth maps; (ii) corresponding error maps. White and yellow dashed contours highlight regions with substantial improvements.
}
\label{fig:qual_depth_tartanair}
\end{figure*}

\begin{figure*}[!tp]
\centering
\includegraphics[width=\textwidth]{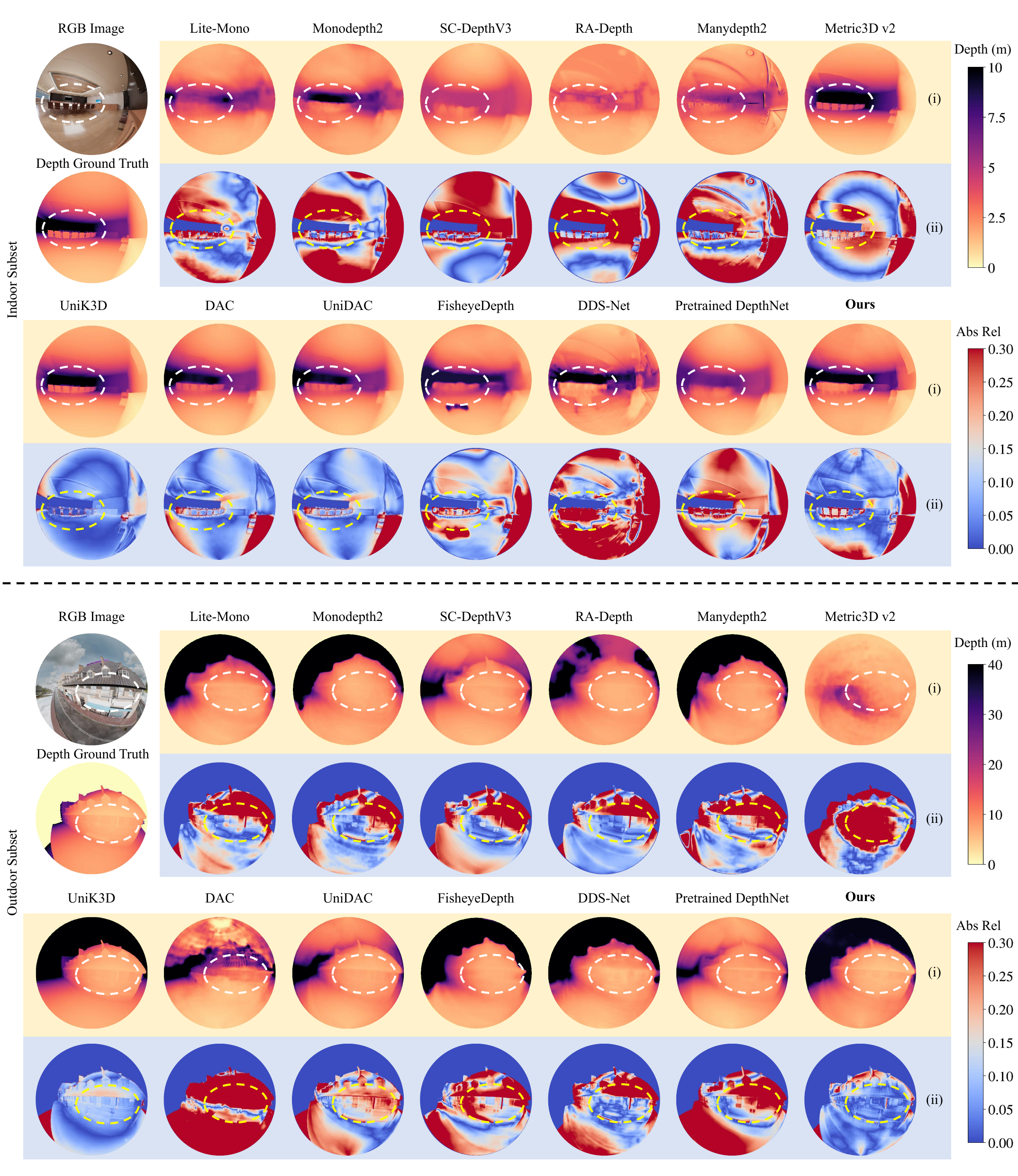}
\caption{Qualitative comparison of monocular fisheye depth estimation on the MIAS-FisheyeVO dataset: (i) estimated depth maps; (ii) corresponding error maps. White and yellow dashed contours highlight regions with substantial improvements.
}
\label{fig:qual_depth_MIAS}
\end{figure*}

This subsection presents a comprehensive evaluation of monocular fisheye depth estimation performance. Qualitative results are shown in Figs.~\ref{fig:qual_depth_kitti360}--\ref{fig:qual_depth_MIAS}, while quantitative results are reported in Tables~\ref{tab:depth_results_aligned} and~\ref{tab:depth_results_aligned_rendered}. These results suggest that the proposed unsupervised monocular fisheye depth learning framework achieves SoTA performance among the compared unsupervised monocular depth estimation methods on all datasets, demonstrating strong effectiveness and robustness in diverse indoor and outdoor scenarios.

The proposed framework consistently outperforms unsupervised monocular depth estimation approaches originally developed for perspective cameras across all datasets. Compared with the best-performing approach developed specifically for perspective cameras, Abs Rel is reduced by $\sim$0.8\%--57.4\%, with the largest reductions reaching 57.4\% on the ScanNet++ dataset, 49.1\% on the outdoor subset of the TartanAir V2 dataset, and 29.9\% on the indoor subset of the MIAS-FisheyeVO dataset. 
To ensure a fair comparison, the projection and unprojection operations involved in the view synthesis process are reformulated according to the fisheye camera model adopted for each dataset. The experimental results suggest that accurate fisheye camera modeling alone is insufficient to provide reliable photometric supervision for depth learning. The nonuniform projection of fisheye cameras causes the image displacement induced by camera motion to vary substantially with image location, increasing the complexity of the motion patterns that PoseNet must learn. As a result, accurate relative pose estimation becomes particularly important for reliable photometric supervision in monocular fisheye depth learning. The proposed framework therefore refines the relative camera poses through spherical bundle adjustment before view synthesis, reducing pixel misalignment and improving the reliability of photometric supervision.

\begin{table*}[!tp]
\centering
\caption{
Quantitative comparison with SoTA networks on the TartanAir V2 and MIAS-FisheyeVO datasets. The best results are shown in bold font. The symbols $\uparrow$ and $\downarrow$ indicate that higher and lower values correspond to better performance, respectively.
}
\label{tab:depth_results_aligned_rendered}

\makebox[\textwidth][c]{%
\begin{tabular}{@{}ccclccccccc@{}}
\toprule
\textbf{Dataset}
& \textbf{Subset}
& \textbf{Camera Model}
& \textbf{Method}
& \textbf{Abs Rel} $\downarrow$
& \textbf{Sq Rel} $\downarrow$
& \textbf{RMSE} $\downarrow$
& \textbf{RMSE$_{\log}$} $\downarrow$
& $\boldsymbol{\delta < 1.25} \uparrow$
& $\boldsymbol{\delta < 1.25^2} \uparrow$
& $\boldsymbol{\delta < 1.25^3} \uparrow$ \\
\midrule

\multirow{16}{*}{TartanAir V2}
& \multirow{8}{*}{Indoor}
& \multirow{5}{*}{Perspective}
& Lite-Mono~\cite{zhang2023litemono}
& 0.262 & 0.235 & 0.730 & 0.315
& 0.568 & 0.850 & 0.951 \\

& & &
Monodepth2~\cite{godard2019digging}
& 0.242 & 0.237 & 0.700 & 0.294
& 0.621 & 0.877 & 0.959 \\

& & &
SC-DepthV3~\cite{sun2023scdepthv3}
& 0.242 & 0.209 & 0.702 & 0.307
& 0.607 & 0.869 & 0.956 \\

& & &
RA-Depth~\cite{he2022radepth}
& 0.243 & 0.183 & 0.647 & 0.290
& 0.614 & 0.873 & 0.960 \\

& & &
Manydepth2~\cite{zhou2025manydepth2}
& 0.237 & 0.207 & 0.686 & 0.294
& 0.615 & 0.875 & 0.960 \\

\cmidrule(lr){3-11}

& &
\multirow{3}{*}{Fisheye}
& FisheyeDepth~\cite{zhao2024fisheyedepth}
& 0.263 & 0.261 & 0.768 & 0.330
& 0.566 & 0.838 & 0.940 \\

& & &
DDS-Net~\cite{xu2026distortion}
& 0.240 & 0.211 & 0.710 & 0.301
& 0.594 & 0.869 & 0.958 \\

& & &
\textbf{Ours}
& \textbf{0.205}
& \textbf{0.137}
& \textbf{0.566}
& \textbf{0.258}
& \textbf{0.649}
& \textbf{0.918}
& \textbf{0.979} \\

\cmidrule(lr){2-11}

& \multirow{8}{*}{Outdoor}
& \multirow{5}{*}{Perspective}
& Lite-Mono~\cite{zhang2023litemono}
& 0.457 & 3.292 & 7.243 & 0.521
& 0.392 & 0.657 & 0.811 \\

& & &
Monodepth2~\cite{godard2019digging}
& 0.391 & 3.560 & 9.405 & 0.504
& 0.425 & 0.693 & 0.835 \\

& & &
SC-DepthV3~\cite{sun2023scdepthv3}
& 0.400 & 2.506 & 7.567 & 0.526
& 0.411 & 0.666 & 0.819 \\

& & &
RA-Depth~\cite{he2022radepth}
& 0.491 & 3.240 & 8.013 & 0.598
& 0.334 & 0.583 & 0.751 \\

& & &
Manydepth2~\cite{zhou2025manydepth2}
& 0.455 & 3.000 & 7.618 & 0.557
& 0.362 & 0.622 & 0.783 \\

\cmidrule(lr){3-11}

& &
\multirow{3}{*}{Fisheye}
& FisheyeDepth~\cite{zhao2024fisheyedepth}
& 0.482 & 3.238 & 7.499 & 0.619
& 0.343 & 0.593 & 0.753 \\

& & &
DDS-Net~\cite{xu2026distortion}
& 0.370 & 1.869 & 5.032 & 0.453
& 0.441 & 0.716 & 0.859 \\

& & &
\textbf{Ours}
& \textbf{0.199}
& \textbf{0.836}
& \textbf{4.218}
& \textbf{0.263}
& \textbf{0.721}
& \textbf{0.912}
& \textbf{0.966} \\

\midrule

\multirow{16}{*}{MIAS-FisheyeVO}
& \multirow{8}{*}{Indoor}
& \multirow{5}{*}{Perspective}
& Lite-Mono~\cite{zhang2023litemono}
& 0.326 & 0.473 & 1.049 & 0.365
& 0.536 & 0.807 & 0.918 \\

& & &
Monodepth2~\cite{godard2019digging}
& 0.333 & 0.465 & 1.051 & 0.365
& 0.523 & 0.803 & 0.918 \\

& & &
SC-DepthV3~\cite{sun2023scdepthv3}
& 0.274 & 0.330 & 1.041 & 0.328
& 0.562 & 0.835 & 0.938 \\

& & &
RA-Depth~\cite{he2022radepth}
& 0.350 & 0.481 & 1.237 & 0.403
& 0.450 & 0.737 & 0.895 \\

& & &
Manydepth2~\cite{zhou2025manydepth2}
& 0.350 & 0.529 & 1.144 & 0.393
& 0.503 & 0.772 & 0.900 \\

\cmidrule(lr){3-11}

& &
\multirow{3}{*}{Fisheye}
& FisheyeDepth~\cite{zhao2024fisheyedepth}
& 0.343 & 0.577 & 1.194 & 0.398
& 0.507 & 0.769 & 0.894 \\

& & &
DDS-Net~\cite{xu2026distortion}
& 0.302 & 0.399 & 1.010 & 0.343
& 0.560 & 0.824 & 0.928 \\

& & &
\textbf{Ours}
& \textbf{0.192}
& \textbf{0.210}
& \textbf{0.701}
& \textbf{0.233}
& \textbf{0.756}
& \textbf{0.921}
& \textbf{0.968} \\

\cmidrule(lr){2-11}

& \multirow{8}{*}{Outdoor}
& \multirow{5}{*}{Perspective}
& Lite-Mono~\cite{zhang2023litemono}
& 0.314 & 2.770 & 6.606 & 0.391
& 0.576 & 0.815 & 0.907 \\

& & &
Monodepth2~\cite{godard2019digging}
& 0.306 & 2.528 & 6.828 & 0.402
& 0.562 & 0.801 & 0.902 \\

& & &
SC-DepthV3~\cite{sun2023scdepthv3}
& 0.230 & 1.599 & 5.910 & 0.328
& 0.676 & 0.873 & 0.938 \\

& & &
RA-Depth~\cite{he2022radepth}
& 0.300 & 2.149 & 6.651 & 0.395
& 0.566 & 0.811 & 0.907 \\

& & &
Manydepth2~\cite{zhou2025manydepth2}
& 0.362 & 4.003 & 7.207 & 0.430
& 0.514 & 0.776 & 0.888 \\

\cmidrule(lr){3-11}

& &
\multirow{3}{*}{Fisheye}
& FisheyeDepth~\cite{zhao2024fisheyedepth}
& 0.314 & 2.553 & 6.426 & 0.397
& 0.549 & 0.800 & 0.905 \\

& & &
DDS-Net~\cite{xu2026distortion}
& 0.290 & 3.082 & 6.694 & 0.378
& 0.612 & 0.841 & 0.919 \\

& & &
\textbf{Ours}
& \textbf{0.203}
& \textbf{1.250}
& \textbf{5.234}
& \textbf{0.286}
& \textbf{0.715}
& \textbf{0.897}
& \textbf{0.954} \\

\bottomrule
\end{tabular}%
}
\end{table*}

The comparison with SC-DepthV3~\cite{sun2023scdepthv3} further demonstrates the effectiveness of the proposed training strategy. SC-DepthV3 generates depth maps from single images using UniDAC~\cite{ganesan2026unidac} and introduces additional losses based on these depth maps, which are jointly minimized with the photometric loss during unsupervised DepthNet training. In contrast, the proposed framework adopts a two-stage training strategy. DepthNet is first pretrained using depth maps generated by UniDAC in a fully supervised manner, followed by unsupervised training using photometric and smoothness losses, with spherical bundle adjustment refining the relative camera poses used for view synthesis. Compared with SC-DepthV3, the proposed framework reduces Abs Rel by $\sim$11.7\%--66.5\% across all datasets and improves $\delta<1.25$ by $\sim$5.8\%--92.3\%. The performance gains are particularly pronounced on the NIO-Fisheye and ScanNet++ datasets, as well as the outdoor subset of the TartanAir V2 dataset, where Abs Rel is reduced by $\sim$52.5\%, 66.5\%, and 50.2\%, respectively. These results indicate that the proposed training strategy effectively refines the pretrained DepthNet through subsequent unsupervised learning using relative camera poses refined by spherical bundle adjustment.

\begin{table*}[!tp]
\centering
\caption{
Quantitative comparison with SoTA foundation models tailored for metric depth estimation. Depth estimates are scaled by aligning their median values before evaluation. The best results are shown in bold font, and the second-best results are underlined. The symbol $\dagger$ denotes methods supporting
different camera models.
}
\label{tab:depth_results}

\begin{tabular*}{\linewidth}{
@{\extracolsep{\fill}}
lll*{7}{c}
@{}
}
\toprule
\textbf{Dataset}
& \textbf{Subset}
& \textbf{Method}
& \textbf{Abs Rel} $\downarrow$
& \textbf{Sq Rel} $\downarrow$
& \textbf{RMSE} $\downarrow$
& \textbf{RMSE$_{\log}$} $\downarrow$
& $\boldsymbol{\delta < 1.25} \uparrow$
& $\boldsymbol{\delta < 1.25^2} \uparrow$
& $\boldsymbol{\delta < 1.25^3} \uparrow$ \\
\midrule

% ==================== KITTI-360 ====================
\multirow{6}{*}{KITTI-360}
& \multirow{6}{*}{--}
& Metric3D v2~\cite{hu2024metric3dv2}
& 0.172 & 0.837 & 5.012 & 0.254
& 0.735 & 0.931 & 0.972 \\

& &
UniK3D$^\dagger$~\cite{piccinelli2025unik3d}
& \textbf{0.093}
& \textbf{0.426}
& \textbf{2.930}
& \textbf{0.159}
& \textbf{0.924}
& \textbf{0.976}
& \textbf{0.988} \\

& &
DAC$^\dagger$~\cite{guo2025depthanycamera}
& 0.193 & 1.042 & 5.097 & 0.283
& 0.689 & 0.903 & 0.963 \\

& &
UniDAC$^\dagger$~\cite{ganesan2026unidac}
& 0.186 & 0.990 & 5.128 & 0.275
& 0.698 & 0.908 & 0.966 \\

& &
Pretrained DepthNet
& 0.179 & 0.808 & 4.588 & 0.251
& 0.703 & 0.922 & 0.977 \\

& &
\textbf{Ours}
& \underline{0.122}
& \underline{0.582}
& \underline{3.613}
& \underline{0.183}
& \underline{0.858}
& \underline{0.959}
& \underline{0.984} \\

\midrule

% ==================== NIO-Fisheye ====================
\multirow{6}{*}{NIO-Fisheye}
& \multirow{6}{*}{--}
& Metric3D v2~\cite{hu2024metric3dv2}
& \underline{0.163}
& \textbf{1.887}
& \underline{8.063}
& \underline{0.271}
& \underline{0.805}
& \underline{0.926}
& \underline{0.961} \\

& &
UniK3D$^\dagger$~\cite{piccinelli2025unik3d}
& \textbf{0.161}
& 2.096
& \textbf{7.728}
& \textbf{0.265}
& \textbf{0.820}
& \textbf{0.928}
& \textbf{0.962} \\

& &
DAC$^\dagger$~\cite{guo2025depthanycamera}
& 0.300
& 4.856
& 11.328
& 0.384
& 0.597
& 0.823
& 0.910 \\

& &
UniDAC$^\dagger$~\cite{ganesan2026unidac}
& 0.303
& 4.704
& 13.964
& 0.446
& 0.482
& 0.737
& 0.858 \\

& &
Pretrained DepthNet
& 0.210
& 2.240
& 9.424
& 0.305
& 0.657
& 0.884
& 0.961 \\

& &
\textbf{Ours}
& 0.187
& \underline{2.068}
& 9.170
& 0.285
& 0.706
& 0.901
& 0.960 \\

\midrule

% ==================== ScanNet++ ====================
\multirow{6}{*}{ScanNet++}
& \multirow{6}{*}{--}
& Metric3D v2~\cite{hu2024metric3dv2}
& 0.154 & 0.102 & 0.548 & 0.214
& 0.774 & 0.958 & 0.989 \\

& &
UniK3D$^\dagger$~\cite{piccinelli2025unik3d}
& \textbf{0.099}
& \textbf{0.043}
& \textbf{0.318}
& \textbf{0.136}
& \textbf{0.925}
& \textbf{0.988}
& \textbf{0.996} \\

& &
DAC$^\dagger$~\cite{guo2025depthanycamera}
& 0.209 & 0.241 & 0.728 & 0.265
& 0.666 & 0.888 & 0.961 \\

& &
UniDAC$^\dagger$~\cite{ganesan2026unidac}
& 0.211 & 0.254 & 0.765 & 0.313
& 0.670 & 0.888 & 0.956 \\

& &
Pretrained DepthNet
& 0.235 & 0.197 & 0.666 & 0.408
& 0.626 & 0.864 & 0.945 \\

& &
\textbf{Ours}
& \underline{0.110}
& \underline{0.056}
& \underline{0.361}
& \underline{0.152}
& \underline{0.894}
& \underline{0.975}
& \underline{0.993} \\

\midrule

% ==================== TartanAir V2 ====================
\multirow{12}{*}{TartanAir V2}
& \multirow{6}{*}{Indoor}
& Metric3D v2~\cite{hu2024metric3dv2}
& 0.261 & 0.239 & 0.783 & 0.331
& 0.561 & 0.849 & 0.939 \\

& &
UniK3D$^\dagger$~\cite{piccinelli2025unik3d}
& \textbf{0.110}
& \textbf{0.040}
& \textbf{0.319}
& \textbf{0.142}
& \textbf{0.893}
& \textbf{0.988}
& \textbf{0.998} \\

& &
DAC$^\dagger$~\cite{guo2025depthanycamera}
& \underline{0.174}
& \underline{0.124}
& \underline{0.557}
& \underline{0.236}
& \underline{0.750}
& \underline{0.932}
& 0.978 \\

& &
UniDAC$^\dagger$~\cite{ganesan2026unidac}
& 0.202 & 0.185 & 0.639 & 0.271
& 0.712 & 0.916 & 0.965 \\

& &
Pretrained DepthNet
& 0.254 & 0.208 & 0.697 & 0.308
& 0.575 & 0.860 & 0.956 \\

& &
\textbf{Ours}
& 0.205 & 0.137 & 0.566 & 0.258
& 0.649 & 0.918 & \underline{0.979} \\

\cmidrule(lr){2-10}

&
\multirow{6}{*}{Outdoor}
& Metric3D v2~\cite{hu2024metric3dv2}
& 0.493 & 3.773 & 8.881 & 0.649
& 0.321 & 0.567 & 0.731 \\

& &
UniK3D$^\dagger$~\cite{piccinelli2025unik3d}
& \textbf{0.125}
& \textbf{0.613}
& \textbf{3.201}
& \textbf{0.174}
& \textbf{0.865}
& \textbf{0.975}
& \textbf{0.992} \\

& &
DAC$^\dagger$~\cite{guo2025depthanycamera}
& 0.516 & 4.814 & 8.480 & 0.627
& 0.325 & 0.576 & 0.748 \\

& &
UniDAC$^\dagger$~\cite{ganesan2026unidac}
& 0.294 & 2.128 & 6.547 & 0.392
& 0.560 & 0.808 & 0.908 \\

& &
Pretrained DepthNet
& 0.407 & 2.730 & 7.140 & 0.518
& 0.399 & 0.667 & 0.822 \\

& &
\textbf{Ours}
& \underline{0.199}
& \underline{0.836}
& \underline{4.218}
& \underline{0.263}
& \underline{0.721}
& \underline{0.912}
& \underline{0.966} \\

\midrule

% ==================== MIAS-FisheyeVO ====================
\multirow{12}{*}{MIAS-FisheyeVO}
& \multirow{6}{*}{Indoor}
& Metric3D v2~\cite{hu2024metric3dv2}
& 0.256 & 0.309 & 0.984 & 0.314
& 0.606 & 0.869 & 0.941 \\

& &
UniK3D$^\dagger$~\cite{piccinelli2025unik3d}
& \textbf{0.116}
& \textbf{0.110}
& \textbf{0.457}
& \textbf{0.149}
& \textbf{0.895}
& \textbf{0.976}
& \textbf{0.990} \\

& &
DAC$^\dagger$~\cite{guo2025depthanycamera}
& 0.263 & 0.312 & 0.859 & 0.307
& 0.618 & 0.869 & 0.947 \\

& &
UniDAC$^\dagger$~\cite{ganesan2026unidac}
& 0.207 & 0.238 & 0.791 & 0.262
& 0.709 & 0.921 & 0.965 \\

& &
Pretrained DepthNet
& 0.310 & 0.402 & 1.015 & 0.360
& 0.506 & 0.803 & 0.926 \\

& &
\textbf{Ours}
& \underline{0.192}
& \underline{0.210}
& \underline{0.701}
& \underline{0.233}
& \underline{0.756}
& \underline{0.921}
& \underline{0.968} \\

\cmidrule(lr){2-10}

&
\multirow{6}{*}{Outdoor}
& Metric3D v2~\cite{hu2024metric3dv2}
& 0.277 & 2.618 & 7.377 & 0.417
& 0.596 & 0.831 & 0.908 \\

& &
UniK3D$^\dagger$~\cite{piccinelli2025unik3d}
& \textbf{0.149}
& 2.802
& 5.553
& \textbf{0.230}
& \textbf{0.897}
& \textbf{0.955}
& \textbf{0.970} \\

& &
DAC$^\dagger$~\cite{guo2025depthanycamera}
& 0.445 & 5.181 & 8.407 & 0.509
& 0.396 & 0.665 & 0.826 \\

& &
UniDAC$^\dagger$~\cite{ganesan2026unidac}
& 0.222
& \underline{1.411}
& \underline{5.384}
& 0.310
& 0.636
& \underline{0.905}
& \underline{0.956} \\

& &
Pretrained DepthNet
& 0.251 & 1.680 & 5.981 & 0.349
& 0.613 & 0.860 & 0.935 \\

& &
\textbf{Ours}
& \underline{0.203}
& \textbf{1.250}
& \textbf{5.234}
& \underline{0.286}
& \underline{0.715}
& 0.897
& 0.954 \\

\bottomrule
\end{tabular*}
\end{table*}

Furthermore, additional comparisons are conducted with FisheyeDepth~\cite{zhao2024fisheyedepth} and DDS-Net~\cite{xu2026distortion}, two SoTA approaches tailored for fisheye cameras. FisheyeDepth relies on externally provided pose scales, whereas DDS-Net introduces dedicated modules to model fisheye distortion. In contrast, the proposed framework refines relative camera poses through spherical bundle adjustment while retaining a standard DepthNet architecture. Compared with FisheyeDepth and DDS-Net, the proposed framework reduces Abs Rel by $\sim$4.7\%--58.7\% and $\sim$5.4\%--46.6\%, respectively, across all datasets. The comparison with FisheyeDepth further suggests that introducing pose scales alone is insufficient to achieve comparable performance, supporting the effectiveness of supervised DepthNet pretraining together with subsequent spherical bundle adjustment-guided unsupervised training. The comparison with DDS-Net also suggests that superior fisheye depth estimation performance can be achieved without introducing dedicated distortion-aware modules.

Consistent performance gains are also observed across datasets with substantially different scene and motion characteristics. Relatively moderate improvements are observed on autonomous driving datasets, such as KITTI-360 and NIO-Fisheye, whereas more pronounced gains are achieved on other datasets. The differences in performance gains across these datasets are closely related to the camera motion, viewpoint changes, and scene structure. In driving scenarios, the motion between adjacent frames is typically more constrained and regular, allowing PoseNet to estimate relative poses with comparatively high reliability. As a result, the poses used for view synthesis already provide sufficiently accurate frame-to-frame transformations for most frame pairs, limiting the potential for further improvement through spherical bundle adjustment. In contrast, more diverse camera motions and larger viewpoint changes increase the difficulty of relative pose estimation and make view synthesis more sensitive to pose errors, thereby weakening the resulting photometric supervision. Under these conditions, the pose refinement provided by spherical bundle adjustment has a greater effect on the quality of photometric supervision, accounting for the larger performance gains observed on these datasets.

\begin{figure*}
    \centering
    \includegraphics[width=\textwidth]{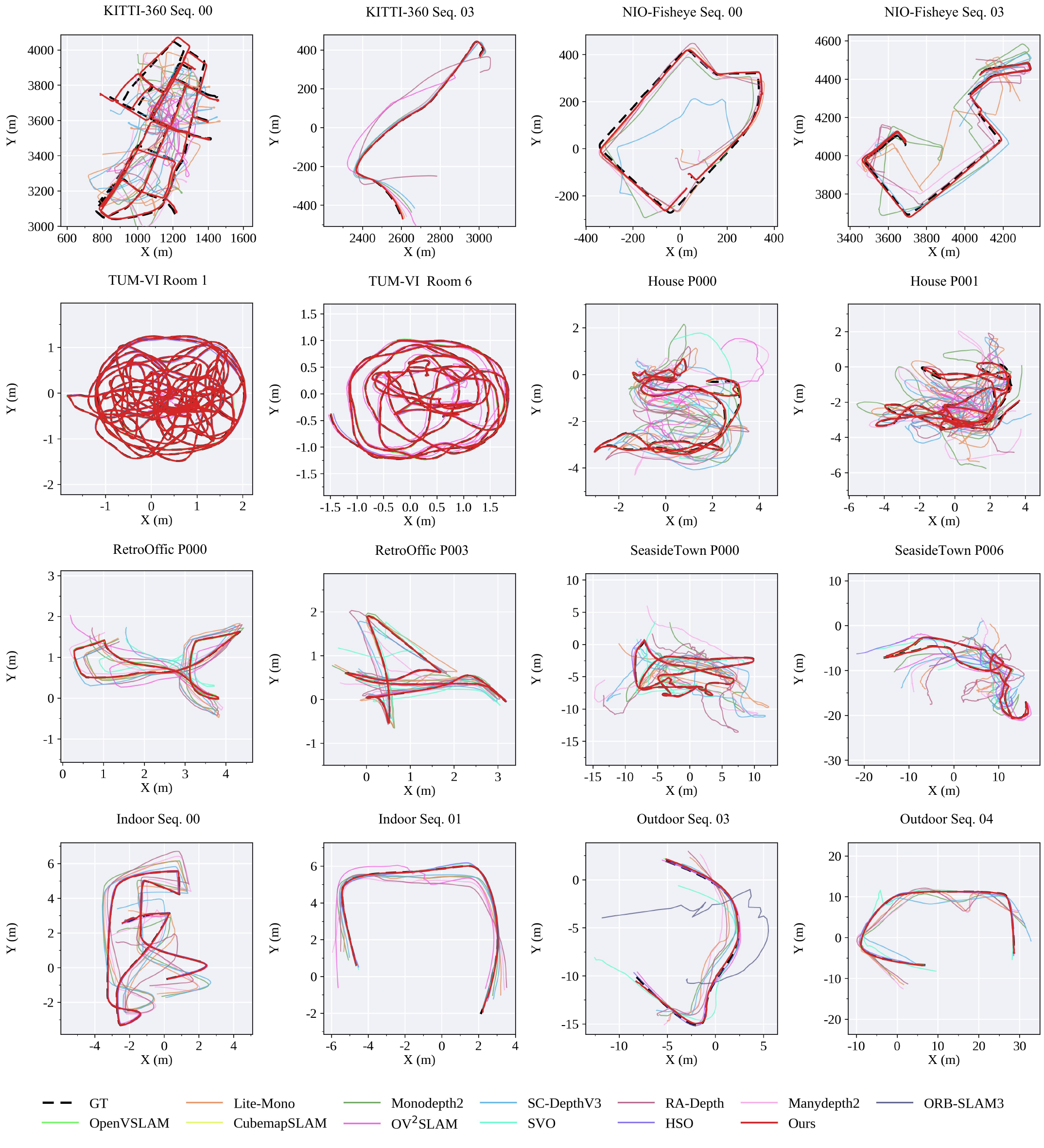}
    \caption{Comparison of estimated camera trajectories on the KITTI-360, NIO-Fisheye, TUM-VI, TartanAir V2, and MIAS-FisheyeVO datasets.}
    \label{fig:vo_results}
\end{figure*}

The quantitative results obtained using different evaluation metrics provide further insight into the performance gains. In particular, the reductions in Sq Rel and RMSE are generally more pronounced than the gains in accuracy under specific thresholds, suggesting that the proposed training strategy is particularly effective in reducing large depth errors. The accuracy under specific thresholds measures only the proportion of depth estimates that fall within predefined relative error ranges and therefore changes less when depth estimates already satisfying the corresponding thresholds are further refined. The qualitative results shown in Figs.~\ref{fig:qual_depth_kitti360}--\ref{fig:qual_depth_MIAS} further support the quantitative findings. Compared with the competing methods, the proposed framework produces depth maps with fewer visible artifacts and much clearer object boundaries, particularly in peripheral image regions affected by pronounced fisheye distortion. These improvements are consistently observed in both indoor and outdoor scenarios.

\begin{figure}[!tp]
\centering
\includegraphics[width=\columnwidth]{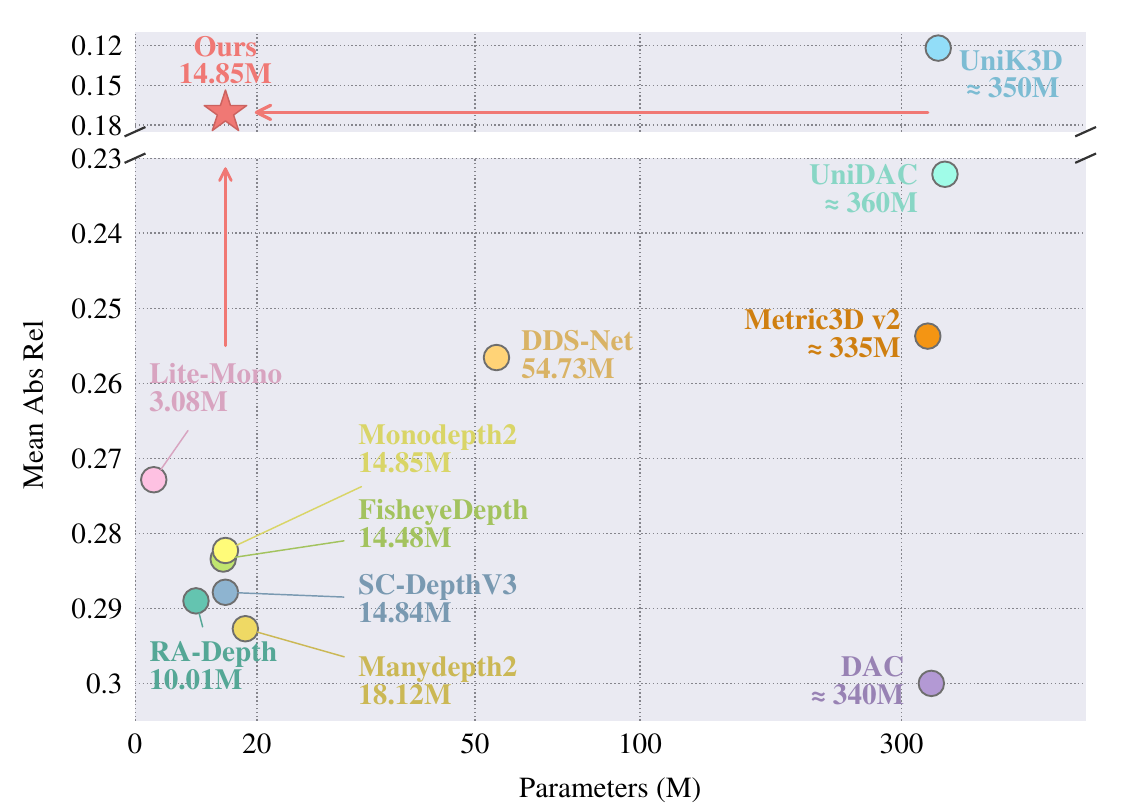}
\caption{Trade-off between depth estimation accuracy and model complexity among representative monocular fisheye depth estimation methods.}
\label{fig:params_absrel}
\end{figure}

Moreover, Table~\ref{tab:depth_results} presents additional comparisons with SoTA foundation models tailored for metric depth estimation, including Metric3D v2, UniK3D, DAC, and UniDAC. The proposed framework remains competitive across all datasets and achieves performance comparable to or better than several SoTA foundation models in multiple scenarios. Since the depth estimates generated by UniDAC are used only to supervise DepthNet pretraining and the pretrained DepthNet still exhibits relatively large depth estimation errors, the substantial improvement achieved through subsequent spherical bundle adjustment-guided unsupervised training further demonstrates the effectiveness of the proposed training strategy.

It should also be noted that the primary objective of this study is monocular fisheye visual odometry rather than standalone monocular fisheye depth estimation. To satisfy the real-time requirements of robotic applications, DepthNet adopts a lightweight encoder-decoder architecture instead of the more complex architectures used by SoTA foundation models. As illustrated in Fig.~\ref{fig:params_absrel}, the proposed method achieves a favorable trade-off between depth estimation accuracy and model complexity, attaining competitive accuracy with substantially fewer trainable parameters.

\subsection{Monocular Fisheye Visual Odometry Results}
\label{sec.exp_vo}
This subsection presents a comprehensive evaluation of monocular fisheye visual odometry performance. Estimated trajectory comparisons are shown in Fig.~\ref{fig:vo_results}, while quantitative results are reported in Tables~\ref{tab:kitti360_vo}--\ref{tab:mias_vo}. The proposed S$^2$D-FVO consistently achieves high trajectory estimation accuracy and camera tracking reliability across diverse scenarios and camera motion patterns, with particularly pronounced improvements over existing methods on challenging sequences.

Among the compared methods, the PoseNets adopted by the aforementioned learning-based methods~\cite{zhang2023litemono,godard2019digging,sun2023scdepthv3,he2022radepth,zhou2025manydepth2} are evaluated for monocular fisheye visual odometry. FisheyeDepth~\cite{zhao2024fisheyedepth} and DDS-Net~\cite{xu2026distortion} are excluded from the comparison, as their pose estimation settings differ from those of the evaluated PoseNets. For the geometry-based methods, the provided camera calibration parameters are adopted, and the corresponding projection and unprojection functions, together with their analytical Jacobians, are implemented accordingly. The same parameter settings are used for all sequences within each dataset. All compared methods are evaluated on both real-world and synthetic datasets.

\begin{table}[!tp]
    \centering
    \caption{Quantitative comparison of monocular visual odometry in terms of translational and rotational drift on the KITTI-360 and NIO-Fisheye datasets.}
    \label{tab:kitti_nio_et_er}
    \begin{tabular*}{\columnwidth}{
    @{\extracolsep{\fill}}
    ll
    cc
    @{}}
    \toprule
    \multicolumn{4}{c}{\textbf{KITTI-360}} \\
    \midrule
    \textbf{Type}
    & \textbf{Method}
    & $e_{t}$ (\%)
    & $e_{r}$ ($^\circ$/100\,m) \\
    \midrule
    
    \multirow{2}{*}{Geometry-based}
    & OV$^{2}$SLAM~\cite{fer2021ov2slam}
    & 57.691 & 6.865 \\
    
    & \textbf{Ours}
    & \textbf{4.451} & \textbf{0.288} \\
    
    \midrule
    
    \multirow{5}{*}{Learning-based}
    & Lite-Mono~\cite{zhang2023litemono}
    & 13.908 & 3.169 \\
    
    & Monodepth2~\cite{godard2019digging}
    & 14.015 & 2.448 \\
    
    & SC-DepthV3~\cite{sun2023scdepthv3}
    & 18.366 & 5.554 \\
    
    & RA-Depth~\cite{he2022radepth}
    & 32.179 & 7.395 \\
    
    & Manydepth2~\cite{zhou2025manydepth2}
    & 15.647 & 3.281 \\
    
    \midrule
    
    \multicolumn{4}{c}{\textbf{NIO-Fisheye}} \\
    \midrule
    \textbf{Type}
    & \textbf{Method}
    & $e_{t}$ (\%)
    & $e_{r}$ ($^\circ$/100\,m) \\
    \midrule
    
    \multirow{2}{*}{Geometry-based}
    & OV$^{2}$SLAM~\cite{fer2021ov2slam}
    & 47.028 & 5.057 \\
    
    & \textbf{Ours}
    & \textbf{11.183} & \textbf{1.140} \\
    
    \midrule
    
    \multirow{5}{*}{Learning-based}
    & Lite-Mono~\cite{zhang2023litemono}
    & 20.796 & 5.480 \\
    
    & Monodepth2~\cite{godard2019digging}
    & 19.722 & 3.735 \\
    
    & SC-DepthV3~\cite{sun2023scdepthv3}
    & 43.880 & 5.857 \\
    
    & RA-Depth~\cite{he2022radepth}
    & 21.209 & 5.610 \\
    
    & Manydepth2~\cite{zhou2025manydepth2}
    & 15.761 & 3.823 \\
    
    \bottomrule
    \end{tabular*}
\end{table}

The KITTI-360 and NIO-Fisheye datasets provide challenging real-world driving scenarios, enabling the evaluation of trajectory estimation accuracy and scale consistency over long sequences, as shown in Tables~\ref{tab:kitti360_vo} and~\ref{tab:nio_vo}. For the KITTI-360 dataset, the lateral viewing direction provides substantial parallax for nearby static features during forward vehicle motion but can also lead to large inter-frame displacements and shorter feature tracks. These challenges are further exacerbated in tunnels and along vegetation-lined roads, where poor illumination, repetitive textures, and unstable visual features can compromise feature matching. The quantitative results demonstrate that there exists a trade-off between tracking completeness and accumulated trajectory errors among the evaluated methods. ORB-SLAM3 and OpenVSLAM achieve average tracking coverage of only 12.6\% and 13.8\%, respectively, while CubemapSLAM, SVO, and HSO also exhibit frequent tracking interruptions. Although OV$^2$SLAM maintains full tracking coverage, its average ATE reaches 335.42\,m, indicating substantial accumulated drift. The learning-based methods provide pose estimates throughout all sequences but exhibit similarly large trajectory errors, with Lite-Mono achieving the lowest average ATE of 140.37\,m. In comparison, S$^2$D-FVO achieves the lowest ATE on all nine sequences, with an average of 31.47\,m, corresponding to reductions of 90.6\% and 77.6\% compared with OV$^2$SLAM and Lite-Mono, respectively. It further achieves the lowest RPE Rot on every sequence, with an average of 0.087$^\circ$, and the lowest average RPE Trans of 0.101\,m, while maintaining continuous camera tracking after pose initialization. These results demonstrate that S$^2$D-FVO substantially reduces both local pose errors and accumulated trajectory drift over long-distance driving sequences.

\begin{table*}[!tp]
    \centering
    \caption{Quantitative comparison of monocular visual odometry on the KITTI-360 dataset. The best results among fully tracked runs are shown in bold font. The symbol $\times$ indicates tracking failures, while $\dagger$ indicates partially tracked results with lower errors than the best fully tracked results. Errors for partially tracked runs are computed only over the successfully tracked trajectory segments.}
    \label{tab:kitti360_vo}
    
    \begin{tabular*}{\textwidth}{
    @{\extracolsep{\fill}}
    ll
    cccccccccc
    @{}
    }
    \toprule
    
    \textbf{Type}
    & \textbf{Method}
    & \textbf{Seq.\,00}
    & \textbf{Seq.\,02}
    & \textbf{Seq.\,03}
    & \textbf{Seq.\,04}
    & \textbf{Seq.\,05}
    & \textbf{Seq.\,06}
    & \textbf{Seq.\,07}
    & \textbf{Seq.\,09}
    & \textbf{Seq.\,10}
    & \textbf{Average} \\
    
    \midrule
    
    % =========================================================
    % ATE
    % =========================================================
    \multicolumn{12}{c}{\textbf{ATE (m)} $\downarrow$} \\
    \midrule
    
    \multirow{7}{*}{Geometry-based}
    & ORB-SLAM3~\cite{campos2021orbslam3}
    & $\times$ & $\times$ & $\times$ & $\times$ & $\times$
    & $\times$ & $\times$ & $\times$ & $\times$ & $\times$ \\
    
    & OpenVSLAM~\cite{sumikura2019openvslam}
    & $\times$ & $\times$ & $\times$ & $\times$ & $\times$ & $\times$ & $\times$ & $\times$ & $\times$ & $\times$ \\
    
    & CubemapSLAM~\cite{wang2018cubemapslam}
    & 308.178 & 354.433 & $\times$ & 869.561 & 287.203
    & $\times$ & $\times$ & $\times$ & $\times$
    & 454.844 \\
    
    & OV$^{2}$SLAM~\cite{fer2021ov2slam}
    & 242.963 & 289.912 & 140.861 & 661.126 & 209.020
    & 244.015 & 645.683 & 213.020 & 372.146
    & 335.416 \\
    
    & SVO~\cite{forster2017svo}
    & 336.580 & $\times$ & 215.743 & $\times$ & $\times$
    & 300.569 & 885.366 & 240.658 & 736.450
    & 452.561 \\
    
    & HSO~\cite{luo2022hso}
    & 311.182 & $\times$ & $\times$ & $\times$ & $\times$
    & $\times$ & $\times$ & $\times$ & $\times$
    & 311.182 \\
    
    & \textbf{Ours}
    & \textbf{19.945} & \textbf{35.034} & \textbf{3.052} & \textbf{53.509} & \textbf{23.577}
    & \textbf{30.906} & \textbf{47.578} & \textbf{33.546} & \textbf{36.062}
    & 31.468 \\

    \midrule
    
    \multirow{5}{*}{Learning-based}
    & Lite-Mono~\cite{zhang2023litemono}
    & 120.390 & 294.160 & 10.920 & 141.280 & 93.730
    & 218.420 & 67.960 & 250.580 & 65.850
    & 140.366 \\
    
    & Monodepth2~\cite{godard2019digging}
    & 136.000 & 228.500 & 20.690 & 205.400 & 77.180
    & 143.560 & 285.680 & 245.440 & 166.070
    & 167.613 \\
    
    & SC-DepthV3~\cite{sun2023scdepthv3}
    & 145.970 & 202.260 & 27.650 & 347.410 & 73.580
    & 227.670 & 363.430 & 198.040 & 191.230
    & 197.471 \\
    
    & RA-Depth~\cite{he2022radepth}
    & 284.390 & 253.500 & 101.050 & 541.770 & 166.110
    & 271.330 & 600.280 & 250.980 & 536.990
    & 334.044 \\
    
    & Manydepth2~\cite{zhou2025manydepth2}
    & 150.370 & 297.770 & 12.800 & 271.840 & 68.960
    & 174.620 & 178.280 & 259.000 & 52.080
    & 162.858 \\
    
    \midrule
    % =========================================================
    % RPE TRANS
    % =========================================================
    \multicolumn{12}{c}{\textbf{RPE Trans (m)} $\downarrow$} \\
    \midrule
    
    \multirow{7}{*}{Geometry-based}
    & ORB-SLAM3~\cite{campos2021orbslam3}
    & $\times$ & $\times$ & $\times$ & $\times$ & $\times$
    & $\times$ & $\times$ & $\times$ & $\times$ & $\times$ \\
    
    & OpenVSLAM~\cite{sumikura2019openvslam}
    & $\times$ & $\times$ & $\times$ & $\times$ & $\times$ & $\times$ & $\times$ & $\times$ & $\times$ & $\times$ \\
    
    & CubemapSLAM~\cite{wang2018cubemapslam}
    & 6.075 & 4.445 & $\times$ & 17.305 & 18.189
    & $\times$ & $\times$ & $\times$ & $\times$
    & 11.503 \\
    
    & OV$^{2}$SLAM~\cite{fer2021ov2slam}
    & 0.603 & 0.768 & 1.215 & 1.637 & 0.565
    & 0.700 & 2.980 & 0.426 & 0.984
    & 1.098 \\
    
    & SVO~\cite{forster2017svo}
    & 12.798 & $\times$ & 34.205 & $\times$ & $\times$
    & 11.201 & 42.854 & 6.119 & 23.047
    & 21.704 \\
    
    & HSO~\cite{luo2022hso}
    & 32.413 & $\times$ & $\times$ & $\times$ & $\times$
    & $\times$ & $\times$ & $\times$ & $\times$
    & 32.413 \\
    
    & \textbf{Ours}
    & \textbf{0.059} & \textbf{0.093} & \textbf{0.031} & 0.143 & 0.094
    & 0.063 & \textbf{0.097} & \textbf{0.118} & 0.213
    & 0.101 \\

    \midrule
    
    \multirow{5}{*}{Learning-based}
    & Lite-Mono~\cite{zhang2023litemono}
    & 0.065 & 0.167 & 0.057 & 0.101 & \textbf{0.054}
    & 0.448 & 0.171 & 0.657 & 0.490
    & 0.246 \\
    
    & Monodepth2~\cite{godard2019digging}
    & 0.181 & 0.167 & 0.062 & \textbf{0.056} & 0.074
    & \textbf{0.050} & 0.392 & 0.594 & 0.166
    & 0.194 \\
    
    & SC-DepthV3~\cite{sun2023scdepthv3}
    & 0.113 & 0.143 & 0.085 & 0.140 & 0.066
    & 0.424 & 0.344 & 0.425 & 0.498
    & 0.249 \\
    
    & RA-Depth~\cite{he2022radepth}
    & 0.464 & 0.421 & 0.176 & 0.193 & 0.224
    & 0.397 & 0.244 & 0.749 & 0.536
    & 0.378 \\
    
    & Manydepth2~\cite{zhou2025manydepth2}
    & 0.202 & 0.352 & 0.060 & 0.103 & 0.090
    & 0.149 & 0.260 & 0.734 & \textbf{0.138}
    & 0.232 \\
    
    \midrule
    % =========================================================
    % RPE ROT
    % =========================================================
    \multicolumn{12}{c}{\textbf{RPE Rot ($^\circ$)} $\downarrow$} \\
    \midrule
    
    \multirow{7}{*}{Geometry-based}
    & ORB-SLAM3~\cite{campos2021orbslam3}
    & $\times$ & $\times$ & $\times$ & $\times$ & $\times$
    & $\times$ & $\times$ & $\times$ & $\times$ & $\times$ \\
    
    & OpenVSLAM~\cite{sumikura2019openvslam}
    & $\times$ & $\times$ & $\times$ & $\times$ & $\times$ & $\times$ & $\times$ & $\times$ & $\times$ & $\times$ \\
    
    & CubemapSLAM~\cite{wang2018cubemapslam}
    & 1.077 & 3.190 & $\times$ & 3.411 & 4.830
    & $\times$ & $\times$ & $\times$ & $\times$
    & 3.127 \\
    
    & OV$^{2}$SLAM~\cite{fer2021ov2slam}
    & 0.419 & 0.429 & 19.661 & 3.563 & 0.538
    & 0.553 & 3.403 & 0.265 & 1.406
    & 3.360 \\
    
    & SVO~\cite{forster2017svo}
    & 7.765 & $\times$ & 6.748 & $\times$ & $\times$
    & 8.477 & 5.390 & 7.629 & 6.958
    & 7.161 \\
    
    & HSO~\cite{luo2022hso}
    & 2.348 & $\times$ & $\times$ & $\times$ & $\times$
    & $\times$ & $\times$ & $\times$ & $\times$
    & 2.348 \\
    
    & \textbf{Ours}
    & \textbf{0.039} & \textbf{0.136} & \textbf{0.060} & \textbf{0.159} & \textbf{0.057}
    & \textbf{0.060} & \textbf{0.092} & \textbf{0.076} & \textbf{0.108}
    & 0.087 \\

    \midrule
    
    \multirow{5}{*}{Learning-based}
    & Lite-Mono~\cite{zhang2023litemono}
    & 0.132 & 0.160 & 0.082 & 0.724 & 0.134
    & 0.123 & 0.108 & 0.119 & 0.240
    & 0.202 \\
    
    & Monodepth2~\cite{godard2019digging}
    & 0.505 & 0.145 & 0.069 & 0.705 & 0.125
    & 0.101 & 0.326 & 0.200 & 0.876
    & 0.339 \\
    
    & SC-DepthV3~\cite{sun2023scdepthv3}
    & 0.468 & 0.184 & 0.113 & 0.739 & 0.179
    & 0.175 & 0.194 & 0.250 & 0.616
    & 0.324 \\
    
    & RA-Depth~\cite{he2022radepth}
    & 0.415 & 0.188 & 0.166 & 0.710 & 0.167
    & 0.168 & 0.342 & 0.194 & 0.707
    & 0.340 \\
    
    & Manydepth2~\cite{zhou2025manydepth2}
    & 0.139 & 0.166 & 0.101 & 0.721 & 0.142
    & 0.128 & 0.144 & 0.139 & 0.402
    & 0.231 \\
    
    \midrule
    % =========================================================
    % TRACKING COVERAGE
    % =========================================================
    \multicolumn{12}{c}{\textbf{Trck (\%)} $\uparrow$} \\
    \midrule
    
    \multirow{7}{*}{Geometry-based}
    & ORB-SLAM3~\cite{campos2021orbslam3}
    & 4.8 & 2.1 & 41.5 & 1.9 & 10.0
    & 8.7 & 6.2 & 10.5 & 27.8 & 12.6 \\
    
    & OpenVSLAM~\cite{sumikura2019openvslam}
    & 23.1 & 21.9 & 27.7 & 1.3 & 10.1 & 0.9 & 2.3 & 8.8 & 27.8 & 13.8 \\
    
    & CubemapSLAM~\cite{wang2018cubemapslam}
    & 88.7 & 84.5 & 3.4 & 72.8 & 61.1
    & 0.9 & 3.5 & 28.2 & 0.7
    & 38.2 \\
    
    & OV$^{2}$SLAM~\cite{fer2021ov2slam}
    & 100.0 & 100.0 & 100.0 & 100.0 & 100.0
    & 100.0 & 100.0 & 100.0 & 100.0
    & 100.0 \\
    
    & SVO~\cite{forster2017svo}
    & 52.4 & 39.7 & 63.2 & 44.1 & 45.8
    & 50.9 & 76.7 & 67.8 & 57.3
    & 55.3 \\
    
    & HSO~\cite{luo2022hso}
    & 86.7 & 17.0 & 37.9 & 1.8 & 25.4
    & 1.8 & 6.3 & 0.9 & 32.0
    & 23.3 \\
    
    & \textbf{Ours}
    & 100.0 & 100.0 & 100.0 & 100.0 & 100.0
    & 100.0 & 100.0 & 100.0 & 100.0
    & 100.0 \\
    
    \midrule
    
    \multirow{5}{*}{Learning-based}
    & Lite-Mono~\cite{zhang2023litemono}
    & 100.0 & 100.0 & 100.0 & 100.0 & 100.0
    & 100.0 & 100.0 & 100.0 & 100.0
    & 100.0 \\
    
    & Monodepth2~\cite{godard2019digging}
    & 100.0 & 100.0 & 100.0 & 100.0 & 100.0
    & 100.0 & 100.0 & 100.0 & 100.0
    & 100.0 \\
    
    & SC-DepthV3~\cite{sun2023scdepthv3}
    & 100.0 & 100.0 & 100.0 & 100.0 & 100.0
    & 100.0 & 100.0 & 100.0 & 100.0
    & 100.0 \\
    
    & RA-Depth~\cite{he2022radepth}
    & 100.0 & 100.0 & 100.0 & 100.0 & 100.0
    & 100.0 & 100.0 & 100.0 & 100.0
    & 100.0 \\
    
    & Manydepth2~\cite{zhou2025manydepth2}
    & 100.0 & 100.0 & 100.0 & 100.0 & 100.0
    & 100.0 & 100.0 & 100.0 & 100.0
    & 100.0 \\
    
    \bottomrule
    \end{tabular*}
    \end{table*}

\begin{table*}[!tp]
    \centering
    \caption{Quantitative comparison of monocular visual odometry on the NIO-Fisheye dataset. The best results among fully tracked runs are shown in bold font. The symbol $\times$ indicates tracking failures, while $\dagger$ indicates partially tracked results with lower errors than the best fully tracked results. Errors for partially tracked runs are computed only over the successfully tracked trajectory segments.}
    \label{tab:nio_vo}
    
    \begin{tabular*}{\textwidth}{
    @{\extracolsep{\fill}}
    ll
    cccccc
    @{}
    }
    \toprule
    
    \textbf{Type}
    & \textbf{Method}
    & \textbf{Seq.\,00}
    & \textbf{Seq.\,01}
    & \textbf{Seq.\,02}
    & \textbf{Seq.\,03}
    & \textbf{Seq.\,04}
    & \textbf{Average} \\
    
    \midrule
    
    % =========================================================
    % ATE
    % =========================================================
    \multicolumn{8}{c}{\textbf{ATE (m)} $\downarrow$} \\
    \midrule
    
    \multirow{7}{*}{Geometry-based}
    & ORB-SLAM3~\cite{campos2021orbslam3}
    & $\times$ & 936.253 & $\times$ & $\times$ & $\times$ & 936.253 \\
    
    & OpenVSLAM~\cite{sumikura2019openvslam}
    & $\times$ & $\times$ & 20.570 & $\times$ & $\times$ & 20.570 \\
    
    & CubemapSLAM~\cite{wang2018cubemapslam}
    & $\times$ & 1352.358 & 222.228 & 369.410 & $\times$
    & 647.999 \\
    
    & OV$^{2}$SLAM~\cite{fer2021ov2slam}
    & 39.003 & 232.083 & 224.715 & 207.867 & 2182.091
    & 577.152 \\
    
    & SVO~\cite{forster2017svo}
    & 225.474 & 1275.617 & 215.834 & 306.793 & 2072.314
    & 819.206 \\
    
    & HSO~\cite{luo2022hso}
    & \textbf{13.037} & 767.302 & \textbf{16.717} & 63.287 & 588.975
    & 289.864 \\
    
    & \textbf{Ours}
    & 16.585 & 178.467 & 28.030 & \textbf{10.129} & \textbf{232.354}
    & 93.113 \\

    % =========================================================
    % RPE TRANS
    % =========================================================
    \midrule
    
    \multirow{5}{*}{Learning-based}
    & Lite-Mono~\cite{zhang2023litemono}
    & 61.067 & 359.522 & 91.147 & 149.415 & 441.247
    & 220.480 \\
    
    & Monodepth2~\cite{godard2019digging}
    & 57.421 & \textbf{116.867} & 42.183 & 121.951 & 1665.746
    & 400.834 \\
    
    & SC-DepthV3~\cite{sun2023scdepthv3}
    & 172.173 & 900.718 & 48.323 & 65.569 & 1236.671
    & 484.691 \\
    
    & RA-Depth~\cite{he2022radepth}
    & 68.622 & 430.269 & 129.757 & 79.704 & 1495.568
    & 440.784 \\
    
    & Manydepth2~\cite{zhou2025manydepth2}
    & 42.883 & 135.723 & 62.623 & 60.492 & 296.079
    & 119.560 \\
    
    \midrule
    \multicolumn{8}{c}{\textbf{RPE Trans (m)} $\downarrow$} \\
    \midrule
    
    \multirow{7}{*}{Geometry-based}
    & ORB-SLAM3~\cite{campos2021orbslam3}
    & $\times$ & 2.597 & $\times$ & $\times$ & $\times$ & 2.597 \\
    
    & OpenVSLAM~\cite{sumikura2019openvslam}
    & $\times$ & $\times$ & 1.095 & $\times$ & $\times$ & 1.095 \\
    
    & CubemapSLAM~\cite{wang2018cubemapslam}
    & $\times$ & 14.194 & 17.434 & 41.579 & $\times$
    & 24.402 \\
    
    & OV$^{2}$SLAM~\cite{fer2021ov2slam}
    & 0.520 & 0.875 & 0.423 & 0.273 & 7.712
    & 1.961 \\
    
    & SVO~\cite{forster2017svo}
    & 1.589 & 32.005 & 8.588 & 1.537 & 34.060
    & 15.556 \\
    
    & HSO~\cite{luo2022hso}
    & 0.208 & 1.212 & 0.148 & 0.178 & 1.993
    & 0.748 \\
    
    & \textbf{Ours}
    & 0.597 & 1.057 & 0.213 & \textbf{0.175} & 0.620
    & 0.532 \\

    % =========================================================
    % RPE ROT
    % =========================================================
    \midrule
    
    \multirow{5}{*}{Learning-based}
    & Lite-Mono~\cite{zhang2023litemono}
    & \textbf{0.162} & 0.324 & 0.233 & 0.266 & 0.498
    & 0.297 \\
    
    & Monodepth2~\cite{godard2019digging}
    & 0.279 & 0.300 & \textbf{0.117} & 0.259 & 0.653
    & 0.322 \\
    
    & SC-DepthV3~\cite{sun2023scdepthv3}
    & 1.161 & 2.919 & 0.217 & 0.506 & 2.514
    & 1.463 \\
    
    & RA-Depth~\cite{he2022radepth}
    & 0.211 & 0.344 & 0.404 & 0.216 & 0.487
    & 0.332 \\
    
    & Manydepth2~\cite{zhou2025manydepth2}
    & 0.187 & \textbf{0.296} & 0.142 & 0.216 & \textbf{0.441}
    & 0.256 \\
    
    \midrule
    \multicolumn{8}{c}{\textbf{RPE Rot ($^\circ$)} $\downarrow$} \\
    \midrule
    
    \multirow{7}{*}{Geometry-based}
    & ORB-SLAM3~\cite{campos2021orbslam3}
    & $\times$ & 0.092 & $\times$ & $\times$ & $\times$ & 0.092 \\
    
    & OpenVSLAM~\cite{sumikura2019openvslam}
    & $\times$ & $\times$ & \textbf{0.049} & $\times$ & $\times$ & 0.049 \\
    
    & CubemapSLAM~\cite{wang2018cubemapslam}
    & $\times$ & 0.217 & 6.164 & 1.353 & $\times$
    & 2.578 \\
    
    & OV$^{2}$SLAM~\cite{fer2021ov2slam}
    & 0.189 & \textbf{0.075} & 6.667 & 0.276 & 2.446
    & 1.931 \\
    
    & SVO~\cite{forster2017svo}
    & 2.048 & 2.164 & 4.525 & 2.833 & 4.650
    & 3.244 \\
    
    & HSO~\cite{luo2022hso}
    & 0.152 & 0.314 & 0.172 & 0.159 & 0.221
    & 0.204 \\
    
    & \textbf{Ours}
    & 0.589 & 1.175 & 0.481 & 0.368 & 0.787
    & 0.680 \\

    % =========================================================
    % TRACKING COVERAGE
    % =========================================================
    \midrule
    
    \multirow{5}{*}{Learning-based}
    & Lite-Mono~\cite{zhang2023litemono}
    & 0.161 & 0.125 & 0.207 & 0.319 & 0.444
    & 0.251 \\
    
    & Monodepth2~\cite{godard2019digging}
    & \textbf{0.127} & 0.094 & 0.208 & \textbf{0.116} & 0.298
    & 0.169 \\
    
    & SC-DepthV3~\cite{sun2023scdepthv3}
    & 0.207 & 0.197 & 0.221 & 0.159 & \textbf{0.210}
    & 0.199 \\
    
    & RA-Depth~\cite{he2022radepth}
    & 0.171 & 0.172 & 0.383 & 0.210 & 0.252
    & 0.238 \\
    
    & Manydepth2~\cite{zhou2025manydepth2}
    & 0.183 & 0.116 & 0.188 & 0.134 & 0.442
    & 0.213 \\
    
    \midrule
    \multicolumn{8}{c}{\textbf{Trck (\%)} $\uparrow$} \\
    \midrule
    
    \multirow{7}{*}{Geometry-based}
    & ORB-SLAM3~\cite{campos2021orbslam3}
    & 47.3 & 72.8 & 42.9 & 42.5 & 12.3 & 43.6 \\
    
    & OpenVSLAM~\cite{sumikura2019openvslam}
    & 29.5 & 30.5 & 100.0 & 42.1 & 13.6 & 43.1 \\
    
    & CubemapSLAM~\cite{wang2018cubemapslam}
    & 42.4 & 100.0 & 78.6 & 78.6 & 13.2
    & 62.6 \\
    
    & OV$^{2}$SLAM~\cite{fer2021ov2slam}
    & 100.0 & 100.0 & 100.0 & 100.0 & 100.0
    & 100.0 \\
    
    & SVO~\cite{forster2017svo}
    & 100.0 & 99.1 & 98.0 & 99.1 & 97.1
    & 98.7 \\
    
    & HSO~\cite{luo2022hso}
    & 100.0 & 100.0 & 100.0 & 100.0 & 67.6
    & 93.5 \\
    
    & \textbf{Ours}
    & 100.0 & 100.0 & 100.0 & 100.0 & 100.0
    & 100.0 \\
    
    \midrule
    
    \multirow{5}{*}{Learning-based}
    & Lite-Mono~\cite{zhang2023litemono}
    & 100.0 & 100.0 & 100.0 & 100.0 & 100.0 
    & 100.0 \\
    
    & Monodepth2~\cite{godard2019digging}
    & 100.0 & 100.0 & 100.0 & 100.0 & 100.0 
    & 100.0 \\
    
    & SC-DepthV3~\cite{sun2023scdepthv3}
    & 100.0 & 100.0 & 100.0 & 100.0 & 100.0 
    & 100.0 \\
    
    & RA-Depth~\cite{he2022radepth}
    & 100.0 & 100.0 & 100.0 & 100.0 & 100.0 
    & 100.0 \\
    
    & Manydepth2~\cite{zhou2025manydepth2}
    & 100.0 & 100.0 & 100.0 & 100.0 & 100.0 
    & 100.0 \\
    
    \bottomrule
    \end{tabular*}
    \end{table*}

\begin{table*}[!tp]
    \centering
    \caption{Quantitative comparison of monocular visual odometry on the TUM-VI dataset. The best results among fully tracked runs are shown in bold font. The symbol $\times$ indicates tracking failures, while $\dagger$ indicates partially tracked results with lower errors than the best fully tracked results. Errors for partially tracked runs are computed only over the successfully tracked trajectory segments.}
    \label{tab:tumvi_vo}
    
    % \settablefont
    % \setlength{\tabcolsep}{2.4pt}
    % \renewcommand{\arraystretch}{1.10}
    
    \begin{tabular*}{\textwidth}{
    @{\extracolsep{\fill}}
    l
    ccccccc
    @{}
    }
    \toprule
    
    \textbf{Method}
    & \textbf{Room\,1}
    & \textbf{Room\,2}
    & \textbf{Room\,3}
    & \textbf{Room\,4}
    & \textbf{Room\,5}
    & \textbf{Room\,6}
    & \textbf{Average} \\
    
    \midrule
    
    % =========================================================
    % ATE RMSE
    % =========================================================
    \multicolumn{8}{c}{\textbf{ATE (m)} $\downarrow$} \\
    \midrule
    
    ORB-SLAM3~\cite{campos2021orbslam3}
    & \textbf{0.013} & 0.014 & 0.157 & 0.010 & 0.104 & 0.010 & 0.051 \\
    
    OpenVSLAM~\cite{sumikura2019openvslam}
    & 0.013 & 0.014 & 0.014 & 0.014 & 0.014 & 0.017 & 0.014 \\
    
    CubemapSLAM~\cite{wang2018cubemapslam}
    & 0.042 & \textbf{0.007} & \textbf{0.009} & \textbf{0.006} & 0.174 & \textbf{0.004} & 0.040 \\
    
    OV$^{2}$SLAM~\cite{fer2021ov2slam}
    & 1.003 & 0.875 & 1.307 & 0.988 & 0.868 & 0.216 & 0.876 \\
    
    SVO~\cite{forster2017svo}
    & 0.912 & 1.235 & 1.279 & 0.987 & 0.869 & 0.892 & 1.029 \\
    
    HSO~\cite{luo2022hso}
    & $\times$ & 0.033 & 1.206 & 0.031 & $\times$ & $\times$ & 0.423 \\
    
    \textbf{Ours}
    & 0.016 & 0.013 & 0.020 & \textbf{0.006} & \textbf{0.010} & 0.005 & 0.012 \\
    
    \midrule
    
    % =========================================================
    % RPE_t RMSE
    % =========================================================
    \multicolumn{8}{c}{\textbf{RPE Trans (m)} $\downarrow$} \\
    \midrule
    
    ORB-SLAM3~\cite{campos2021orbslam3}
    & 0.008 & 0.010 & 0.062 & 0.006 & 0.017 & 0.005 & 0.018 \\
    
    OpenVSLAM~\cite{sumikura2019openvslam}
    & 0.011 & 0.014 & 0.017 & 0.012 & 0.011 & 0.011 & 0.013 \\
    
    CubemapSLAM~\cite{wang2018cubemapslam}
    & 0.012 & \textbf{0.008} & \textbf{0.011} & \textbf{0.004} & 0.050 & \textbf{0.003} & 0.015 \\
    
    OV$^{2}$SLAM~\cite{fer2021ov2slam}
    & 0.065 & 0.049 & 0.100 & 0.034 & 0.055 & 0.012 & 0.053 \\
    
    SVO~\cite{forster2017svo}
    & 0.126 & 0.130 & 0.193 & 0.155 & 0.171 & 0.055 & 0.138 \\
    
    HSO~\cite{luo2022hso}
    & $\times$ & 0.009 & 0.280 & 0.005 & $\times$ & $\times$ & 0.098 \\
    
    \textbf{Ours}
    & \textbf{0.006} & \textbf{0.008} & 0.012 & \textbf{0.004} & \textbf{0.005} & 0.004 & 0.007 \\
    
    \midrule
    
    % =========================================================
    % RPE_r RMSE
    % =========================================================
    \multicolumn{8}{c}{\textbf{RPE Rot ($\circ$)} $\downarrow$} \\ 
    \midrule
    
    ORB-SLAM3~\cite{campos2021orbslam3}
    & 0.489 & 0.714 & 0.775 & 0.480 & 0.530 & 0.345 & 0.555 \\
    
    OpenVSLAM~\cite{sumikura2019openvslam}
    & 0.538 & 0.811 & 0.866 & 0.614 & 0.495 & 0.402 & 0.621 \\
    
    CubemapSLAM~\cite{wang2018cubemapslam}
    & 0.552 & \textbf{0.709} & \textbf{0.716} & \textbf{0.469} & 0.449 & \textbf{0.337} & 0.539 \\
    
    OV$^{2}$SLAM~\cite{fer2021ov2slam}
    & 2.257 & 1.002 & 3.001 & 1.263 & 3.957 & 0.446 & 1.988 \\
    
    SVO~\cite{forster2017svo}
    & 6.159 & 4.037 & 8.170 & 10.887 & 12.220 & 3.162 & 7.439 \\
    
    HSO~\cite{luo2022hso}
    & $\times$ & 0.712 & 4.396 & 0.480 & $\times$ & $\times$ & 1.863 \\
    
    \textbf{Ours}
    & \textbf{0.482} & 0.710 & 0.744 & 0.477 & \textbf{0.423} & 0.343 & 0.530 \\
    
    \midrule
    
    % =========================================================
    % TRACKING COVERAGE
    % =========================================================
    \multicolumn{8}{c}{\textbf{Trck (\%)} $\uparrow$} \\
    \midrule
    
    ORB-SLAM3~\cite{campos2021orbslam3}
    & 100.0 & 100.0 & 100.0 & 100.0 & 100.0 & 100.0 & 100.0 \\
    
    OpenVSLAM~\cite{sumikura2019openvslam}
    & 95.9 & 98.0 & 100.0 & 96.8 & 100.0 & 99.5 & 98.4 \\
    
    CubemapSLAM~\cite{wang2018cubemapslam}
    & 100.0 & 100.0 & 100.0 & 100.0 & 100.0 & 100.0 & 100.0 \\
    
    OV$^{2}$SLAM~\cite{fer2021ov2slam}
    & 100.0 & 100.0 & 100.0 & 100.0 & 100.0 & 100.0 & 100.0 \\
    
    SVO~\cite{forster2017svo}
    & 72.9 & 84.1 & 75.6 & 56.5 & 58.5 & 88.8 & 72.7 \\
    
    HSO~\cite{luo2022hso}
    & 23.2 & 100.0 & 63.3 & 100.0 & 23.0 & 20.9 & 55.1 \\
    
    \textbf{Ours}
    & 100.0 & 100.0 & 100.0 & 100.0 & 100.0 & 100.0 & 100.0 \\
    
    \bottomrule
    \end{tabular*}
    \end{table*}

\begin{table*}[!tp]
    \centering
    \caption{Quantitative comparison of monocular visual odometry on the House and RetroOffice scenarios of the TartanAir V2 dataset. Sequences P000--P007 and P000--P005
are denoted as H0--H7 and R0--R5 in the House
and RetroOffice scenarios, respectively. The best results among fully tracked runs are shown in bold font. The symbol $\times$ indicates tracking failures, while $\dagger$ indicates partially tracked results with lower errors than the best fully tracked results. Errors for partially tracked runs are computed only over the successfully tracked trajectory segments.}
    \label{tab:tartan_vo_indoor}

    \begin{tabular*}{\textwidth}{
    @{\extracolsep{\fill}}
    ll
    *{9}{c}
    *{7}{c}
    @{}
    }
    \toprule

    \multirow{2}{*}{\textbf{Type}}
    & \multirow{2}{*}{\textbf{Method}}
    & \multicolumn{9}{c}{\textbf{House Subset}}
    & \multicolumn{7}{c}{\textbf{RetroOffice Subset}} \\
    \cmidrule(lr){3-11}
    \cmidrule(lr){12-18}

    &
    & \textbf{H0}
    & \textbf{H1}
    & \textbf{H2}
    & \textbf{H3}
    & \textbf{H4}
    & \textbf{H5}
    & \textbf{H6}
    & \textbf{H7}
    & \textbf{Average}
    & \textbf{R0}
    & \textbf{R1}
    & \textbf{R2}
    & \textbf{R3}
    & \textbf{R4}
    & \textbf{R5}
    & \textbf{Average} \\

    \midrule

    % =========================================================
    % ATE
    % =========================================================
    \multicolumn{18}{c}{\textbf{ATE (m)} $\downarrow$} \\
    \midrule

    \multirow{7}{*}{Geometry-based}
    & ORB-SLAM3~\cite{campos2021orbslam3}
    & $\times$ & $\times$ & 1.653 & $\times$
    & $\times$ & $\times$ & 0.012 & 0.185$^{\dagger}$
    & 0.617
    & 0.026 & 0.006$^{\dagger}$ & $\times$ & $\times$
    & 0.011 & $\times$
    & 0.014 \\

    & OpenVSLAM~\cite{sumikura2019openvslam}
    & $\times$ & $\times$ & $\times$ & $\times$
    & $\times$ & $\times$ & 0.055 & $\times$
    & 0.055
    & 0.035 & \textbf{0.016} & 0.012 & 0.013
    & 0.016 & 0.062
    & 0.026 \\

    & CubemapSLAM~\cite{wang2018cubemapslam}
    & $\times$ & 0.017$^{\dagger}$ & \textbf{0.058} & $\times$
    & 0.027$^{\dagger}$ & \textbf{0.026} & 0.012 & $\times$
    & 0.028
    & \textbf{0.006} & 0.003$^{\dagger}$ & 0.005$^{\dagger}$ & \textbf{0.004}
    & 0.004 & \textbf{0.005}
    & 0.005 \\

    & OV$^{2}$SLAM~\cite{fer2021ov2slam}
    & 2.565 & 2.757 & 2.884 & 1.810
    & 3.895 & 2.139 & 1.677 & 2.846
    & 2.572
    & 0.616 & 0.048 & 0.023 & 0.020
    & 0.012 & 0.115
    & 0.139 \\

    & SVO~\cite{forster2017svo}
    & 3.980 & 3.630 & 3.261 & 2.964
    & 3.111 & 3.377 & 2.459 & 3.505
    & 3.286
    & 0.735 & 0.603 & 0.675 & 0.281
    & 0.009 & 0.739
    & 0.507 \\

    & HSO~\cite{luo2022hso}
    & $\times$ & $\times$ & $\times$ & $\times$
    & $\times$ & $\times$ & $\times$ & $\times$
    & $\times$
    & $\times$ & 0.237 & $\times$ & $\times$
    & 0.155 & $\times$
    & 0.196 \\

    & \textbf{Ours}
    & \textbf{0.106} & \textbf{0.100} & 0.067 & \textbf{0.029}
    & \textbf{0.045} & 0.041 & \textbf{0.011} & \textbf{0.194}
    & 0.074
    & \textbf{0.006} & 0.160 & \textbf{0.007} & \textbf{0.004}
    & \textbf{0.004} & 0.006
    & 0.031 \\

    \midrule

    \multirow{5}{*}{Learning-based}
    & Lite-Mono~\cite{zhang2023litemono}
    & 1.750 & 2.497 & 1.682 & 0.978
    & 1.027 & 1.082 & 0.914 & 2.748
    & 1.585
    & 0.371 & 0.197 & 0.355 & 0.332
    & 0.171 & 0.371
    & 0.300 \\

    & Monodepth2~\cite{godard2019digging}
    & 1.464 & 1.724 & 1.332 & 1.030
    & 0.965 & 1.182 & 0.942 & 1.508
    & 1.268
    & 0.328 & 0.183 & 0.480 & 0.259
    & 0.173 & 0.502
    & 0.321 \\

    & SC-DepthV3~\cite{sun2023scdepthv3}
    & 1.659 & 1.699 & 1.711 & 1.448
    & 1.094 & 1.200 & 0.882 & 2.354
    & 1.506
    & 0.483 & 0.258 & 0.578 & 0.263
    & 0.204 & 0.388
    & 0.362 \\

    & RA-Depth~\cite{he2022radepth}
    & 1.867 & 2.491 & 1.683 & 0.824
    & 1.199 & 1.162 & 0.631 & 1.835
    & 1.462
    & 0.365 & 0.404 & 0.467 & 0.346
    & 0.204 & 0.388
    & 0.362 \\

    & Manydepth2~\cite{zhou2025manydepth2}
    & 1.936 & 1.149 & 1.607 & 1.747
    & 1.662 & 1.270 & 1.503 & 2.251
    & 1.641
    & 0.199 & 0.265 & 0.533 & 0.401
    & 0.274 & 0.448
    & 0.353 \\

    \midrule
    % =========================================================
    % RPE TRANS
    % =========================================================
    \multicolumn{18}{c}{\textbf{RPE Trans (m)} $\downarrow$} \\
    \midrule

    \multirow{7}{*}{Geometry-based}
    & ORB-SLAM3~\cite{campos2021orbslam3}
    & $\times$ & $\times$ & 0.081 & $\times$
    & $\times$ & $\times$ & 0.009 & 0.023
    & 0.038
    & 0.029 & 0.006$^{\dagger}$ & $\times$ & $\times$
    & 0.012 & $\times$
    & 0.016 \\

    & OpenVSLAM~\cite{sumikura2019openvslam}
    & $\times$ & $\times$ & $\times$ & $\times$
    & $\times$ & $\times$ & 0.014 & $\times$
    & 0.014
    & 0.013 & \textbf{0.013} & 0.010 & 0.008
    & 0.007 & 0.014
    & 0.011 \\

    & CubemapSLAM~\cite{wang2018cubemapslam}
    & $\times$ & 0.010 & \textbf{0.006} & $\times$
    & 0.007 & \textbf{0.006} & 0.004 & $\times$
    & 0.007
    & \textbf{0.002} & 0.002$^{\dagger}$ & 0.002$^{\dagger}$ & \textbf{0.002}
    & 0.002 & \textbf{0.003}
    & 0.002 \\

    & OV$^{2}$SLAM~\cite{fer2021ov2slam}
    & 0.105 & 0.129 & 0.100 & 0.089
    & 0.179 & 0.084 & 0.073 & 0.117
    & 0.109
    & 0.040 & 0.037 & 0.007 & 0.013
    & 0.009 & 0.012
    & 0.020 \\

    & SVO~\cite{forster2017svo}
    & 0.429 & 0.184 & 0.160 & 0.333
    & 0.423 & 0.228 & 0.285 & 0.167
    & 0.276
    & 0.099 & 0.065 & 0.141 & 0.049
    & 0.004 & 0.066
    & 0.071 \\

    & HSO~\cite{luo2022hso}
    & $\times$ & $\times$ & $\times$ & $\times$
    & $\times$ & $\times$ & $\times$ & $\times$
    & $\times$
    & $\times$ & 0.077 & $\times$ & $\times$
    & 0.069 & $\times$
    & 0.073 \\

    & \textbf{Ours}
    & \textbf{0.016} & \textbf{0.010} & 0.009 & \textbf{0.021}
    & \textbf{0.005} & 0.010 & \textbf{0.003} & \textbf{0.016}
    & 0.011
    & \textbf{0.002} & 0.020 & \textbf{0.004} & \textbf{0.002}
    & \textbf{0.002} & \textbf{0.003}
    & 0.006 \\

    \midrule

    \multirow{5}{*}{Learning-based}
    & Lite-Mono~\cite{zhang2023litemono}
    & 0.056 & 0.072 & 0.051 & 0.043
    & 0.059 & 0.043 & 0.047 & 0.094
    & 0.058
    & 0.034 & 0.031 & 0.036 & 0.024
    & 0.040 & 0.048
    & 0.036 \\

    & Monodepth2~\cite{godard2019digging}
    & 0.052 & 0.060 & 0.044 & 0.043
    & 0.056 & 0.042 & 0.045 & 0.062
    & 0.050
    & 0.029 & 0.029 & 0.037 & 0.024
    & 0.041 & 0.043
    & 0.034 \\

    & SC-DepthV3~\cite{sun2023scdepthv3}
    & 0.047 & 0.063 & 0.043 & 0.054
    & 0.055 & 0.041 & 0.043 & 0.083
    & 0.054
    & 0.029 & 0.030 & 0.050 & 0.021
    & 0.038 & 0.039
    & 0.035 \\

    & RA-Depth~\cite{he2022radepth}
    & 0.051 & 0.067 & 0.046 & 0.041
    & 0.060 & 0.041 & 0.048 & 0.077
    & 0.054
    & 0.037 & 0.030 & 0.043 & 0.030
    & 0.044 & 0.045
    & 0.038 \\

    & Manydepth2~\cite{zhou2025manydepth2}
    & 0.057 & 0.060 & 0.054 & 0.055
    & 0.075 & 0.046 & 0.064 & 0.084
    & 0.062
    & 0.030 & 0.033 & 0.047 & 0.029
    & 0.044 & 0.047
    & 0.038 \\

    \midrule
    % =========================================================
    % RPE ROT
    % =========================================================
    \multicolumn{18}{c}{\textbf{RPE Rot ($^\circ$)} $\downarrow$} \\
    \midrule

    \multirow{7}{*}{Geometry-based}
    & ORB-SLAM3~\cite{campos2021orbslam3}
    & $\times$ & $\times$ & 0.537 & $\times$
    & $\times$ & $\times$ & 0.111 & 0.358
    & 0.335
    & 0.132 & 0.126$^{\dagger}$ & $\times$ & $\times$
    & 0.172 & $\times$
    & 0.143 \\

    & OpenVSLAM~\cite{sumikura2019openvslam}
    & $\times$ & $\times$ & $\times$ & $\times$
    & $\times$ & $\times$ & 0.212 & $\times$
    & 0.212
    & 0.268 & 0.438 & 0.187 & 0.215
    & 0.281 & 0.413
    & 0.300 \\

    & CubemapSLAM~\cite{wang2018cubemapslam}
    & $\times$ & 0.114$^{\dagger}$ & \textbf{0.083} & $\times$
    & 0.092 & 0.164 & 0.070 & $\times$
    & 0.105
    & 0.068 & 0.064$^{\dagger}$ & 0.051$^{\dagger}$ & 0.054
    & 0.061 & 0.074
    & 0.062 \\

    & OV$^{2}$SLAM~\cite{fer2021ov2slam}
    & 1.272 & 1.763 & 1.309 & 0.784
    & 3.588 & 1.575 & 1.514 & 1.529
    & 1.667
    & 0.944 & 1.350 & 0.145 & 0.623
    & 0.266 & 0.296
    & 0.604 \\

    & SVO~\cite{forster2017svo}
    & 6.192 & 6.123 & 3.446 & 10.686
    & 10.923 & 8.088 & 15.999 & 5.787
    & 8.406
    & 5.545 & 1.529 & 3.360 & 1.100
    & 0.091 & 2.828
    & 2.409 \\

    & HSO~\cite{luo2022hso}
    & $\times$ & $\times$ & $\times$ & $\times$
    & $\times$ & $\times$ & $\times$ & $\times$
    & $\times$
    & $\times$ & 2.170 & $\times$ & $\times$
    & 2.222 & $\times$
    & 2.196 \\

    & \textbf{Ours}
    & \textbf{0.373} & \textbf{0.300} & 0.465 & \textbf{0.540}
    & \textbf{0.076} & \textbf{0.120} & \textbf{0.049} & \textbf{0.255}
    & 0.272
    & \textbf{0.048} & \textbf{0.432} & \textbf{0.052} & \textbf{0.044}
    & \textbf{0.054} & \textbf{0.061}
    & 0.115 \\

    \midrule

    \multirow{5}{*}{Learning-based}
    & Lite-Mono~\cite{zhang2023litemono}
    & 1.187 & 1.120 & 1.106 & 1.006
    & 1.176 & 0.978 & 0.907 & 1.210
    & 1.086
    & 0.725 & 0.835 & 0.871 & 0.669
    & 0.906 & 0.944
    & 0.825 \\

    & Monodepth2~\cite{godard2019digging}
    & 1.065 & 1.038 & 0.911 & 0.951
    & 1.036 & 0.880 & 0.842 & 1.049
    & 0.971
    & 0.649 & 0.791 & 0.724 & 0.629
    & 0.933 & 0.741
    & 0.744 \\

    & SC-DepthV3~\cite{sun2023scdepthv3}
    & 0.904 & 0.883 & 0.829 & 0.734
    & 0.937 & 0.749 & 0.732 & 0.987
    & 0.844
    & 0.644 & 0.801 & 0.617 & 0.554
    & 0.902 & 0.727
    & 0.708 \\

    & RA-Depth~\cite{he2022radepth}
    & 1.110 & 1.179 & 1.077 & 1.001
    & 1.097 & 0.948 & 0.826 & 1.289
    & 1.066
    & 0.724 & 0.822 & 0.820 & 0.710
    & 1.086 & 0.905
    & 0.845 \\

    & Manydepth2~\cite{zhou2025manydepth2}
    & 1.288 & 1.218 & 1.124 & 1.086
    & 1.306 & 1.006 & 1.001 & 1.222
    & 1.156
    & 0.706 & 0.862 & 0.912 & 0.747
    & 1.053 & 0.968
    & 0.875 \\

    \midrule
    % =========================================================
    % TRACKING COVERAGE
    % =========================================================
    \multicolumn{18}{c}{\textbf{Trck (\%)} $\uparrow$} \\
    \midrule

    \multirow{7}{*}{Geometry-based}
    & ORB-SLAM3~\cite{campos2021orbslam3}
    & 17.6 & 16.8 & 100.0 & 23.0
    & 0.0 & 43.8 & 100.0 & 58.0
    & 44.9
    & 84.5 & 84.5 & 19.3 & 0.0
    & 71.6 & 0.0
    & 43.3 \\

    & OpenVSLAM~\cite{sumikura2019openvslam}
    & 2.7 & 11.3 & 3.2 & 12.7
    & 6.8 & 26.3 & 96.0 & 26.6
    & 23.2
    & 100.0 & 100.0 & 100.0 & 100.0
    & 100.0 & 100.0
    & 100.0 \\

    & CubemapSLAM~\cite{wang2018cubemapslam}
    & 29.1 & 54.5 & 100.0 & 26.2
    & 73.8 & 100.0 & 100.0 & 22.1
    & 63.2
    & 100.0 & 89.0 & 79.2 & 100.0
    & 86.4 & 100.0
    & 92.4 \\

    & OV$^{2}$SLAM~\cite{fer2021ov2slam}
    & 100.0 & 100.0 & 100.0 & 100.0
    & 100.0 & 100.0 & 100.0 & 100.0
    & 100.0
    & 100.0 & 100.0 & 100.0 & 100.0
    & 100.0 & 100.0
    & 100.0 \\

    & SVO~\cite{forster2017svo}
    & 89.5 & 98.7 & 100.0 & 93.7
    & 98.2 & 95.7 & 93.8 & 98.9
    & 96.1
    & 100.0 & 100.0 & 73.6 & 100.0
    & 100.0 & 97.7
    & 95.2 \\

    & HSO~\cite{luo2022hso}
    & 0.0 & 0.0 & 0.0 & 26.2
    & 0.0 & 0.0 & 22.5 & 0.0
    & 6.1
    & 0.0 & 100.0 & 5.5 & 11.0
    & 100.0 & 9.2
    & 37.6 \\

    & \textbf{Ours}
    & 100.0 & 100.0 & 100.0 & 100.0
    & 100.0 & 100.0 & 100.0 & 100.0
    & 100.0
    & 100.0 & 100.0 & 100.0 & 100.0
    & 100.0 & 100.0
    & 100.0 \\

    \midrule

    \multirow{5}{*}{Learning-based}
    & Lite-Mono~\cite{zhang2023litemono}
    & 100.0 & 100.0 & 100.0 & 100.0
    & 100.0 & 100.0 & 100.0 & 100.0
    & 100.0
    & 100.0 & 100.0 & 100.0 & 100.0
    & 100.0 & 100.0
    & 100.0 \\

    & Monodepth2~\cite{godard2019digging}
    & 100.0 & 100.0 & 100.0 & 100.0
    & 100.0 & 100.0 & 100.0 & 100.0
    & 100.0
    & 100.0 & 100.0 & 100.0 & 100.0
    & 100.0 & 100.0
    & 100.0 \\

    & SC-DepthV3~\cite{sun2023scdepthv3}
    & 100.0 & 100.0 & 100.0 & 100.0
    & 100.0 & 100.0 & 100.0 & 100.0
    & 100.0
    & 100.0 & 100.0 & 100.0 & 100.0
    & 100.0 & 100.0
    & 100.0 \\

    & RA-Depth~\cite{he2022radepth}
    & 100.0 & 100.0 & 100.0 & 100.0
    & 100.0 & 100.0 & 100.0 & 100.0
    & 100.0
    & 100.0 & 100.0 & 100.0 & 100.0
    & 100.0 & 100.0
    & 100.0 \\

    & Manydepth2~\cite{zhou2025manydepth2}
    & 100.0 & 100.0 & 100.0 & 100.0
    & 100.0 & 100.0 & 100.0 & 100.0
    & 100.0
    & 100.0 & 100.0 & 100.0 & 100.0
    & 100.0 & 100.0
    & 100.0 \\

    \bottomrule
    \end{tabular*}
    \end{table*}

\begin{table*}[!tp]
    \centering
    \caption{Quantitative comparison of monocular visual odometry on the SeasideTown scenario of the TartanAir V2 dataset. Sequences P000--P007 are denoted as S0--S7,
respectively. The best results among fully tracked runs are shown in bold font. The symbol $\times$ indicates tracking failures, while $\dagger$ indicates partially tracked results with lower errors than the best fully tracked results. Errors for partially tracked runs are computed only over the successfully tracked trajectory segments.}
    \label{tab:tartan_vo_seaside}

    \begin{tabular*}{\textwidth}{
    @{\extracolsep{\fill}}
    ll
    *{9}{c}
    @{}
    }
    \toprule

    \textbf{Type}
    & \textbf{Method}
    & \textbf{S0}
    & \textbf{S1}
    & \textbf{S2}
    & \textbf{S3}
    & \textbf{S4}
    & \textbf{S5}
    & \textbf{S6}
    & \textbf{S7}
    & \textbf{Average} \\

    \midrule

    % =========================================================
    % ATE
    % =========================================================
    \multicolumn{11}{c}{\textbf{ATE (m)} $\downarrow$} \\
    \midrule

    \multirow{7}{*}{Geometry-based}
    & ORB-SLAM3~\cite{campos2021orbslam3}
    & $\times$ & 0.033$^{\dagger}$ & $\times$ & $\times$
    & $\times$ & $\times$ & $\times$ & $\times$
    & 0.033 \\

    & OpenVSLAM~\cite{sumikura2019openvslam}
    & $\times$ & 0.248 & 0.601 & 0.409
    & 1.114 & $\times$ & 0.916 & 0.706
    & 0.666 \\

    & CubemapSLAM~\cite{wang2018cubemapslam}
    & \textbf{0.017} & \textbf{0.036} & \textbf{0.013} & 0.031
    & 0.022$^{\dagger}$ & 0.031$^{\dagger}$ & 0.039$^{\dagger}$ & \textbf{0.023}
    & 0.026 \\

    & OV$^{2}$SLAM~\cite{fer2021ov2slam}
    & 0.172 & 0.209 & 0.807 & 0.112
    & 0.226 & \textbf{2.228} & 4.968 & 2.005
    & 1.341 \\

    & SVO~\cite{forster2017svo}
    & 1.992 & 0.154 & 8.231 & 9.688
    & 9.337 & 6.277 & 7.192 & 7.047
    & 6.240 \\

    & HSO~\cite{luo2022hso}
    & 0.289 & 0.274 & 0.256 & 0.354
    & 1.767 & 5.287 & 4.825 & 5.481
    & 2.317 \\

    & \textbf{Ours}
    & 0.089 & 0.074 & 0.017 & \textbf{0.015}
    & \textbf{0.109} & 3.712 & \textbf{0.175} & 0.184
    & 0.547 \\

    \midrule

    \multirow{5}{*}{Learning-based}
    & Lite-Mono~\cite{zhang2023litemono}
    & 7.325 & 9.765 & 5.979 & 6.074
    & 9.173 & 6.540 & 16.320 & 10.667
    & 8.980 \\

    & Monodepth2~\cite{godard2019digging}
    & 6.605 & 7.808 & 2.846 & 3.639
    & 3.458 & 4.186 & 7.125 & 5.875
    & 5.193 \\

    & SC-DepthV3~\cite{sun2023scdepthv3}
    & 7.015 & 8.274 & 2.966 & 4.907
    & 4.423 & 7.715 & 4.572 & 6.762
    & 5.829 \\

    & RA-Depth~\cite{he2022radepth}
    & 5.826 & 11.615 & 7.624 & 7.953
    & 12.619 & 10.976 & 15.705 & 9.422
    & 10.217 \\

    & Manydepth2~\cite{zhou2025manydepth2}
    & 5.830 & 9.640 & 6.702 & 6.442
    & 9.897 & 7.858 & 14.027 & 8.345
    & 8.593 \\

    \midrule
    % =========================================================
    % RPE TRANS
    % =========================================================
    \multicolumn{11}{c}{\textbf{RPE Trans (m)} $\downarrow$} \\
    \midrule

    \multirow{7}{*}{Geometry-based}
    & ORB-SLAM3~\cite{campos2021orbslam3}
    & $\times$ & 0.019 & $\times$ & $\times$
    & $\times$ & $\times$ & $\times$ & $\times$
    & 0.019 \\

    & OpenVSLAM~\cite{sumikura2019openvslam}
    & $\times$ & 0.150 & 0.125 & 0.335
    & 0.442 & $\times$ & 0.045 & 0.151
    & 0.208 \\

    & CubemapSLAM~\cite{wang2018cubemapslam}
    & 0.009 & 0.010 & 0.007 & 0.008
    & 0.007 & 0.011$^{\dagger}$ & 0.018 & \textbf{0.008}
    & 0.010 \\

    & OV$^{2}$SLAM~\cite{fer2021ov2slam}
    & 0.027 & 0.035 & 0.031 & 0.020
    & 0.013 & \textbf{0.059} & 0.106 & 0.058
    & 0.044 \\

    & SVO~\cite{forster2017svo}
    & 0.065 & 0.017 & 0.645 & 0.185
    & 0.380 & 1.544 & 0.688 & 0.085
    & 0.451 \\

    & HSO~\cite{luo2022hso}
    & 0.065 & 0.065 & 0.063 & 0.075
    & 1.703 & 1.779 & 2.570 & 0.099
    & 0.802 \\

    & \textbf{Ours}
    & \textbf{0.005} & \textbf{0.006} & \textbf{0.004} & \textbf{0.005}
    & \textbf{0.006} & 0.074 & \textbf{0.005} & 0.009
    & 0.014 \\

    \midrule

    \multirow{5}{*}{Learning-based}
    & Lite-Mono~\cite{zhang2023litemono}
    & 0.093 & 0.126 & 0.087 & 0.085
    & 0.101 & 0.155 & 0.108 & 0.122
    & 0.110 \\

    & Monodepth2~\cite{godard2019digging}
    & 0.090 & 0.140 & 0.076 & 0.082
    & 0.094 & 0.136 & 0.139 & 0.102
    & 0.107 \\

    & SC-DepthV3~\cite{sun2023scdepthv3}
    & 0.094 & 0.137 & 0.081 & 0.081
    & 0.094 & 0.124 & 0.122 & 0.109
    & 0.105 \\

    & RA-Depth~\cite{he2022radepth}
    & 0.114 & 0.132 & 0.112 & 0.095
    & 0.132 & 0.159 & 0.119 & 0.130
    & 0.124 \\

    & Manydepth2~\cite{zhou2025manydepth2}
    & 0.096 & 0.124 & 0.098 & 0.080
    & 0.108 & 0.145 & 0.123 & 0.120
    & 0.112 \\

    \midrule
    % =========================================================
    % RPE ROT
    % =========================================================
    \multicolumn{11}{c}{\textbf{RPE Rot ($^\circ$)} $\downarrow$} \\
    \midrule

    \multirow{7}{*}{Geometry-based}
    & ORB-SLAM3~\cite{campos2021orbslam3}
    & $\times$ & 0.062 & $\times$ & $\times$
    & $\times$ & $\times$ & $\times$ & $\times$
    & 0.062 \\

    & OpenVSLAM~\cite{sumikura2019openvslam}
    & $\times$ & 0.162 & 0.727 & 0.379
    & 0.629 & $\times$ & 0.181 & 0.456
    & 0.422 \\

    & CubemapSLAM~\cite{wang2018cubemapslam}
    & 0.051 & 0.050 & 0.046 & 0.055
    & 0.039 & 0.048$^{\dagger}$ & 0.045 & 0.050
    & 0.048 \\

    & OV$^{2}$SLAM~\cite{fer2021ov2slam}
    & 0.177 & 0.170 & 0.124 & 0.265
    & 0.091 & \textbf{0.284} & 0.477 & 0.203
    & 0.224 \\

    & SVO~\cite{forster2017svo}
    & 0.278 & 0.059 & 2.319 & 0.667
    & 1.107 & 3.305 & 1.916 & 0.269
    & 1.240 \\

    & HSO~\cite{luo2022hso}
    & 2.255 & 1.972 & 2.151 & 2.169
    & 3.955 & 6.155 & 5.137 & 2.106
    & 3.237 \\

    & \textbf{Ours}
    & \textbf{0.027} & \textbf{0.021} & \textbf{0.026} & \textbf{0.035}
    & \textbf{0.023} & 0.493 & \textbf{0.024} & \textbf{0.041}
    & 0.086 \\

    \midrule

    \multirow{5}{*}{Learning-based}
    & Lite-Mono~\cite{zhang2023litemono}
    & 0.608 & 0.621 & 0.776 & 0.730
    & 0.814 & 0.820 & 0.845 & 0.895
    & 0.764 \\

    & Monodepth2~\cite{godard2019digging}
    & 0.254 & 0.244 & 0.301 & 0.288
    & 0.396 & 0.424 & 0.372 & 0.426
    & 0.338 \\

    & SC-DepthV3~\cite{sun2023scdepthv3}
    & 0.479 & 0.451 & 0.522 & 0.530
    & 0.678 & 0.640 & 0.755 & 0.627
    & 0.585 \\

    & RA-Depth~\cite{he2022radepth}
    & 0.810 & 0.808 & 1.098 & 0.834
    & 0.983 & 0.948 & 0.984 & 1.135
    & 0.950 \\

    & Manydepth2~\cite{zhou2025manydepth2}
    & 0.639 & 0.648 & 0.877 & 0.683
    & 0.843 & 0.859 & 0.865 & 0.878
    & 0.786 \\

    \midrule
    % =========================================================
    % TRACKING COVERAGE
    % =========================================================
    \multicolumn{11}{c}{\textbf{Trck (\%)} $\uparrow$} \\
    \midrule

    \multirow{7}{*}{Geometry-based}
    & ORB-SLAM3~\cite{campos2021orbslam3}
    & 0.0 & 64.5 & 0.0 & 0.0
    & 45.5 & 0.0 & 47.0 & 0.0
    & 19.6 \\

    & OpenVSLAM~\cite{sumikura2019openvslam}
    & 3.9 & 100.0 & 100.0 & 100.0
    & 80.8 & 46.8 & 59.8 & 64.6
    & 69.5 \\

    & CubemapSLAM~\cite{wang2018cubemapslam}
    & 100.0 & 100.0 & 100.0 & 100.0
    & 85.4 & 76.8 & 64.5 & 100.0
    & 90.8 \\

    & OV$^{2}$SLAM~\cite{fer2021ov2slam}
    & 100.0 & 100.0 & 100.0 & 100.0
    & 100.0 & 100.0 & 100.0 & 100.0
    & 100.0 \\

    & SVO~\cite{forster2017svo}
    & 100.0 & 100.0 & 78.8 & 100.0
    & 99.1 & 84.9 & 93.0 & 100.0
    & 94.5 \\

    & HSO~\cite{luo2022hso}
    & 100.0 & 100.0 & 100.0 & 100.0
    & 83.4 & 75.1 & 63.4 & 100.0
    & 90.2 \\

    & \textbf{Ours}
    & 100.0 & 100.0 & 100.0 & 100.0
    & 100.0 & 100.0 & 100.0 & 100.0
    & 100.0 \\

    \midrule

    \multirow{5}{*}{Learning-based}
    & Lite-Mono~\cite{zhang2023litemono}
    & 100.0 & 100.0 & 100.0 & 100.0
    & 100.0 & 100.0 & 100.0 & 100.0
    & 100.0 \\

    & Monodepth2~\cite{godard2019digging}
    & 100.0 & 100.0 & 100.0 & 100.0
    & 100.0 & 100.0 & 100.0 & 100.0
    & 100.0 \\

    & SC-DepthV3~\cite{sun2023scdepthv3}
    & 100.0 & 100.0 & 100.0 & 100.0
    & 100.0 & 100.0 & 100.0 & 100.0
    & 100.0 \\

    & RA-Depth~\cite{he2022radepth}
    & 100.0 & 100.0 & 100.0 & 100.0
    & 100.0 & 100.0 & 100.0 & 100.0
    & 100.0 \\

    & Manydepth2~\cite{zhou2025manydepth2}
    & 100.0 & 100.0 & 100.0 & 100.0
    & 100.0 & 100.0 & 100.0 & 100.0
    & 100.0 \\

    \bottomrule
    \end{tabular*}
    \end{table*}
    
\begin{table*}[!tp]
    \centering
    \caption{Quantitative comparison of monocular visual odometry on the MIAS-FisheyeVO dataset. The best results among fully tracked runs are shown in bold font. The symbol $\times$ indicates tracking failures, while $\dagger$ indicates partially tracked results with lower errors than the best fully tracked results. Errors for partially tracked runs are computed only over the successfully tracked trajectory segments.}
    \label{tab:mias_vo}
    
    \begin{tabular*}{\textwidth}{
    @{\extracolsep{\fill}}
    ll
    *{7}{c}
    *{6}{c}
    @{}
    }
    \toprule
    
    \multirow{2}{*}{\textbf{Type}}
    & \multirow{2}{*}{\textbf{Method}}
    & \multicolumn{7}{c}{\textbf{Indoor Scenario}}
    & \multicolumn{6}{c}{\textbf{Outdoor Scenario}} \\
    \cmidrule(lr){3-9}
    \cmidrule(lr){10-15}
    
    &
    & \textbf{Seq.\,00}
    & \textbf{Seq.\,01}
    & \textbf{Seq.\,02}
    & \textbf{Seq.\,03}
    & \textbf{Seq.\,04}
    & \textbf{Seq.\,05}
    & \textbf{Average}
    & \textbf{Seq.\,00}
    & \textbf{Seq.\,01}
    & \textbf{Seq.\,02}
    & \textbf{Seq.\,03}
    & \textbf{Seq.\,04}
    & \textbf{Average} \\
    
    \midrule
    
    % =========================================================
    % ATE
    % =========================================================
    \multicolumn{15}{c}{\textbf{ATE (m)} $\downarrow$} \\
    \midrule

    \multirow{7}{*}{Geometry-based}
    & ORB-SLAM3~\cite{campos2021orbslam3}
    & 0.037 & 0.084 & 0.271 & 0.135 & 0.029 & 1.186
    & 0.290
    & 45.997 & $\times$ & 16.386 & 4.141 & 0.136
    & 16.665 \\

    & OpenVSLAM~\cite{sumikura2019openvslam}
    & 0.076 & $\times$ & $\times$ & 0.294 & 0.142 & 0.029
    & 0.135
    & $\times$ & 0.054$^{\dagger}$ & $\times$ & 0.370 & 0.125
    & 0.183 \\

    & CubemapSLAM~\cite{wang2018cubemapslam}
    & 0.223 & $\times$ & $\times$ & $\times$ & \textbf{0.006} & 0.002$^{\dagger}$
    & 0.077
    & $\times$ & \textbf{0.729} & $\times$ & 0.032$^{\dagger}$ & 0.013$^{\dagger}$
    & 0.258 \\

    & OV$^{2}$SLAM~\cite{fer2021ov2slam}
    & 0.077 & 0.888 & 1.021 & 0.093 & 0.944 & 0.046
    & 0.511
    & 43.257 & 1.693 & 14.996 & 0.299 & 0.142
    & 12.077 \\

    & SVO~\cite{forster2017svo}
    & 1.050 & 0.100 & $\times$ & 3.248 & 0.217 & 1.382
    & 1.199
    & 45.660 & 2.598 & 10.748 & 1.652 & 1.744
    & 12.480 \\

    & HSO~\cite{luo2022hso}
    & $\times$ & 0.113 & $\times$ & $\times$ & 0.235 & 0.346
    & 0.231
    & $\times$ & 3.735 & $\times$ & $\times$ & 0.391
    & 2.063 \\

    & \textbf{Ours}
    & \textbf{0.011} & \textbf{0.008} & \textbf{0.012} & \textbf{0.043} & 0.012 & \textbf{0.019}
    & 0.017
    & \textbf{17.299} & 1.644 & \textbf{4.259} & \textbf{0.185} & \textbf{0.032}
    & 4.684 \\

    \midrule

    \multirow{5}{*}{Learning-based}
    & Lite-Mono~\cite{zhang2023litemono}
    & 1.002 & 0.664 & 1.093 & 2.286 & 0.988 & 0.349
    & 1.064
    & 57.167 & 4.541 & 15.872 & 2.210 & 6.125
    & 17.183 \\

    & Monodepth2~\cite{godard2019digging}
    & 0.985 & 0.550 & 0.922 & 1.832 & 0.903 & 0.587
    & 0.963
    & 58.203 & 2.690 & 13.745 & 1.843 & 4.332
    & 16.163 \\

    & SC-DepthV3~\cite{sun2023scdepthv3}
    & 0.961 & 0.514 & 1.021 & 2.344 & 0.702 & 0.715
    & 1.043
    & 53.816 & 6.857 & 12.837 & 1.680 & 2.569
    & 15.552 \\

    & RA-Depth~\cite{he2022radepth}
    & 1.173 & 0.961 & 1.185 & 2.292 & 1.338 & 0.443
    & 1.232
    & 59.611 & 7.069 & 17.254 & 1.776 & 4.874
    & 18.117 \\

    & Manydepth2~\cite{zhou2025manydepth2}
    & 1.020 & 0.882 & 1.589 & 2.200 & 1.157 & 0.459
    & 1.218
    & 60.498 & 2.931 & 17.504 & 2.429 & 6.140
    & 17.900 \\

    \midrule
    % =========================================================
    % RPE TRANS
    % =========================================================
    \multicolumn{15}{c}{\textbf{RPE Trans (m)} $\downarrow$} \\
    \midrule

    \multirow{7}{*}{Geometry-based}
    & ORB-SLAM3~\cite{campos2021orbslam3}
    & 0.023 & 0.039 & 0.320 & 0.058$^{\dagger}$ & 0.028 & 1.000
    & 0.245
    & 3.926 & $\times$ & 3.866 & 0.602 & 0.110
    & 2.126 \\

    & OpenVSLAM~\cite{sumikura2019openvslam}
    & 0.025 & $\times$ & $\times$ & 0.137 & 0.058 & 0.038
    & 0.065
    & $\times$ & 0.034$^{\dagger}$ & $\times$ & 0.090 & 0.101
    & 0.075 \\

    & CubemapSLAM~\cite{wang2018cubemapslam}
    & 0.044 & $\times$ & $\times$ & $\times$ & \textbf{0.004} & 0.002$^{\dagger}$
    & 0.017
    & $\times$ & \textbf{0.134} & $\times$ & 0.020$^{\dagger}$ & 0.004$^{\dagger}$
    & 0.053 \\

    & OV$^{2}$SLAM~\cite{fer2021ov2slam}
    & 0.030 & 0.095 & 0.119 & 0.111 & 0.140 & 0.046
    & 0.090
    & 0.717 & 0.563 & 0.732 & 0.075 & 0.123
    & 0.442 \\

    & SVO~\cite{forster2017svo}
    & 0.329 & 0.018 & $\times$ & 0.543 & 0.055 & 0.171
    & 0.223
    & 4.454 & 3.227 & 3.495 & 0.168 & 0.577
    & 2.384 \\

    & HSO~\cite{luo2022hso}
    & $\times$ & 0.020 & $\times$ & $\times$ & 0.040 & 0.082
    & 0.047
    & $\times$ & 1.765 & $\times$ & $\times$ & 0.088
    & 0.926 \\

    & \textbf{Ours}
    & \textbf{0.006} & \textbf{0.004} & \textbf{0.007} & \textbf{0.064} & 0.010 & \textbf{0.012}
    & 0.017
    & \textbf{0.346} & 0.263 & \textbf{0.237} & \textbf{0.039} & \textbf{0.010}
    & 0.179 \\

    \midrule

    \multirow{5}{*}{Learning-based}
    & Lite-Mono~\cite{zhang2023litemono}
    & 0.074 & 0.053 & 0.113 & 0.261 & 0.085 & 0.063
    & 0.108
    & 0.604 & 0.848 & 0.883 & 0.218 & 0.410
    & 0.593 \\

    & Monodepth2~\cite{godard2019digging}
    & 0.065 & 0.042 & 0.107 & 0.232 & 0.074 & 0.074
    & 0.099
    & 0.568 & 0.575 & 0.795 & 0.188 & 0.313
    & 0.488 \\

    & SC-DepthV3~\cite{sun2023scdepthv3}
    & 0.064 & 0.041 & 0.132 & 0.310 & 0.069 & 0.064
    & 0.113
    & 0.518 & 0.623 & 0.839 & 0.165 & 0.199
    & 0.469 \\

    & RA-Depth~\cite{he2022radepth}
    & 0.079 & 0.061 & 0.115 & 0.275 & 0.119 & 0.063
    & 0.119
    & 0.663 & 0.863 & 0.931 & 0.185 & 0.339
    & 0.596 \\

    & Manydepth2~\cite{zhou2025manydepth2}
    & 0.074 & 0.058 & 0.149 & 0.270 & 0.095 & 0.083
    & 0.121
    & 0.658 & 0.569 & 0.944 & 0.215 & 0.408
    & 0.559 \\

    \midrule
    % =========================================================
    % RPE ROT
    % =========================================================
    \multicolumn{15}{c}{\textbf{RPE Rot ($^\circ$)} $\downarrow$} \\
    \midrule

    \multirow{7}{*}{Geometry-based}
    & ORB-SLAM3~\cite{campos2021orbslam3}
    & 0.263 & 0.475 & 0.561 & 0.380 & 0.313 & 0.488
    & 0.413
    & 6.617 & $\times$ & 13.661 & 2.116 & 0.202
    & 5.649 \\

    & OpenVSLAM~\cite{sumikura2019openvslam}
    & 0.303 & $\times$ & $\times$ & 0.842 & 0.912 & 0.458
    & 0.629
    & $\times$ & 0.124$^{\dagger}$ & $\times$ & 0.493 & 0.403
    & 0.340 \\

    & CubemapSLAM~\cite{wang2018cubemapslam}
    & 0.765 & $\times$ & $\times$ & $\times$ & \textbf{0.078} & 0.042$^{\dagger}$
    & 0.295
    & $\times$ & 0.985 & $\times$ & 0.091$^{\dagger}$ & 0.037$^{\dagger}$
    & 0.371 \\

    & OV$^{2}$SLAM~\cite{fer2021ov2slam}
    & 0.652 & 0.844 & 2.317 & 0.326 & 1.933 & 0.162
    & 1.039
    & 0.613 & 0.661 & 5.741 & 0.583 & 0.332
    & 1.586 \\

    & SVO~\cite{forster2017svo}
    & 2.180 & 0.182 & $\times$ & 9.652 & 1.306 & 4.081
    & 3.480
    & 6.155 & 21.680 & 22.657 & 0.510 & 0.973
    & 10.395 \\

    & HSO~\cite{luo2022hso}
    & $\times$ & 0.239 & $\times$ & $\times$ & 0.868 & 0.484
    & 0.530
    & $\times$ & 9.726 & $\times$ & $\times$ & 0.271
    & 4.999 \\

    & \textbf{Ours}
    & \textbf{0.059} & \textbf{0.077} & \textbf{0.088} & \textbf{0.206} & 0.109 & \textbf{0.136}
    & 0.113
    & \textbf{0.183} & \textbf{0.400} & \textbf{0.426} & \textbf{0.363} & \textbf{0.044}
    & 0.283 \\

    \midrule

    \multirow{5}{*}{Learning-based}
    & Lite-Mono~\cite{zhang2023litemono}
    & 1.313 & 0.809 & 2.384 & 3.319 & 2.173 & 1.096
    & 1.849
    & 2.677 & 6.645 & 7.349 & 1.045 & 1.248
    & 3.793 \\

    & Monodepth2~\cite{godard2019digging}
    & 1.140 & 0.669 & 2.272 & 2.349 & 2.618 & 1.192
    & 1.707
    & 1.889 & 6.605 & 6.370 & 0.821 & 1.055
    & 3.348 \\

    & SC-DepthV3~\cite{sun2023scdepthv3}
    & 1.030 & 0.606 & 2.233 & 2.382 & 1.840 & 0.892
    & 1.497
    & 1.659 & 6.110 & 5.075 & 0.721 & 0.693
    & 2.852 \\

    & RA-Depth~\cite{he2022radepth}
    & 1.427 & 0.983 & 2.538 & 2.704 & 3.409 & 1.024
    & 2.014
    & 2.213 & 5.398 & 6.701 & 1.032 & 0.956
    & 3.260 \\

    & Manydepth2~\cite{zhou2025manydepth2}
    & 1.424 & 0.871 & 3.208 & 3.052 & 2.731 & 1.278
    & 2.094
    & 3.178 & 6.513 & 8.280 & 1.238 & 1.183
    & 4.078 \\

    \midrule
    % =========================================================
    % TRACKING COVERAGE
    % =========================================================
    \multicolumn{15}{c}{\textbf{Trck (\%)} $\uparrow$} \\
    \midrule

    \multirow{7}{*}{Geometry-based}
    & ORB-SLAM3~\cite{campos2021orbslam3}
    & 100.0 & 100.0 & 87.5 & 67.5 & 100.0 & 100.0
    & 92.5
    & 87.2 & 4.2 & 100.0 & 100.0 & 81.0
    & 74.5 \\

    & OpenVSLAM~\cite{sumikura2019openvslam}
    & 61.2 & 35.0 & 38.8 & 61.3 & 83.0 & 60.0
    & 56.6
    & 33.5 & 87.5 & 32.7 & 100.0 & 100.0
    & 70.7 \\

    & CubemapSLAM~\cite{wang2018cubemapslam}
    & 100.0 & 22.0 & 0.0 & 0.0 & 100.0 & 78.0
    & 50.0
    & 33.3 & 100.0 & 39.9 & 88.9 & 52.0
    & 62.8 \\

    & OV$^{2}$SLAM~\cite{fer2021ov2slam}
    & 100.0 & 100.0 & 100.0 & 71.2 & 100.0 & 100.0
    & 95.2
    & 100.0 & 100.0 & 100.0 & 100.0 & 100.0
    & 100.0 \\

    & SVO~\cite{forster2017svo}
    & 55.6 & 100.0 & 24.7 & 60.5 & 82.3 & 100.0
    & 70.5
    & 88.1 & 94.7 & 68.2 & 100.0 & 100.0
    & 90.2 \\

    & HSO~\cite{luo2022hso}
    & 21.1 & 100.0 & 0.0 & 32.4 & 100.0 & 100.0
    & 58.9
    & 40.9 & 92.7 & 5.9 & 21.2 & 100.0
    & 52.1 \\

    & \textbf{Ours}
    & 100.0 & 100.0 & 100.0 & 100.0 & 100.0 & 100.0
    & 100.0
    & 100.0 & 100.0 & 100.0 & 100.0 & 100.0
    & 100.0 \\

    \midrule

    \multirow{5}{*}{Learning-based}
    & Lite-Mono~\cite{zhang2023litemono}
    & 100.0 & 100.0 & 100.0 & 100.0 & 100.0 & 100.0
    & 100.0
    & 100.0 & 100.0 & 100.0 & 100.0 & 100.0
    & 100.0 \\

    & Monodepth2~\cite{godard2019digging}
    & 100.0 & 100.0 & 100.0 & 100.0 & 100.0 & 100.0
    & 100.0
    & 100.0 & 100.0 & 100.0 & 100.0 & 100.0
    & 100.0 \\

    & SC-DepthV3~\cite{sun2023scdepthv3}
    & 100.0 & 100.0 & 100.0 & 100.0 & 100.0 & 100.0
    & 100.0
    & 100.0 & 100.0 & 100.0 & 100.0 & 100.0
    & 100.0 \\

    & RA-Depth~\cite{he2022radepth}
    & 100.0 & 100.0 & 100.0 & 100.0 & 100.0 & 100.0
    & 100.0
    & 100.0 & 100.0 & 100.0 & 100.0 & 100.0
    & 100.0 \\

    & Manydepth2~\cite{zhou2025manydepth2}
    & 100.0 & 100.0 & 100.0 & 100.0 & 100.0 & 100.0
    & 100.0
    & 100.0 & 100.0 & 100.0 & 100.0 & 100.0
    & 100.0 \\

    \bottomrule
    \end{tabular*}
\end{table*}

All methods are also evaluated on the newly collected NIO-Fisheye dataset, which contains video sequences captured by a forward-facing fisheye camera. When the vehicle navigates straight ahead, visual features can undergo substantial appearance and scale changes across consecutive images, with regions corresponding to previously distant image pixels expanding rapidly as the vehicle approaches them. Moreover, visual features corresponding to distant 3-D landmarks may exhibit limited parallax when their viewing rays are nearly aligned with the camera translation direction, increasing the uncertainty of triangulated depth estimates and weakening the constraints on camera translation~\cite{yang2018dvso}. These characteristics are evident in the video sequences of the NIO-Fisheye dataset. Unlike the compared geometry-based methods, S$^2$D-FVO incorporates depth estimates produced by DepthNet as a scale reference during relative pose initialization, thereby reducing its reliance on the depth obtained through 3-D landmark triangulation when parallax is limited. The resulting scaled relative poses provide more reliable initial camera pose estimates for subsequent camera tracking and local mapping. As shown in Table~\ref{tab:nio_vo}, S$^2$D-FVO and OV$^2$SLAM maintain continuous camera tracking after pose initialization on all five sequences, while the other geometry-based methods fail to maintain camera tracking on several sequences. Although several learning-based methods achieve lower RPE Trans and RPE Rot, S$^2$D-FVO still achieves an average ATE of 93.113\,m, with the lowest ATE among methods with full tracking coverage across all five sequences, demonstrating its advantage in overall trajectory estimation accuracy.

To further evaluate translational and rotational drift over long-distance driving sequences, the corresponding errors are measured over path lengths of 100--800\,m. As reported in Table~\ref{tab:kitti_nio_et_er}, S$^2$D-FVO achieves reductions of 92.3\% in translational error and 95.8\% in rotational error relative to OV$^2$SLAM on the KITTI-360 dataset. On the NIO-Fisheye dataset, the corresponding reductions reach 76.2\% and 77.5\%. These results demonstrate the effectiveness of S$^2$D-FVO in mitigating scale drift while maintaining accurate rotational estimates over long trajectories.

The six room sequences of the TUM-VI dataset capture brisk walking with largely steady camera motion and few abrupt changes in direction or speed~\cite{schubert2018tum}. Almost all evaluated methods achieve high tracking coverage and generally low trajectory errors across these sequences. More pronounced differences are observed on sequences Room 3 and Room 5, where the proposed system reduces ATE by 87.3\% and 90.4\% compared with ORB-SLAM3, respectively, and by 98.5\% and 98.8\% compared with OV$^2$SLAM, respectively. Although OpenVSLAM and CubemapSLAM achieve lower ATE on Room 3, S$^2$D-FVO achieves the lowest mean ATE of 0.012\,m among all evaluated methods, with ATE ranging from 0.005 to 0.020\,m across the six sequences, as reported in Table~\ref{tab:tumvi_vo}. These results demonstrate the stability of S$^2$D-FVO across sequences of the TUM-VI dataset.

In addition, experimental results on the TartanAir V2 dataset demonstrate that S$^2$D-FVO achieves high trajectory estimation accuracy, with ATE below 0.2\,m on 21 of the 22 evaluated sequences while maintaining continuous camera tracking after pose initialization, as shown in Tables~\ref{tab:tartan_vo_indoor} and~\ref{tab:tartan_vo_seaside}. The most pronounced improvements are observed in the House scenario, where S$^2$D-FVO reduces ATE by 93.2\%--99.3\% compared with OV$^2$SLAM across all eight sequences, while ORB-SLAM3, OpenVSLAM, CubemapSLAM, and HSO frequently fail to maintain camera tracking. In the RetroOffice scenario, OpenVSLAM and CubemapSLAM also achieve low trajectory errors on several sequences. The proposed system obtains ATE values of 0.004--0.007\,m on five of the six sequences, achieving errors comparable to or lower than those of these methods. On sequences P002 and P004, it further achieves higher tracking coverage while maintaining ATE comparable to that of CubemapSLAM. In the outdoor SeasideTown scenario, S$^2$D-FVO reduces ATE by 51.8\% and 96.5\% compared with OV$^2$SLAM on sequences P004 and P006, respectively. However, sequence P005 presents substantial challenges for all evaluated methods, resulting in either large trajectory errors or reduced tracking coverage. Although S$^2$D-FVO achieves a relatively high ATE of 3.712\,m on this sequence, it still outperforms all evaluated learning-based methods and several geometry-based methods in trajectory estimation accuracy while maintaining continuous camera tracking after pose initialization. Overall, the results on the TartanAir V2 dataset demonstrate the applicability of S$^2$D-FVO across diverse indoor and outdoor scenarios, including sequences that pose substantial challenges to existing methods. 

The comparison on the MIAS-FisheyeVO dataset further demonstrates substantial improvements in local translational and rotational estimation accuracy across indoor and outdoor scenes, as shown in Table~\ref{tab:mias_vo}. Compared with OV$^2$SLAM, the average RPE Trans and RPE Rot are reduced by 81.1\% and 89.1\% for indoor scenes and by 59.5\% and 82.2\% for outdoor scenes, respectively. S$^2$D-FVO also achieves lower ATE on all 11 sequences, with average reductions of 96.7\% and 61.2\% for indoor and outdoor scenes, respectively. Among methods with full tracking coverage, S$^2$D-FVO achieves the lowest ATE on five of the six indoor sequences. Although CubemapSLAM achieves a lower ATE on indoor Seq.\,04 and outdoor Seq.\,01, its average tracking coverage is limited to 50.0\% and 62.8\% in indoor and outdoor scenes, respectively, highlighting the superior balance between trajectory estimation accuracy and tracking coverage achieved by S$^2$D-FVO.

\begin{figure}
    \centering
    \includegraphics[width=\columnwidth]{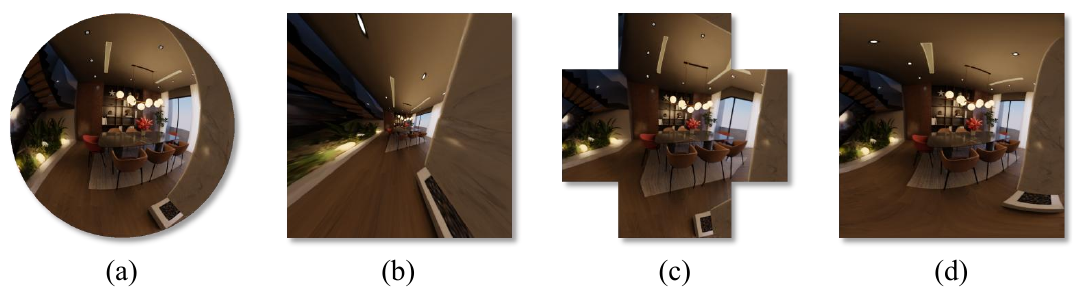}
    \caption{Comparison of image representations adopted by different monocular fisheye visual odometry systems: (a) original fisheye image; (b) undistorted perspective image; (c) unfolded cubemap image; (d) ERP image.}
    \label{fig:fisheye_representation}
\end{figure}

Beyond the quantitative trajectory comparisons, the evaluated geometry-based methods are also analyzed in terms of their fisheye image representations and reprojection residuals. Fig.~\ref{fig:fisheye_representation} illustrates the corresponding image representations. ORB-SLAM3 and SVO~\cite{forster2017svo} retain the original fisheye images without rectification and compute reprojection residuals using their respective fisheye camera models. OpenVSLAM and OV$^2$SLAM also extract and track visual features in the original fisheye images. The detected visual features are mapped onto a perspective image plane, and their reprojection residuals are computed accordingly. In contrast, HSO rectifies the entire fisheye image into a perspective image before camera tracking, which significantly reduces the FoV. CubemapSLAM instead employs perspective piecewise remapping to project viewing rays onto the faces of a cube, retaining a wider FoV at the cost of additional processing of the resulting perspective images. S$^2$D-FVO adopts a different approach by decoupling the image representation from the definition of reprojection residuals. ERP provides a unified image representation for visual feature extraction and depth learning across different fisheye camera models, while reprojection residuals are defined directly on the unit sphere rather than computed from differences in ERP pixel coordinates. The proposed tangent-space angular reprojection residual uses the spherical logarithmic map to represent directional discrepancies as tangent-space vectors whose norms equal the corresponding geodesic distances, rather than the chordal distances measured by Euclidean differences between bearing vectors. Accordingly, spherical bundle adjustment directly minimizes angular reprojection errors on the unit sphere, avoiding small-angle approximations and dependence on the equirectangular projection. This design allows seamless adaptation to different fisheye camera models.

The comparison with learning-based methods further highlights the benefits of incorporating DepthNet estimates into the monocular fisheye visual odometry system. The evaluated PoseNets estimate camera trajectories by accumulating relative poses obtained from consecutive images, without explicit multi-frame pose optimization. In contrast, the proposed system uses depth estimates from DepthNet to provide a scale reference, while multi-frame visual correspondences and spherical bundle adjustment constrain and refine the camera trajectory. These results indicate that integrating depth estimates from DepthNet with geometry-based camera pose estimation provides an effective approach to scale-consistent monocular fisheye visual odometry.

The above experimental results suggest that monocular fisheye visual odometry remains susceptible to camera tracking failures and accumulated trajectory drift. Camera tracking may fail when visual features are insufficient or parallax is limited, while errors in relative pose and scale may accumulate even when camera tracking remains continuous. By incorporating DepthNet outputs as a scale reference, S$^2$D-FVO reduces its reliance on both the availability of triangulated 3-D landmarks and the accuracy of their depth estimates, thereby improving the robustness of camera tracking and local mapping.

\subsection{Runtime Analysis}

The runtime of the proposed S$^2$D-FVO is measured on a heterogeneous system consisting of an Intel Core Ultra 9 285K CPU and an NVIDIA GeForce RTX 5090 GPU. SuperPoint, SphereGlue, and DepthNet are deployed in ONNX format, with neural network inference performed on the GPU using ONNX Runtime. The remaining computations are performed on the CPU. Table~\ref{tab:runtime_breakdown} reports the average runtime of the main components on sequence Room\,1 of the TUM-VI dataset. The runtime of SphereGlue is included in that of pose estimation. The runtime of the subsequently executed processes is measured only when a loop closure is verified.

ERP remapping requires an average of only $1.5$\,ms per frame, introducing only a small computational overhead. Camera tracking requires an average of $70.7$\,ms per frame, with the highest computational cost arising from visual feature matching using SphereGlue. In comparison, depth estimation using the incorporated DepthNet requires only $8.5$\,ms per frame. For local mapping, the average runtime is $165.4$\,ms per keyframe, including spherical bundle adjustment, which requires an average of $123.1$\,ms. During loop closing, verified loop closures are used to correct accumulated trajectory drift through pose graph optimization, with each loop correction taking less than $1.0$\,s on average. Overall, S$^2$D-FVO operates at $\sim10$\,FPS, indicating that depth estimation can be integrated into the complete system with relatively low additional computational cost.

\begin{table}[t]
    \centering
    \caption{Runtime of the main components of S$^2$D-FVO.}
    \label{tab:runtime_breakdown}
    \resizebox{\columnwidth}{!}{%
    \begin{tabular}{llc}
        \toprule
        Module & Main Component & Time (ms) \\
        \midrule
        Data Preprocessing
            & ERP Remapping & $1.5 \pm 0.4$ \\
        \midrule
        Depth Estimation
            & DepthNet & $8.5 \pm 6.9$ \\
        \midrule
        \multirow{6}{*}{Camera Tracking}
            & SuperPoint & $17.8 \pm 1.6$ \\
            & SphereGlue & $38.9 \pm 5.4$ \\
            & Pose Estimation & $47.9 \pm 5.3$ \\
            & Local Map Tracking & $0.01 \pm 0.05$ \\
            & New Keyframe Decision & $3.6 \pm 5.3$ \\
            & \textbf{Total} & $\mathbf{70.7 \pm 14.3}$ \\
        \midrule
        \multirow{6}{*}{Local Mapping}
            & Keyframe Insertion & $13.5 \pm 8.0$ \\
            & Map Point Culling & $0.1 \pm 0.1$ \\
            & 3-D Landmark Triangulation & $16.7 \pm 31.5$ \\
            & Spherical Bundle Adjustment & $123.1 \pm 129.9$ \\
            & Keyframe Culling & $12.4 \pm 17.0$ \\
            & \textbf{Total} & $\mathbf{165.4 \pm 170.6}$ \\
        \midrule
        \multirow{6}{*}{Loop Closing}
            & Database Query & $12.9 \pm 6.0$ \\
            & Geometric Verification & $72.0 \pm 93.2$ \\
            & \textbf{Total}
            & $\mathbf{45.2 \pm 78.5}$ \\
            \cline{2-3}
            & Loop Fusion & $206.8 \pm 160.1$ \\
            & Pose-Graph Optimization
            & $182.4 \pm 53.3$ \\
            & \textbf{Total}
            & $\mathbf{457.0 \pm 229.1}$ \\
        \bottomrule
    \end{tabular}%
    }
\end{table}

\subsection{Ablation Studies}
\label{sec.exp_ablation}
Two ablation studies are conducted on the KITTI-360 dataset to validate the effectiveness of the main components of the proposed monocular fisheye depth estimation and visual odometry framework. The first ablation study validates the effectiveness of DepthNet pretraining and the relative pose estimation strategies used for view synthesis during unsupervised DepthNet training. The second ablation study investigates the effect of incorporating DepthNet outputs into monocular fisheye visual odometry and validates the effectiveness of the proposed median depth-constraint scale regularization term.

\begin{figure}[!tp]
\centering
\includegraphics[width=\columnwidth]{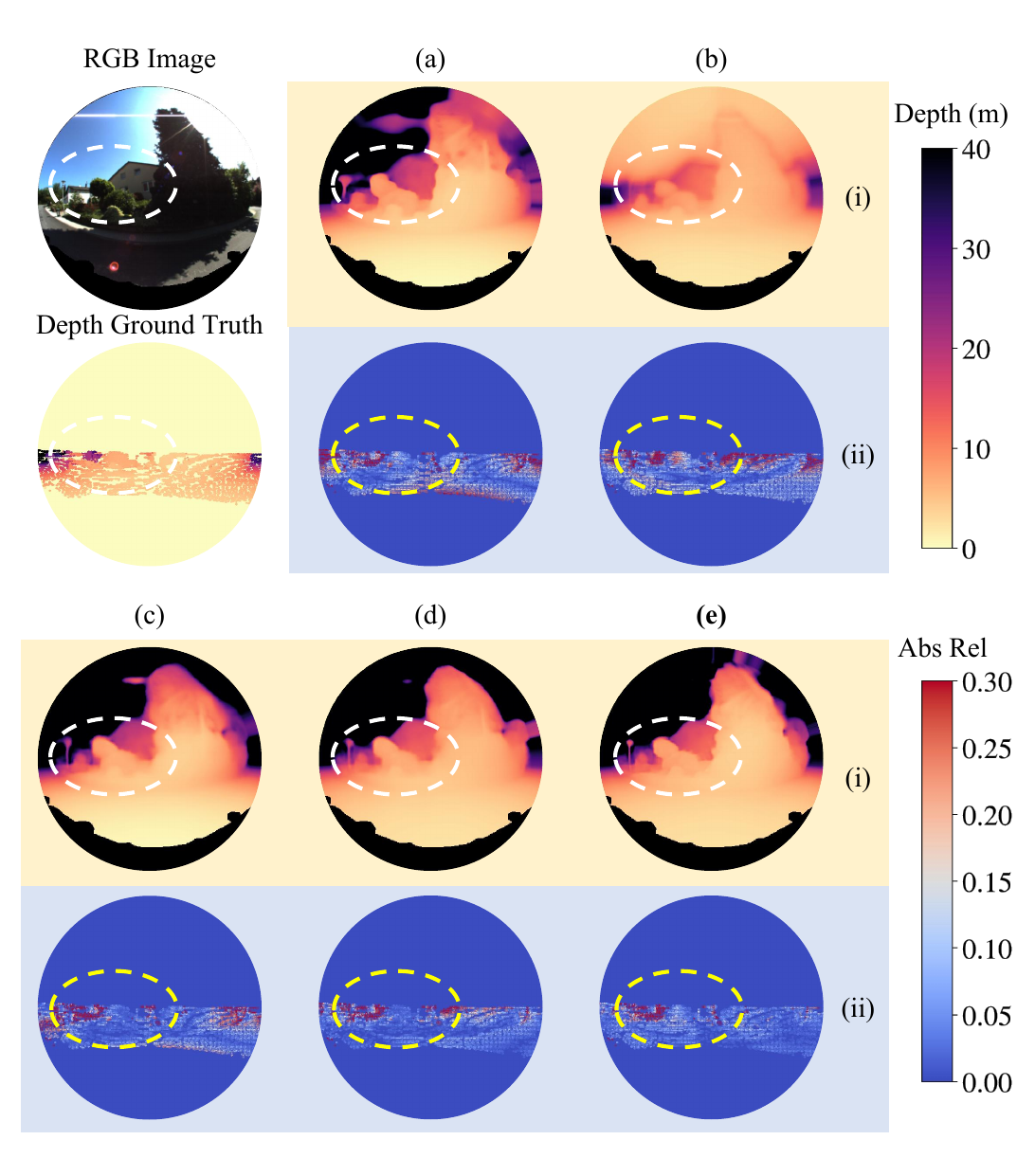}
    \caption{
    Qualitative results of the monocular fisheye depth learning ablation study on the KITTI-360 dataset: (i) estimated depth maps; (ii) corresponding error maps. (a)--(e) correspond to the ablation study configurations. White and yellow dashed contours highlight regions with substantial improvements.
    }
    \label{fig:ablation_depth}
\end{figure}

In the first ablation study, five configurations (a)--(e) are evaluated. Configuration (a) uses Monodepth2~\cite{godard2019digging} as the baseline DepthNet, configuration (b) pretrains DepthNet using pseudo depth maps generated by UniDAC, and configurations (c)--(e) further train the pretrained DepthNet in an unsupervised manner and differ only in the source of the relative poses used for view synthesis: (c) uses the relative poses estimated by PoseNet, (d) uses those estimated using the birotation solution~\cite{zhao2025birotation}, and (e) further refines the relative poses estimated using the birotation solution through spherical bundle adjustment. The results of this ablation study and the corresponding qualitative visualizations are presented in Table~\ref{tab:ablation_training} and Fig.~\ref{fig:ablation_depth}, respectively.

\begin{table*}[!tp]
\centering
\caption{
Ablation study of different pose estimation strategies used for view synthesis during unsupervised monocular fisheye depth learning on the KITTI-360 dataset. ``BS'' denotes the birotation solution, while ``SBA'' denotes the spherical bundle adjustment. 
}
\label{tab:ablation_training}

\begin{tabular*}{\textwidth}{
@{\extracolsep{\fill}}
l
ccccccc
@{}
}
\toprule

\textbf{Configuration}
& \textbf{Abs Rel} $\downarrow$
& \textbf{Sq Rel} $\downarrow$
& \textbf{RMSE} $\downarrow$
& \textbf{RMSE$_{\log}$} $\downarrow$
& $\boldsymbol{\delta < 1.25} \uparrow$
& $\boldsymbol{\delta < 1.25^2} \uparrow$
& $\boldsymbol{\delta < 1.25^3} \uparrow$ \\
\midrule

(a) Baseline
& 0.212
& 1.186
& 4.616
& 0.307
& 0.656
& 0.896
& 0.957
\\

(b) Pretrained DepthNet
& 0.179
& 0.808
& 4.588
& 0.251
& 0.703
& 0.922
& 0.977
\\

(c) Pretrained DepthNet + PoseNet
& 0.190
& 0.904
& 4.110
& 0.272
& 0.696
& 0.917
& 0.968
\\

(d) Pretrained DepthNet + BS 
& 0.135
& 0.610
& 3.708
& 0.204
& 0.831
& 0.953
& 0.982
\\

(e) Pretrained DepthNet + BS + SBA
& \textbf{0.122}
& \textbf{0.582}
& \textbf{3.613}
& \textbf{0.183}
& \textbf{0.858}
& \textbf{0.959}
& \textbf{0.984}
\\

\bottomrule
\end{tabular*}
\end{table*}

Configurations (a) and (b) are first compared to validate the contribution of DepthNet pretraining. Compared with the baseline used in configuration (a), the pretrained DepthNet in configuration (b) achieves a 15.6\% reduction in Abs Rel, while consistent improvements are observed across all other reported evaluation metrics, demonstrating the effectiveness of pretraining with the pseudo depth maps generated by UniDAC. Configurations (b) and (c) are then compared to investigate the efficacy of subsequent unsupervised training using the relative poses estimated by PoseNet. After enforcing photometric consistency using the PoseNet outputs, Abs Rel increases by 6.1\%, while six of the seven reported evaluation metrics indicate degraded depth estimation performance. These results suggest that the accuracy of the relative poses used for view synthesis plays a critical role in unsupervised depth learning. Larger relative pose errors increase the misalignment between the target and synthesized images, causing the photometric loss to be increasingly affected by relative pose errors. Since photometric loss is highly sensitive to image alignment and may exhibit inconsistent update directions~\cite{shu2020featdepth}, such relative pose errors may lead to erroneous DepthNet updates and consequently degrade depth estimation performance. 

The comparison between configurations (c) and (d) further demonstrates the importance of relative pose accuracy for view synthesis during unsupervised DepthNet training. Configuration (d) replaces PoseNet with the birotation solution~\cite{zhao2025birotation} for relative pose estimation. Unlike PoseNet, which implicitly regresses relative poses from image pairs, the birotation solution explicitly estimates relative poses from matched correspondences and achieves higher pose estimation accuracy~\cite{zhao2025birotation}. Compared with configuration (c), configuration (d) reduces Abs Rel by 28.9\%, with reductions of up to 32.5\% across other error metrics and improvements of up to 19.4\% in threshold-based accuracy. These performance gains demonstrate that more accurate relative poses provide more reliable photometric supervision, thereby further improving the depth estimation performance of the pretrained DepthNet.

Finally, the comparison between configurations (d) and (e) validates the effectiveness of spherical bundle adjustment. While the birotation solution estimates relative poses independently for each frame pair, spherical bundle adjustment jointly refines the camera poses using visual correspondences across multiple frames. Compared with configuration (d), configuration (e) reduces Abs Rel by 9.6\%, indicating that multi-frame pose refinement provides more reliable relative poses for view synthesis and further improves the performance of the proposed monocular fisheye depth learning framework.

\begin{figure}
    \centering
    \includegraphics[width=\columnwidth]{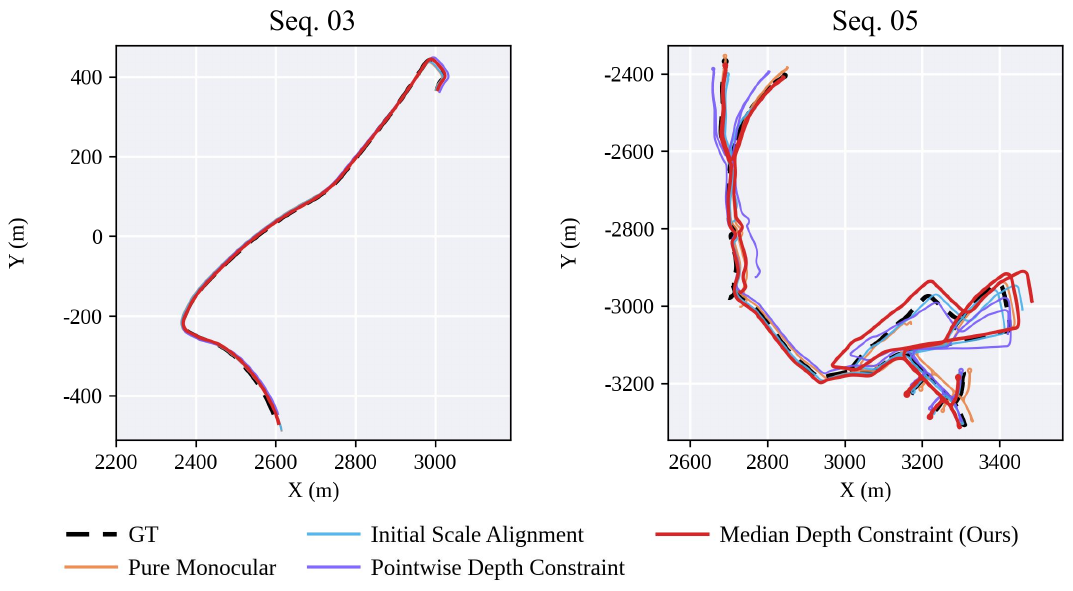}
    \caption{Ablation study on different scale regularization strategies for monocular fisheye visual odometry on the KITTI-360 dataset.}
    \label{fig:vo_ablation}
\end{figure}

In the second ablation study, four configurations (a)--(d) are evaluated to investigate the effectiveness of different scale regularization strategies for monocular fisheye visual odometry. Configuration (a) performs monocular fisheye visual odometry without using depth estimates as a scale reference. Configuration (b) incorporates depth estimates from DepthNet to initialize the translation magnitude during relative pose estimation. Building on configuration (b), configurations (c) and (d) further introduce scale regularization into spherical bundle adjustment, with the former constraining individual 3-D landmark depth and the latter constraining the common scale using median depth. Quantitative and qualitative results of this ablation study are presented in Table~\ref{tab:ablation_vo} and Fig.~\ref{fig:vo_ablation}, respectively.

\begin{table*}[!tp]
\centering
\caption{
Ablation study on different scale regularization strategies on the KITTI-360 dataset.}
\label{tab:ablation_vo}
\begin{tabular*}{\textwidth}{
@{\extracolsep{\fill}}
lcccccc
@{}
}
\toprule

\multirow{2}{*}{\textbf{Configuration}}
& \multicolumn{3}{c}{\textbf{Seq.\,03}}
& \multicolumn{3}{c}{\textbf{Seq.\,05}} \\
\cmidrule(lr){2-4}
\cmidrule(lr){5-7}

& \textbf{ATE (m)} $\downarrow$
& \textbf{RPE Trans (m)} $\downarrow$
& $\boldsymbol{e_t}$ \textbf{(\%)} $\downarrow$
& \textbf{ATE (m)} $\downarrow$
& \textbf{RPE Trans (m)} $\downarrow$
& $\boldsymbol{e_t}$ \textbf{(\%)} $\downarrow$ \\
\midrule

(a) No Scale Reference
& 6.108 & 1.233 & 2.420
& 55.019 & 0.221 & 5.866 \\

(b) Scale Reference for Pose Initialization
& 4.978 & 0.087 & 1.907
& 44.176 & 0.194 & 7.174 \\

(c) Scale Regularization Using Pointwise Depth
& 10.066 & 0.145 & 4.024
& 28.579 & \textbf{0.074} & \textbf{3.355} \\

(d) \textbf{Scale Regularization Using Median Depth (Ours)}
& \textbf{3.052} & \textbf{0.031} & \textbf{1.183}
& \textbf{23.577} & 0.094 & 4.593 \\

\bottomrule
\end{tabular*}
\end{table*}

Configurations (a) and (b) are first compared to evaluate the effectiveness of using depth estimates from DepthNet as a scale reference during relative pose initialization. Compared with configuration (a), configuration (b) reduces ATE by 18.50\% and 19.71\% on Seqs.\,03 and 05, respectively, with corresponding reductions of 92.94\% and 12.22\% in RPE Trans. Nevertheless, $e_t$ increases from 5.866\% to 7.174\% on Seq.\,05, indicating that incorporating depth estimates only during relative pose initialization does not consistently improve trajectory estimation accuracy. Since configuration (b) uses depth estimates only for pose initialization, the subsequent spherical bundle adjustment does not explicitly preserve the initialized scale, leaving the estimated trajectory susceptible to accumulated scale drift.

Configurations (b) and (c) are then compared to validate the effectiveness of scale regularization using pointwise depth in spherical bundle adjustment. Compared with configuration (b), configuration (c) reduces ATE and RPE Trans on Seq.\,05 by 35.31\% and 61.86\%, respectively. Nevertheless, on Seq.\,03, ATE increases from 4.978 m to 10.066 m, while RPE Trans increases from 0.087 m to 0.145 m. These results indicate that scale regularization using pointwise depth does not consistently improve trajectory estimation accuracy. In configuration (c), the scale regularization term enforces consistency between the depth of each 3-D landmark and the corresponding depth estimated by DepthNet. Errors in these depth estimates may perturb the optimized 3-D landmark positions and subsequently affect camera pose estimation through joint optimization.

Finally, the comparison between configurations (c) and (d) validates the effectiveness of the proposed median depth-based scale regularization term. Rather than using individual depth estimates from DepthNet as scale references, the proposed strategy provides a common scale reference for the reconstructed 3-D structure by enforcing consistency between the median depth of the triangulated 3-D landmarks and that of the corresponding DepthNet estimates. Compared with configuration (c), configuration (d) reduces ATE by 69.68\% and 17.50\% on Seqs.\,03 and 05, respectively. On Seq.\,03, RPE Trans and $e_t$ are further reduced by 78.62\% and 70.60\%, respectively. By providing a common scale reference while allowing individual 3-D landmark positions to be determined primarily by multi-view correspondences, the proposed strategy reduces the dependence on potentially inaccurate individual depth estimates. These results demonstrate the effectiveness of the proposed scale regularization strategy in improving trajectory estimation accuracy.

\section{Discussion}
\label{sec.Discussion}

\begin{figure*}[!tp]
\centering
\includegraphics[width=\textwidth]{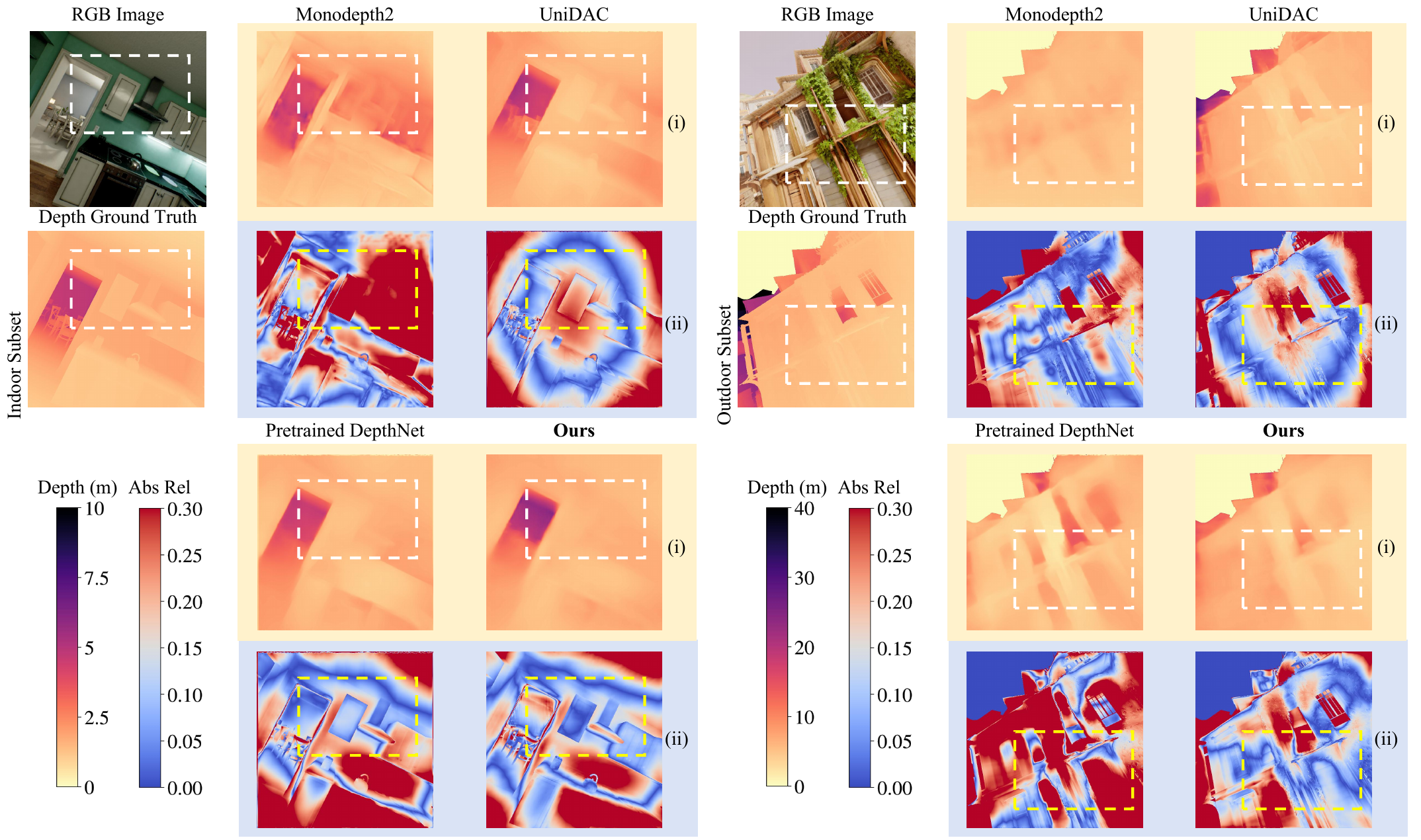}
\caption{
Qualitative comparison of monocular depth estimation on perspective video sequences from the TartanAir V2 dataset: (i) estimated depth maps; (ii) corresponding error maps. White and yellow dashed contours highlight regions with substantial improvements in depth estimation accuracy.
}
\label{fig:discussion_depth_front}
\end{figure*}

\begin{figure*}[!tp]
\centering
\includegraphics[width=\textwidth]{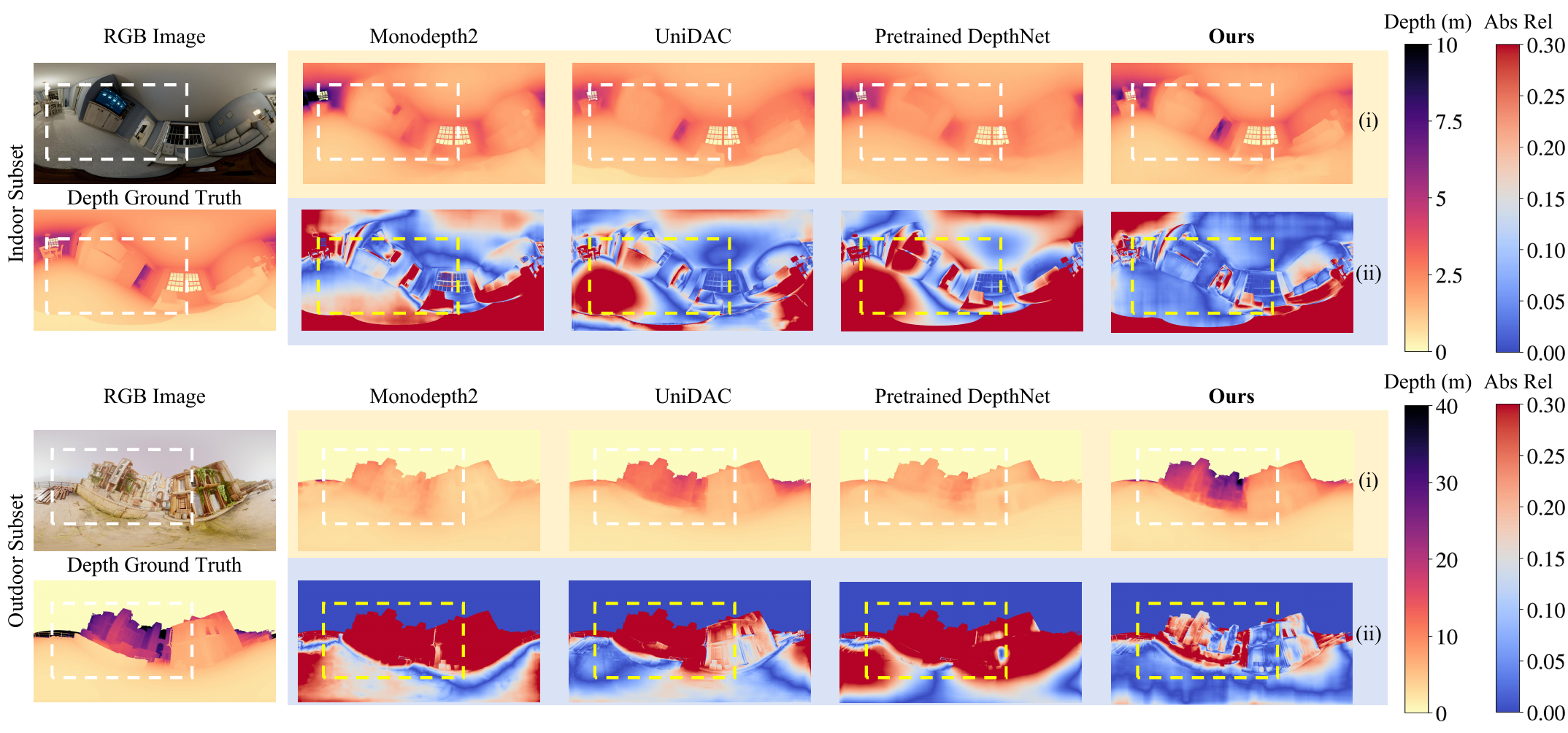}
\caption{
Qualitative comparison of monocular depth estimation on equirectangular video sequences from the TartanAir V2 dataset: (i) estimated depth maps; (ii) corresponding error maps. White and yellow dashed contours highlight regions with substantial improvements in depth estimation accuracy.
}
\label{fig:discussion_depth_ERP}
\end{figure*}

\subsection{Applicability to Different Camera Models}
\label{sec.applicability}

Although S$^2$D-FVO is developed primarily for monocular fisheye cameras, its spherical geometric pipeline is not confined to a specific camera model. For any single-viewpoint camera model, 2-D visual features can be unprojected into bearing vectors on the unit sphere and further parameterized by spherical longitude and latitude in the ERP domain, providing a unified representation for the subsequent geometric operations. Unlike visual odometry systems whose residuals and Jacobians depend on a specific camera projection model, the proposed system can be adapted to different camera models by replacing only the corresponding projection and unprojection functions, while retaining the same spherical-to-ERP parameterization, residual definitions, optimization variables, analytical Jacobians, and numerical optimization procedures.

\begin{figure}[!tp]
\centering
\includegraphics[width=\columnwidth]{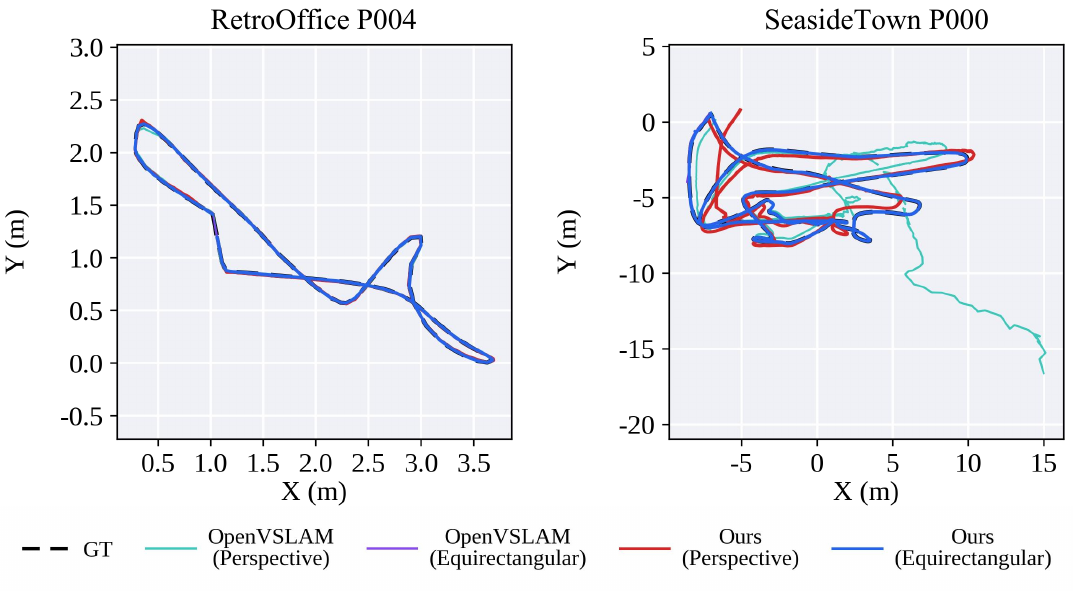}
\caption{
Qualitative comparison of monocular visual odometry on perspective and equirectangular video sequences from the TartanAir V2 dataset.}
\label{fig:discussion_vo}
\end{figure}

The applicability of the proposed S$^2$D-FVO framework across different camera models is further evaluated through both monocular visual odometry and depth estimation experiments on the TartanAir V2 dataset, as shown in Tables~\ref{tab:camera_generalization} and~\ref{tab:depth_camera_generalization}, as well as Figs.~\ref{fig:discussion_vo}, \ref{fig:discussion_depth_front}, and \ref{fig:discussion_depth_ERP}. For visual odometry, the same pose estimation and optimization implementation is used across different camera models, with only the corresponding projection and unprojection functions replaced. The results demonstrate that S$^2$D-FVO can be readily applied to different camera models. Moreover, S$^2$D-FVO achieves lower ATE on equirectangular video sequences than on perspective video sequences. Similar improvements are observed in depth estimation accuracy, with Abs Rel reduced by 35.22\% and 48.40\% for indoor and outdoor scenarios, respectively. These results further suggest that a wider FoV can improve the accuracy of relative poses estimated through spherical bundle adjustment, thereby benefiting visual odometry and providing more reliable poses for view synthesis in unsupervised monocular depth learning.

\begin{table*}[!tp]
\centering
\caption{
Quantitative comparison of monocular visual odometry on perspective and equirectangular video sequences from the TartanAir V2 dataset. Results are reported in terms of ATE (m), with lower values indicating better performance. The best results are shown in bold font.
}
\label{tab:camera_generalization}

\begin{tabular*}{\textwidth}{
@{\extracolsep{\fill}}
llccccccccc
@{}
}
\toprule

\multirow{2}{*}{\textbf{Camera Model}}
& \multirow{2}{*}{\textbf{Method}}
& \multicolumn{3}{c}{\textbf{House}}
& \multicolumn{3}{c}{\textbf{RetroOffice}}
& \multicolumn{3}{c}{\textbf{SeasideTown}} \\

\cmidrule(lr){3-5}
\cmidrule(lr){6-8}
\cmidrule(lr){9-11}

&
& \textbf{H0}
& \textbf{H5}
& \textbf{H6}
& \textbf{R0}
& \textbf{R2}
& \textbf{R4}
& \textbf{S0}
& \textbf{S1}
& \textbf{S3} \\

\midrule

\multirow{2}{*}{Perspective}
& OpenVSLAM~\cite{sumikura2019openvslam}
& $\times$ & $\times$ & \textbf{0.015}
& \textbf{0.121} & \textbf{0.007} & 0.013
& 3.474 & \textbf{0.059} & \textbf{0.104} \\

& \textbf{Ours}
& \textbf{2.978} & \textbf{1.660} & 1.249
& 0.364 & 0.077 & \textbf{0.009}
& \textbf{2.856} & 5.447 & 2.691 \\

\midrule

\multirow{2}{*}{Equirectangular}
& OpenVSLAM~\cite{sumikura2019openvslam}
& \textbf{0.010} & \textbf{0.008} & \textbf{0.005}
& 0.005 & 0.003 & 0.002
& 0.048 & 0.035 & 0.023 \\

& \textbf{Ours}
& 0.047 & 0.015 & 0.016
& \textbf{0.003} & \textbf{0.003} & \textbf{0.002}
& \textbf{0.017} & \textbf{0.023} & \textbf{0.017} \\

\bottomrule
\end{tabular*}
\end{table*}

\begin{table}[!tp]
\centering
\caption{
Quantitative comparison of monocular depth estimation on perspective and equirectangular video sequences from the indoor and outdoor scenarios
of the TartanAir V2 dataset. The best results are shown in bold font.
}
\label{tab:depth_camera_generalization}
\begin{tabular*}{\columnwidth}{
@{\extracolsep{\fill}}
llccc
@{}
}
\toprule

\textbf{Camera Model}
& \textbf{Method}
& \textbf{Abs Rel} $\downarrow$
& \textbf{RMSE (m)} $\downarrow$
& $\boldsymbol{\delta_1}$ $\uparrow$ \\
\midrule

\multicolumn{5}{c}{\textbf{Indoor}} \\
\midrule

\multirow{4}{*}{Perspective}
& Monodepth2~\cite{godard2019digging} & 0.291 & 0.673 & 0.536 \\
& UniDAC~\cite{ganesan2026unidac} & \textbf{0.155} & \textbf{0.412} & \textbf{0.791} \\
& Pretrained DepthNet & 0.290 & 0.664 & 0.526 \\
& \textbf{Ours} & 0.230 & 0.688 & 0.536 \\

\midrule

\multirow{4}{*}{Equirectangular}
& Monodepth2~\cite{godard2019digging} & 0.222 & 0.660 & 0.653 \\
& UniDAC~\cite{ganesan2026unidac} & 0.162 & \textbf{0.440} & 0.762 \\
& Pretrained DepthNet & 0.217 & 0.583 & 0.642 \\
& \textbf{Ours} & \textbf{0.149} & 0.447 & \textbf{0.800} \\

\midrule
\multicolumn{5}{c}{\textbf{Outdoor}} \\
\midrule

\multirow{4}{*}{Perspective}
& Monodepth2~\cite{godard2019digging} & 0.407 & 5.948 & 0.398 \\
& UniDAC~\cite{ganesan2026unidac} & \textbf{0.218} & \textbf{4.139} & \textbf{0.678} \\
& Pretrained DepthNet & 0.421 & 5.608 & 0.397 \\
& \textbf{Ours} & 0.374 & 5.331 & 0.452 \\

\midrule

\multirow{4}{*}{Equirectangular}
& Monodepth2~\cite{godard2019digging} & 0.414 & 7.188 & 0.374 \\
& UniDAC~\cite{ganesan2026unidac} & 0.288 & 6.219 & 0.520 \\
& Pretrained DepthNet & 0.406 & 7.569 & 0.409 \\
& \textbf{Ours} & \textbf{0.193} & \textbf{4.166} & \textbf{0.751} \\

\bottomrule
\end{tabular*}
\end{table}

\subsection{Limitations}
\label{sec.limitations}
Although the proposed framework effectively couples monocular fisheye depth estimation with spherical bundle adjustment to improve scale consistency in visual odometry, the proposed scale regularization strategy still has two main limitations. The first limitation is related to the reliability of the scale reference derived from DepthNet. This reference is determined by aligning the median depth of the DepthNet estimates with that of the 3-D landmarks, implicitly assuming that these estimates are sufficiently accurate for scale alignment. Nevertheless, DepthNet may exhibit systematic depth estimation errors in specific scenes, causing these estimates to be consistently overestimated or underestimated. Although the median operation can suppress isolated outliers, it cannot eliminate a systematic bias affecting a substantial portion of the depth estimates. The resulting scale reference may therefore be biased and propagate an incorrect scale into the local reconstruction.

The second limitation is related to the reliability of 3-D landmark depth determined during spherical bundle adjustment. This reliability depends strongly on the parallax available for triangulation. When the parallax is limited, the viewing rays corresponding to different observations have only a small angular separation, such that small errors in matched correspondences or camera poses can lead to large errors in the recovered 3-D landmark depth. In the extreme case of pure rotation, the parallax is zero and 3-D landmark depth cannot be determined from the corresponding observations. Therefore, the recovered 3-D landmark depth may be unreliable under such conditions. Since the scale reference derived from DepthNet is incorporated into the scale regularization by minimizing the discrepancy between the median 3-D landmark depth and that estimated by DepthNet, even an accurate scale reference may fail to provide an effective scale constraint when the 3-D landmark depth is unreliable.

The two limitations above indicate that reliable scale regularization requires both an accurate scale reference from DepthNet and sufficiently reliable 3-D landmark depth in spherical bundle adjustment. As this study focuses on validating the effectiveness of coupling depth estimates with spherical bundle adjustment, the uncertainties associated with the scale reference and 3-D landmark depth are not explicitly modeled during optimization. Explicitly accounting for both sources of uncertainty is therefore important for improving the robustness of scale regularization under challenging conditions.

\section{Conclusion and Future Work}
\label{sec.Conclusion}

This article presented a scale-consistent monocular fisheye visual odometry framework S$^2$D-FVO that couples monocular depth estimation with spherical bundle adjustment. The main contributions of this study are threefold. 
First, it develops a spherical bundle adjustment formulation that combines
tangent-space angular reprojection residuals with median depth-based scale
regularization to mitigate scale drift without additional sensing modalities. Second, it demonstrates that spherical bundle adjustment can substantially improve unsupervised fisheye depth learning by providing more reliable relative camera poses for photometric supervision. Third, comprehensive evaluations on multiple synthetic and real-world fisheye datasets validate the effectiveness of the proposed framework for both monocular depth estimation and visual odometry. The proposed method achieves SoTA performance among compared unsupervised monocular depth estimation methods across all evaluated datasets, remains competitive with recently proposed metric depth foundation models despite adopting a lightweight DepthNet architecture, and consistently reduces scale drift while improving trajectory estimation accuracy across diverse indoor and outdoor scenarios.

Future work will focus on developing adaptive scale regularization strategies capable of quantifying the reliability of DepthNet estimates and triangulated 3-D landmark depth, thereby reducing the influence of unreliable depth information on spherical bundle adjustment.

\bibliographystyle{IEEEtran}
\bibliography{sn-bibliography-revised}
\end{document}